\documentclass{article}

\usepackage{listings}
\usepackage{microtype}
\usepackage{graphicx}
\usepackage{booktabs} %

\usepackage{hyperref} %

\usepackage[accepted]{icml2024}

\usepackage{amsmath}
\usepackage{amssymb}
\usepackage{mathtools}
\usepackage{amsthm}

\usepackage[capitalize,noabbrev]{cleveref}

\theoremstyle{plain}

\theoremstyle{definition}

\theoremstyle{remark}

\usepackage{amsmath}
\usepackage{amssymb}
\usepackage{bm}
\usepackage{mathtools}

\def\ry{{\textnormal{y}}}

\def\rvf{{\mathbf{f}}}

\def\rvx{{\mathbf{x}}}
\def\rvy{{\mathbf{y}}}

\def\vzero{{\bm{0}}}
\def\vone{{\bm{1}}}

\def\vmu{{\bm{\mu}}}

\def\vtheta{{\bm{\theta}}}

\def\va{{\bm{a}}}
\def\vb{{\bm{b}}}

\def\vf{{\bm{f}}}
\def\vg{{\bm{g}}}
\def\vh{{\bm{h}}}

\def\vsigma{{\bm{\sigma}}}

\def\vu{{\bm{u}}}
\def\vv{{\bm{v}}}

\def\vx{{\bm{x}}}
\def\vy{{\bm{y}}}

\def\vz{{\bm{z}}}

\def\mA{{\bm{A}}}
\def\mB{{\bm{B}}}
\def\mC{{\bm{C}}}
\def\mD{{\bm{D}}}
\def\mE{{\bm{E}}}
\def\mF{{\bm{F}}}
\def\mG{{\bm{G}}}

\def\mI{{\bm{I}}}

\def\mK{{\bm{K}}}

\def\mS{{\bm{S}}}

\def\mV{{\bm{V}}}
\def\mW{{\bm{W}}}
\def\mX{{\bm{X}}}

\def\mZ{{\bm{Z}}}

\def\mSigma{{\bm{\Sigma}}}

\DeclareMathAlphabet{\mathsfit}{\encodingdefault}{\sfdefault}{m}{sl}
\SetMathAlphabet{\mathsfit}{bold}{\encodingdefault}{\sfdefault}{bx}{n}
\newcommand{\tens}[1]{\bm{\mathsfit{#1}}}
\def\tA{{\tens{A}}}
\def\tB{{\tens{B}}}

\def\tG{{\tens{G}}}
\def\tH{{\tens{H}}}

\def\tJ{{\tens{J}}}

\def\tU{{\tens{U}}}
\def\tV{{\tens{V}}}

\def\gC{{\mathcal{C}}}

\def\gF{{\mathcal{F}}}

\def\gL{{\mathcal{L}}}

\def\gN{{\mathcal{N}}}
\def\gO{{\mathcal{O}}}

\def\gX{{\mathcal{X}}}
\def\gY{{\mathcal{Y}}}

\def\sD{{\mathbb{D}}}

\def\sR{{\mathbb{R}}}

\DeclareSymbolFont{bbold}{U}{bbold}{m}{n}
\DeclareSymbolFontAlphabet{\mathbbold}{bbold}

\newcommand{\etens}[1]{\mathsfit{#1}}

\def\etA{{\etens{A}}}

\newcommand{\E}{\mathbb{E}}

\newcommand{\R}{\mathbb{R}}

\newcommand{\softmax}{\mathrm{softmax}}
\newcommand{\onehot}{\mathrm{onehot}}

\DeclareMathOperator*{\argmin}{arg\,min}

\DeclareMathOperator{\diag}{diag}

\DeclareMathOperator{\rank}{rank} %

\newcommand{\jac}{\mathrm{J}}
\newcommand{\hess}{\mathrm{H}}
\newcommand{\ggn}{\mathrm{G}}

\newcommand{\kfac}{\mathrm{KFAC}}

\usepackage{comment}
\usepackage{paracol}
\usepackage{blindtext}
\usepackage{enumitem}

\usepackage{xspace}
\newcommand*{\ie}{i.e.\@\xspace}
\newcommand*{\iid}{i.i.d.\@\xspace}
\newcommand*{\wrt}{w.r.t.\@\xspace}
\newcommand*{\eg}{e.g.\@\xspace}

\definecolor{VectorBlack}{RGB}{34, 34, 34}
\definecolor{VectorGray}{RGB}{239, 238, 237}

\definecolor{VectorBlue}{RGB}{59, 69, 227}
\definecolor{VectorPink}{RGB}{253, 8, 238}
\definecolor{VectorOrange}{RGB}{250, 173, 26}
\definecolor{VectorTeal}{RGB}{82, 199, 222}
\newcommand{\colored}[2][VectorBlue]{{\color{#1}#2}}

\hypersetup{%
  colorlinks,
  citecolor = VectorBlue,%
  linkcolor = VectorBlue,%
  urlcolor = VectorPink,%
}%
\crefname{listing}{snippet}{snippets}
\lstdefinestyle{vector_institute}{
  backgroundcolor=\color{VectorGray!50},
  commentstyle=\bfseries\color{VectorBlue},
  keywordstyle=\bfseries\color{VectorBlack},
  numberstyle=\tiny\color{VectorBlack!50},
  stringstyle=\bfseries\color{VectorBlue},
  basicstyle=\ttfamily\scriptsize,
  xleftmargin=3.2ex,
  breakatwhitespace=false,
  breaklines=true,
  captionpos=t,
  keepspaces=true,
  numbers=left,
  numbersep=7pt,
  showspaces=false,
  showstringspaces=false,
  showtabs=false,
  tabsize=2,
  escapebegin={\color{VectorBlue}},
  linewidth=\linewidth,
  mathescape=true,
}
\usepackage{caption}
\newcommand{\repourl}{https://github.com/f-dangel/kfac-tutorial/blob/main}
\newcommand{\repofile}[1]{
  \href{\repourl/kfs/#1.py}{\texttt{kfs/\detokenize{#1}.py}}
}
\newcommand{\codeblock}[1]{
  \lstinputlisting[%
  language=python,%
  caption={\repofile{#1}},%
  label=#1,%
  ]{kfs/#1.py}
}

\usepackage{nicefrac}
\DeclareMathOperator{\rvec}{rvec}
\DeclareMathOperator{\cvec}{cvec}
\let\vec\relax %
\DeclareMathOperator{\vec}{vec}
\DeclareMathOperator{\mat}{mat}
\DeclareMathOperator{\lin}{lin}

\usepackage{mdframed}
\mdfdefinestyle{custom}{%
  linecolor=black,%
  topline=false,%
  bottomline=false,%
  rightline=false,%
  linewidth=1.25pt,%
  backgroundcolor=VectorGray!50,%
  innerleftmargin=5pt,%
}
\theoremstyle{definition}
\newmdtheoremenv[style=custom]{definition}{Definition}
\newmdtheoremenv[style=custom]{setup}{Setup}
\newmdtheoremenv[%
style=custom,%
linecolor=VectorOrange,%
backgroundcolor=VectorOrange!10,%
]{caveat}{Caveat}
\newmdtheoremenv[%
style=custom,%
linecolor=VectorTeal,%
backgroundcolor=VectorTeal!10,%
]{test}{Test}
\newmdtheoremenv[%
style=custom,%
linecolor=VectorTeal,%
backgroundcolor=VectorTeal!10,%
]{example}{Example}

\usepackage{tikz}
\usetikzlibrary{arrows.meta}
\usetikzlibrary{positioning}

\begin{document}

\globalcounter{figure}
\globalcounter{example}

\onecolumn
\newcommand{\papertitle}{%
  Kronecker-factored Approximate Curvature (KFAC) From Scratch
}%
\title{\papertitle}

\icmltitlerunning{\papertitle}

\icmltitle{\papertitle}

\icmlsetsymbol{equal}{*}

\begin{icmlauthorlist}
\icmlauthor{Felix Dangel}{equal,vector}
\icmlauthor{B\'alint Mucs\'anyi}{equal,tue}
\icmlauthor{Tobias Weber}{equal,tue}
\icmlauthor{Runa Eschenhagen}{cambridge}
\end{icmlauthorlist}

\icmlaffiliation{vector}{Vector Institute, Canada}
\icmlaffiliation{cambridge}{University of Cambridge, United Kingdom}
\icmlaffiliation{tue}{University of T\"ubingen, Germany}

\icmlcorrespondingauthor{Felix Dangel}{fdangel@vectorinstitute.ai}
\icmlkeywords{KFAC, Natural gradient descent}

\vskip 0.3in

\printAffiliationsAndNotice{\icmlEqualContribution} %

\vspace*{-2ex}

\begin{center}
\begin{tikzpicture}
    \node[inner sep=0.7pt, opacity=0.5, draw opacity=1, draw=black!50!white, ultra thick, rounded corners]{\includegraphics[scale=0.66]{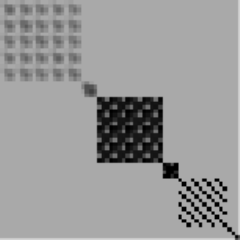}};
\end{tikzpicture}
\end{center}

\vfill

\begin{abstract}
  Kronecker-factored approximate curvature \citep[KFAC,][]{martens2015optimizing} is arguably one of the most prominent curvature approximations in deep learning.
  Its applications range from optimization to Bayesian deep learning, training data attribution with influence functions, and model compression or merging.
  While the intuition behind KFAC is easy to understand, its implementation is tedious: It comes in many flavours, has common pitfalls when translating the math to code, and is challenging to test, which complicates ensuring a properly functioning implementation.
  Some of the authors themselves have dealt with these challenges and experienced the discomfort of not being able to fully test their code.
  Thanks to recent advances in understanding KFAC, we are now able to provide test cases and a recipe for a reliable KFAC implementation.
  \emph{This tutorial is meant as a ground-up introduction to KFAC.}
  In contrast to the existing work, our focus lies on providing both math and code side-by-side and providing test cases based on the latest insights into KFAC that are scattered throughout the literature.
  We hope this tutorial provides a contemporary view of KFAC that
  allows beginners to gain a deeper understanding of this curvature approximation while lowering the barrier to its implementation, extension, and usage in practice.
\end{abstract}

\vfill

\paragraph{Version:} \today\,(v1.0.0)

\paragraph{About the length of this document.}
Before you close this document because you saw the page count:
the \emph{effective length is much shorter than suggested by its page number}.
This is because \emph{we use an experimental two-column layout which presents text and code in parallel} and leads to a large amount of white space.
The left column contains the main text with explanations and mathematical descriptions.
The right column accompanies the left one with code snippets to make the ideas precise in code; it can safely be skipped if you are in a rush.
And if you already know the basics, it suffices to read the KFAC-specific part (pages \pageref{sec:kfac-overview}--\pageref{sec:kfac-cheatsheet}).

\paragraph{Follow along in code.} The \LaTeX\,\& Python source code is available at~\href{\repourl}{\texttt{github.com/f-dangel/kfac-tutorial}}.
This allows you to run the code as you read:
Clone the repository and follow the installation instructions.
You can then run each snippet from the repository root, for instance by calling \texttt{python kfs/basics/forward\_pass.py}.
If you find typos or have suggestions for improving explanations, math, or code, please open issues and pull requests.
In doing so, you are contributing to making this tutorial a valuable reference for newcomers.

\vspace{\baselineskip}

\clearpage

\section*{How to Read This Document?}

\begin{figure*}[!h]
  \centering
  \begin{tikzpicture}[ultra thick, node distance=3cm]
  \tikzstyle{box} = [rectangle, align=center, minimum height=2.5ex, rounded corners, draw=black, fill=VectorGray!50!white, inner sep=5pt]
  \tikzstyle{answer} = [minimum height=2.5ex, fill=white, align=center]

  \node [circle, fill=black] (start) {};
  \node [anchor=south] at (start.north) {Start};

  \node [box, below=1cm of start] (check-basics) {I know derivatives (Jacobians, Hessians)\\ and the maximum likelihood interpretation\\ of empirical risk minimization};
  \draw[-{LaTeX}] (start) -- (check-basics);

  \node [box, below right=of check-basics] (read-basics) {Read the basics (\cref{sec:basics})};
  \node [box, below=of check-basics] (read-cheatsheet) {Read the cheatsheet (\cref{sec:cheatsheet-basics})};
  \draw[-{LaTeX}] (check-basics) to[out=270, in=90] node[midway, answer] {yes} (read-cheatsheet);
  \draw[-{LaTeX}] (check-basics) to[out=270, in=90] node[midway, answer] {no} (read-basics);
  \draw[-{LaTeX}] (read-basics) to[out=0, in=0] (check-basics);

  \node [box, below=of read-cheatsheet] (scaffold) {Scaffold (\cref{sec:kfac-overview})};
  \draw[-{LaTeX}] (read-cheatsheet) to node [midway, answer] {Looks good, give \\ me an overview\\ of KFAC} (scaffold);
  \draw[-{LaTeX}] (read-cheatsheet) to [out=315, in=225] node [midway, answer] {I did not understand \\ some of the concepts \\ or notation} (read-basics);

  \node[box, below=of scaffold] (kfac-cheatsheet) {Read KFAC cheatsheet (\cref{sec:kfac-cheatsheet})};
  \node[box, below right=of scaffold] (kfac-details) {KFAC details (\cref{sec:kfac-linear})};
  \draw[-{LaTeX}] (scaffold) to[out=315, in=135] node [midway, answer] {Give me all \\ the details} (kfac-details);
  \draw[-{LaTeX}] (scaffold) to[out=270, in=90] node [midway, answer] {I just want the \\ one page summary} (kfac-cheatsheet);
  \draw[-{LaTeX}] (kfac-details) to[out=270, in=0] (kfac-cheatsheet);

  \node[box, below=of kfac-cheatsheet] (outlook) {Future plans (\cref{sec:outlook})};
  \draw[-{LaTeX}] (kfac-cheatsheet) to node [midway, answer] {What's next?} (outlook);

  \node[circle, fill=black, below=1cm of outlook] (end) {};
  \node[anchor=north] at (end.south) {End};
  \draw[-{LaTeX}] (outlook) -- (end);
\end{tikzpicture}
\end{figure*}

\clearpage

\setcounter{tocdepth}{2}
\tableofcontents
\clearpage

\section{Preface}\label{sec:preface}
\paragraph{Relevance.} Kronecker-factored approximate curvature (KFAC) is arguably one of the most prominent curvature approximations in deep learning.
Its applications range from optimization~\cite{martens2015optimizing,grosse2016kroneckerfactored,eschenhagen2023kroneckerfactored,benzing2022gradient,petersen2023isaac} to Bayesian deep learning~\cite{daxberger2021laplace}, training data attribution with influence functions~\cite{grosse2023studying,bae2024training}, and model pruning~\cite{wang2019eigendamage} and merging~\cite{tam2024merging}.
The intuition behind KFAC is easy to understand: it approximates curvature information in the form of Kronecker-factored matrices that are both cheap to store and compute with.
There exist several packages that compute KFAC~\cite{botev2022kfac-jax,dangel2020backpack,osawa2023asdl,grosse2023studying}.
However, their details are often undocumented, making them hard to extend, \eg adding support for new layers, or adding features like eigenvalue correction~\cite{george2018fast}.
We claim that achieving a bug-free KFAC implementation is non-trivial, and our understanding is still evolving.

\paragraph{Goals.}
This is an attempt to bundle the scattered knowledge about KFAC into a single document, explain all the technicalities and pitfalls, and present tests to ensure bug-free implementations.
Our goal with this tutorial is to explain, from scratch, how to obtain a KFAC implementation that is easy to build on, allowing newcomers to fully grasp the concept and providing experienced practitioners a code base to build on.
We use PyTorch~\cite{paszke2019pytorch} and rely on its modular interface (\texttt{torch.nn} modules), which is familiar to most deep learning practitioners.
The tutorial's goals are the following:
\begin{enumerate}
\item \textbf{[Basics] Provide a self-contained introduction to curvature matrices (the `C' in KFAC).}
  To understand KFAC, we first need to understand the objects it aims to approximate, namely, curvature matrices of neural network loss functions.
  These curvature matrices are all based on the Hessian, which gives rise to the curvature if we Taylor-expand the loss up to quadratic order.
  We will introduce common approximations of the Hessian, like the generalized Gauss-Newton (GGN), Fisher, and empirical Fisher (EF), and how to work with them through autodiff (\cref{sec:basics}).
  This establishes the basis of our tutorial and serves as a foundation to test our KFAC implementation (\cref{sec:cheatsheet-basics} shows a compact summary).

\item \textbf{[Intuition] Show how Kronecker products naturally emerge in curvature matrices (the `KF' in KFAC).}
  Doing so motivates Kronecker products as a `natural' structure to approximate curvature, which is what KFAC does (\cref{sec:kfac-overview}).

\item \textbf{[Code] Explain how to implement and test KFAC (the `A' in KFAC).}
  Present the different flavours of KFAC and show how they are related to the curvature matrices, which allows them to be tested (\cref{sec:kfac-linear}).
\end{enumerate}
\paragraph{Scope \& limitations.} To make this tutorial pedagogically valuable while providing a different experience than just reading the original papers introducing KFAC, we had to make compromises, as understanding and implementing KFAC in all its different variations quickly becomes daunting.
Here is a rough breakdown of the degrees of freedom:
\begin{center}

  \begin{tabular}[!h]{cc}
    \textbf{Degree of freedom}
    &
      \textbf{Choices}
    \\
    Which software framework should we use?
    &
      $
      \begin{Bmatrix}
        \text{PyTorch}
        \\
        \text{JAX}
      \end{Bmatrix}
      $
    \\
    & $\times$
    \\
    Which neural network layers should we discuss?
    &
      $
      \begin{Bmatrix}
        \text{fully-connected layers}
        \\
        \text{convolutional layers}
        \\
        \text{recurrent layers}
        \\
        \text{attention layers}
      \end{Bmatrix}
      $
    \\
    & $\times$
    \\
    Which curvature matrices should we discuss?
    &
      $
      \begin{Bmatrix}
        \text{Hessian}
        \\
        \text{GGN}
        \\
        \text{Fisher (type-I/II, empirical)}
      \end{Bmatrix}
      $
    \\
    & $\times$
    \\
    Which flattening convention should we choose?
    &
      $
      \begin{Bmatrix}
        \cvec\,\text{(used by literature)}
        \\
        \rvec\,\text{(used by implementations)}
      \end{Bmatrix}
      $
    \\
    & $\times$
    \\
    Should we discuss flavours from the presence of weight sharing?
    &
      $
      \begin{Bmatrix}
        \text{yes (expand and reduce)}
        \\
        \text{no}
      \end{Bmatrix}
      $
  \end{tabular}
\end{center}
We decided to center this tutorial around the original work from~\citet{martens2015optimizing} that corresponds to
\begin{align*}
  \begin{Bmatrix}
    \text{fully-connected layers}
  \end{Bmatrix}
  \times
  \begin{Bmatrix}
    \text{type-I Fisher}
  \end{Bmatrix}
  \times
  \begin{Bmatrix}
    \cvec
  \end{Bmatrix}
  \times
  \begin{Bmatrix}
    \text{no weight sharing}
  \end{Bmatrix}\,.
\end{align*}
However, to provide some additional value, this document aims to provide an introduction to the following flavours:
\begin{align*}
  \begin{Bmatrix}
    \text{PyTorch}
  \end{Bmatrix}
  \times
  \begin{Bmatrix}
    \text{fully-connected}
  \end{Bmatrix}
  \times
  \begin{Bmatrix}
    \text{GGN}
    \\
    \text{type-I/II Fisher}
    \\
    \text{empirical Fisher}
  \end{Bmatrix}
  \times
  \begin{Bmatrix}
    \rvec
    \\
    \cvec
  \end{Bmatrix}
  \times
  \begin{Bmatrix}
    \text{no weight sharing}
  \end{Bmatrix}\,.
\end{align*}
This allows us to (i) highlight challenges when translating math to code, (ii) point out various connections between curvature matrices and KFAC, and (iii) produce a working, tested version of the original KFAC paper with slight generalizations.
A more fully-featured KFAC implementation is provided by the \texttt{curvlinops} library~\cite{dangel2025position}.

\clearpage

\columnratio{0.42}
\begin{paracol}{2}
  \section{Basics}\label{sec:basics}
  \switchcolumn[1]*
\codeblock{basics/reduction_factors}
\switchcolumn[0]

This tutorial is meant to be self-contained.
Therefore, we will start with an extensive introduction to KFAC-relevant concepts.
This allows us to build the core functionality we will later need to verify our implementation.

\paragraph{Roadmap.} First, we introduce the empirical risk (`the loss', \cref{subsec:empirical-risk-minimization}) whose curvature KFAC approximates, and neural networks (\cref{subsec:deep-neural-networks}).
One recurring theme in our discussion will be that the loss and neural net have probabilistic interpretations in most deep learning settings: Minimizing the empirical risk corresponds to maximum likelihood estimation where the neural net models a likelihood (\cref{subsec:probabilistic-interpretation}).
Next, since curvature information is based on the Hessian, which contains second-order partial derivatives, we will talk about first- and second-order derivatives, and how to compute with them using PyTorch's automatic differentiation (\cref{subsec:derivatives}).
We conclude with an introduction to all curvature matrices relevant to our discussion (\cref{subsec:curvature-matrices}).
These include the Hessian, generalized Gauss-Newton (GGN) matrix, and different flavours of the Fisher information matrix, which follows from the probabilistic interpretation from \cref{subsec:probabilistic-interpretation}.

\subsection{Empirical Risk Minimization}\label{subsec:empirical-risk-minimization}
We consider supervised learning with a neural network $f\colon \gX \times \Theta \to \gF$ that maps a given input $\vx \in \gX$ from a domain $\gX$ to a prediction $f(\vx, \vtheta) \in \gF$ in a prediction space $\gF$ using parameters $\vtheta \in \Theta$ from a parameter space $\Theta$.
Predictions are scored with a criterion function $c\colon \gF \times \gY \to \sR$ that compares the prediction to the true target $\vy \in \gY$ from a label space $\gY$, producing a single number called the loss on datum $(\vx, \vy)$.

For a data set $\sD = \{(\vx_n, \vy_n) \mid n=1, \dots, N\}$ of collected labelled examples, we evaluate the per-datum criteria and accumulate them into the total loss, using an accumulation factor $R \in \sR$,
\begin{align}\label{eq:empirical_risk}
  \begin{split}
    \gL_{\sD}(\vtheta) & = R \sum_{n=1}^N \ell_n(\vtheta)
    \\
                       & = R \sum_{n=1}^N c(f(\vx_n, \vtheta), \vy_n)\,.
  \end{split}
\end{align}
Common choices for $R$ are $\nicefrac{1}{N}, 1, \nicefrac{1}{N \dim(\gY)}$; see \Cref{basics/reduction_factors} for a function that computes $R$.
The goal of training is to find the parameters $\vtheta$ that reduce the empirical risk $\gL_{\sD}(\vtheta)$ without overfitting to the training data.

\Cref{eq:empirical_risk} disentangles `the loss' into three components: the neural network $f$, the criterion function $c$, and the reduction factor $R$.
The most common loss functions are the square loss for regression and the softmax cross-entropy loss for classification.
We show their criterion and reduction factors in \cref{ex:square_loss,ex:cross_entropy_loss}).

\switchcolumn[1]
\begin{example}[Square loss, \Cref{basics/reduction_factors}]\label{ex:square_loss}
  For least squares regression with vector-valued targets ($\gY = \sR^C = \gF$), the criterion and reduction factor of PyTorch's \texttt{nn.MSELoss} are
  \begin{align*}
    &c(\vf, \vy)
      =
      \frac{1}{2}\sum_{c=1}^C [\vf - \vy]_c^2\,,
    \\
    R
    &=
      \begin{cases}
        2                     & \text{\texttt{reduction="sum"}}
        \\
        \frac{2}{N \dim(\gY)} & \text{\texttt{reduction="mean"}}
      \end{cases}
  \end{align*}
  where $\dim(\gY) = C$ in the vector case, but $\gY = \gF$ could also be a matrix or tensor space.
\end{example}

\begin{example}[Cross-entropy loss, \Cref{basics/reduction_factors}]\label{ex:cross_entropy_loss}
  For classification, with categorical targets ($\gY = \{1, \dots, C\}$ and $\gF = \sR^C$), PyTorch's \texttt{nn.CrossEntropyLoss} uses the following criterion function and reduction factor
  \begin{align*}
    &c(\vf, y)
      =
      - \log([\softmax(\vf)]_y)\,,
    \\
    R
    &=
      \begin{cases}
        1                     & \text{\texttt{reduction="sum"}}
        \\
        \frac{1}{N \dim(\gY)} & \text{\texttt{reduction="mean"}}
      \end{cases}
  \end{align*}
  with $[\softmax(\vf)]_i = \nicefrac{\exp([\vf]_i)}{\sum_{j=1}^C \exp([\vf]_{j})}$.
  For the vector case $\dim(\gY) = 1$, but $\gY, \gF$ could also be compatible matrix or tensor spaces in a more general setup where we aim to classify sequences of categorical labels.
\end{example}
\switchcolumn[0]

\begin{caveat}[Scaling]
  Implementations of loss functions mix the concepts of criterion and reduction.
  This is often fine, but sometimes makes it difficult to translate to new loss functions without accidentally forgetting a factor.
  By keeping both concepts separate, we reduce the chance of introducing scaling bugs.
\end{caveat}

\switchcolumn[1]
\codeblock{basics/forward_pass}
\switchcolumn[0]
\subsection{Deep Neural Networks}\label{subsec:deep-neural-networks}
We consider sequential neural networks as they are simple and widely used.
They are composed of $L$ layers $f^{(l)}(\cdot, \vtheta^{(l)}), l=1,\dots, L$, each of which can have its own parameters $\vtheta^{(l)}$.
The whole network is simply a stack of layers, \ie $f = f^{(L)} \circ \dots \circ f^{(1)}$, and the evaluation produces a set of intermediate features,
\begin{align*}
  \vx^{(l)} = f^{(l)}(\vx^{(l-1)}, \vtheta^{(l)}), \quad l=1,\dots, L,
\end{align*}
starting with $\vx^{(0)} \leftarrow \vx$, ending in $\vx^{(L)} \leftarrow f(\vx, \vtheta)$.
We refer to the hidden representations $\vx^{(l-1)}$ as the \emph{input to layer $l$}, and to $\vx^{(l)}$ as the \emph{output of layer $l$}.
Some layers have empty parameters, e.g.\,activation, pooling, or dropout layers.

To compute KFAC, we will need access to the inputs and outputs of certain layers.
We can obtain these by intercepting the neural net's forward pass using PyTorch's hook mechanism, see \Cref{basics/forward_pass}.

\subsection{Probabilistic Interpretation}\label{subsec:probabilistic-interpretation}
So far, we have considered minimizing an empirical risk over a data set given an arbitrary criterion function $c$.
Now, we take a step back and first describe empirical risk minimization as an approximation of minimizing the intractable population risk.
This perspective connects to a probabilistic interpretation of empirical risk minimization as maximum likelihood estimation that will occur throughout the tutorial, \eg when we define the Fisher information matrix as curvature matrix (\cref{subsec:curvature-matrices}).

\paragraph{Empirical risk as expectation.} Recall the empirical risk from \cref{eq:empirical_risk} which we minimize during training,
\begin{align*}
  \min_{\vtheta} \gL_{\sD}(\vtheta) = \min_{\vtheta} R \sum_{n=1}^N \ell_n(\vtheta)\,.
\end{align*}
Why is it called an `empirical' risk?
Because it can be expressed as expectation over an empirical distribution in the following sense:

Assume there exists a data-generating process $p_{\text{data}}(\rvx, \rvy)$ over input-target pairs.
Ideally, we want to minimize the risk over this distribution,
\begin{align*}
  \argmin_{\vtheta} \E_{(\vx, \vy) \sim p_{\text{data}}(\rvx, \rvy)}[c(f(\vx, \vtheta), \vy)]\,.
\end{align*}
However, $p_{\text{data}}$ is intractable.
Therefore, we draw a finite collection of samples into a data set
\begin{align*}
  \sD = \{ (\vx_n, \vy_n) \mid (\vx_n, \vy_n) \stackrel{\text{\iid}}{\sim} p_{\text{data}}(\vx, \vy) \}\,.
\end{align*}
Then, we can replace the intractable data-generating process $p_{\text{data}}$ with the tractable empirical distribution $p_{\sD}(\rvx, \rvy)$ implied by data set $\sD$.
It consists of a uniformly weighted sum of delta peaks around the collected data points,
\begin{align*}
  p_{\sD}(\rvx, \rvy) = \frac{1}{N} \sum_{n=1}^N \delta(\rvx - \vx_n) \delta(\rvy - \vy_n)\,.
\end{align*}
This turns risk minimization into a tractable task:
\begin{align*}
  & \argmin_{\vtheta} \E_{(\vx, \vy) \sim \colored{p_{\text{data}}(\rvx, \rvy)}}[c(f(\vx, \vtheta), \vy)]
  \\
  \approx & \argmin_{\vtheta} \E_{(\vx, \vy) \sim \colored{p_{\sD}(\rvx, \rvy)}}[c(f(\vx, \vtheta), \vy)].
  \\
  \intertext{By writing out the expectation, we obtain}
  =       & \argmin_{\vtheta} \frac{1}{N} \sum_n c(f(\vx_n, \vtheta), \vy_n).
  \\
  \intertext{Note that the minimized objective is the empirical risk in \cref{eq:empirical_risk}
  scaled by $\nicefrac{1}{NR}$.
  However, we can arbitrarily scale objectives without changing the
  location of their minima.
  Hence, the above is equivalent to minimizing the empirical risk
  }
  =& \argmin_{\vtheta} \gL_{\sD}(\vtheta)\,.
\end{align*}
We have thus shown that empirical risk minimization is minimizing a (scaled) expectation of the criterion $c$ over an empirical density $p_{\sD}(\rvx, \rvy)$.

\paragraph{Neural nets parameterize likelihoods.}
Let's approach this from a probabilistic perspective now.
Assume we want to learn $p_{\text{data}}(\rvx, \rvy) = p_{\text{data}}(\rvy \mid \rvx) p_{\text{data}}(\rvx)$ using a parameterized density of the form $p(\rvx, \rvy \mid \vtheta) = p(\rvy \mid \rvx, \vtheta) p_{\text{data}}(\rvx)$ where $p_{\text{data}}(\rvx) = \int p_{\text{data}}(\rvx, \rvy)\ \mathrm{d}\rvy$ is the marginal density of the input data.
Note that we only model the likelihood of the labels with parameters $\vtheta$.

One plausible approach to make $p$ resemble $p_{\text{data}}$ is to minimize their KL divergence,
\begin{align*}
  & \argmin_{\vtheta} \mathrm{KL}(p_{\text{data}}(\rvx, \rvy) \mid\mid p(\rvx, \rvy \mid \vtheta))\,.
  \\
  \intertext{We can simplify this expression by substituting the definition of the KL divergence and dropping terms that do not depend on $\vtheta$,}
  \Leftrightarrow & \argmin_{\vtheta} \E_{p_{\text{data}}(\rvx, \rvy)}[- \log( p(\rvx, \rvy \mid \vtheta))]\,.
  \\
  \intertext{This looks very similar to the expected risk from above.
  Next, let's factorize our model distribution using its conditional and marginal densities and drop $p_{\text{data}}(\rvx)$ as it does not depend on $\vtheta$,
  }
  \Leftrightarrow & \argmin_{\vtheta} \E_{p_{\text{data}}(\rvx, \rvy)}[- \log( p(\rvy \mid \rvx, \vtheta))]\,.
                    \intertext{To make this problem tractable, we need to replace the intractable data-generating process $p_{\text{data}}(\rvx, \rvy)$ with the empirical distribution $p_{\sD}(\rvx, \rvy)$ again:}
                    \approx         & \argmin_{\vtheta} \E_{p_{\sD}(\rvx, \rvy)}[- \log( p(\rvy \mid \rvx, \vtheta))].
  \\
  \intertext{Writing out the expectation, we obtain}
  =               & \argmin_{\vtheta} \frac{1}{N} \sum_n - \log( p(\rvy = \vy_n \mid \rvx=\vx_n,\vtheta)).
  \\
  \intertext{To make this expression more similar to the empirical risk, we introduce a general scaling $R$ and change the likelihood's parameterization from $p(\rvy \mid \rvx, \vtheta)$ to $r(\rvy \mid f(\rvx, \vtheta))$ with a neural net $f$:}
  =               & \argmin_{\vtheta} R \sum_n - \log( r(\rvy = \vy_n \mid f(\vx_n,\vtheta)))
\end{align*}
This parameterization makes it clear that the neural network represents a conditional distribution over the labels given the inputs (and parameters).

\paragraph{The criterion is the negative log-likelihood.}
We are now very close to writing down the explicit connection between empirical risk minimization and maximum likelihood estimation.
The last remaining step is to connect the model's likelihood $r(\rvy \mid f(\rvx, \vtheta))$ with the criterion function $c(f(\vx, \vtheta), \vy)$ from empirical risk minimization.
It turns out that empirical risk minimization with square loss (\cref{ex:square_loss}) corresponds to maximum likelihood estimation (or equivalently, negative log-likelihood minimization) of a Gaussian distribution over the labels (\cref{ex:square_loss_probabilistic}).
Similarly, classification with softmax cross-entropy criterion amounts to maximum likelihood estimation where the neural net parameterizes a categorical distribution (\cref{ex:cross_entropy_loss_probabilistic}).
\Cref{basics/label_sampling} provides functions to sample labels from these distributions.

The neural net's interpretation as a likelihood allows using probabilistic concepts to measure similarity for comparing two networks.
This will be useful when we define the Fisher information matrix (\cref{sec:fisher}), a common curvature matrix.

\switchcolumn[1]
\begin{example}[Probabilistic interpretation of the square loss]\label{ex:square_loss_probabilistic}
  For the square loss from \Cref{ex:square_loss}, we have that $c(\vf, \vy) = - \log( \mathrm{const.}
  \cdot \gN(\rvy \mid \vmu = f(\vx, \vtheta), \mSigma = \mI))$ where $\gN(\bullet \mid \vmu, \mSigma)$ is a multivariate Gaussian distribution with mean $\vmu \in \sR^C$ and positive definite covariance $\mSigma \in \sR^{C \times C}$,
  \begin{align*}
    \gN(\rvy \mid \vmu, \mSigma)
    =
    \frac{
    \exp\left( -\frac{1}{2} {(\rvy - \vmu)}^\top \mSigma^{-1} (\rvy - \vmu) \right)
    }{{(2\pi)}^{C/2} \sqrt{\det(\mSigma)}}\,.
  \end{align*}
  We can safely neglect the constant factor for the optimization problem and, by setting the covariance to the identity matrix and the mean to the neural net's prediction, identify that empirical risk minimization with square loss corresponds to maximum likelihood estimation of a Gaussian likelihood with unit covariance and mean parameterized by the network:
  \begin{align*}
    c                             & = \text{\texttt{MSELoss}}
    \\
                                  & \Leftrightarrow
    \\
    r(\rvy \mid f(\rvx, \vtheta)) & = \gN(\rvy \mid \vmu = f(\vx, \vtheta), \mSigma = \mI)\,
    \\
                                  & \Leftrightarrow
    \\
    p(\rvy \mid \rvx, \vtheta)    & = \gN(\rvy \mid \vmu = f(\vx, \vtheta), \mSigma = \mI)\,.
  \end{align*}
\end{example}

\begin{example}[Probabilistic interpretation of softmax cross-entropy loss]\label{ex:cross_entropy_loss_probabilistic}
  For softmax cross-entropy loss from \Cref{ex:cross_entropy_loss}, we have that $c(\vf, y) = - \log( \gC(\ry \mid \vsigma = \softmax(\vf) ))$ where $\gC(\bullet \mid \vsigma)$ is a categorical distribution over $\{1, \dots, C\}$ with probabilities $\vsigma \in \sR^C_{\ge 0}$ and $\vsigma^\top \vone = 1$,
  \begin{align*}
    \gC(\ry \mid \vsigma)
    =
    \prod_{c=1}^C [\vsigma]_c^{\delta_{\ry,c}}\,.
  \end{align*}
  Hence, we can identify that empirical risk minimization with softmax cross-entropy loss amounts to maximum likelihood estimation with a categorical likelihood parameterized by the softmax of the network's output:
  \begin{align*}
    c                             & = \text{\texttt{CrossEntropyLoss}}
    \\
                                  & \Leftrightarrow
    \\
    r(\ry \mid f(\rvx, \vtheta)) & = \gC(\ry \mid \vsigma = \softmax(f(\vx, \vtheta)))\,
    \\
                                  & \Leftrightarrow
    \\
    p(\ry \mid \rvx, \vtheta)    & = \gC(\ry \mid \vsigma = \softmax(f(\vx, \vtheta)))\,.
  \end{align*}
\end{example}

\codeblock{basics/label_sampling}
\switchcolumn[0]

\subsection{Derivatives \& Automatic Differentiation}\label{subsec:derivatives}
Let's talk about derivatives, which play a fundamental role in our goal to understand curvature matrices.
All of them are based on the Hessian, which emerges in a second-order Taylor expansion and contains the second-order derivatives.
Here, we will build up to the Hessian and how to compute with it using PyTorch's automatic differentiation.
For simplicity, we will fully rely on PyTorch's backpropagation and neglect other modes like forward-mode automatic differentiation.

\switchcolumn[0]
\subsubsection{Flattening}
\switchcolumn[1]
\begin{example}[Matrix flattening, \Cref{basics/flattening}]\label{ex:flattening}
  For a matrix
  \begin{equation*}
    \mA = \begin{pmatrix} 1 & 2 \\ 3 & 4 \end{pmatrix}
  \end{equation*}
  we have
  \begin{equation*}
    \rvec(\mA)
    =
    \begin{pmatrix}
      1 \\ 2 \\ 3 \\ 4
    \end{pmatrix}\,,
    \qquad
    \cvec(\mA)
    =
    \begin{pmatrix}
      1 \\ 3 \\ 2 \\ 4
    \end{pmatrix}\,.
  \end{equation*}
\end{example}
\switchcolumn[0]

\vspace{\baselineskip}
\begin{caveat}[Flattening]
  In deep learning, we often work with matrices, or higher-dimensional tensors.
  We want to use matrix linear algebra expressions to avoid using heavy index notation.
  This can be achieved by flattening all tensors back into vectors and reusing definitions of derivatives from the vector case.
  However, we must be careful when translating the results back to the tensor format, as the translation process depends on the flattening convention.
  Classically, the mathematical derivations prefer a \emph{different} flattening scheme than the one used in deep learning libraries.
  This can cause confusion and bugs.
\end{caveat}

\switchcolumn[1]
\codeblock{basics/flattening}
\switchcolumn[0]

There are many ways to flatten the entries of a tensor into a vector.
The two by far most common conventions are (i) last-varies-fastest ($\rvec$) and (ii) first-varies-fastest ($\cvec$).
Their names are easy to remember from their action on a matrix (see \Cref{ex:flattening}): $\cvec$-flattening concatenates columns into a vector (column flattening); $\rvec$-flattening concatenates rows into a vector (row flattening).

Column-flattening is popular in mathematical presentations, while row-flattening is popular in deep learning libraries, which lay out tensors in row-major format in memory.
To see their differences, we will implement both (\Cref{basics/flattening}).
For arbitrary tensors, we can generalize the matrix flattenings by ordering entries such that either their first index ($\cvec$, \Cref{def:cvec}) or last index ($\rvec$, \Cref{def:rvec}) varies fastest:

\begin{setup}[Rank-$A$ tensor]\label{setup:flattening}
  Let $\tA \in \sR^{N_1 \times \dots \times N_A}$ be a tensor of rank $A$ whose entries are indexed through a tuple $(n_1, \dots, n_A)$ where $n_a \in \{1, \dots, N_a\}$ for $a \in \{1, \dots, A\}$.
  Vectors are rank-1 tensors, and matrices are rank-2 tensors.
\end{setup}
\begin{definition}[$\cvec$, \Cref{basics/flattening}]\label{def:cvec}
  The first-varies-fastest flattening of tensor $\tA$ from \Cref{setup:flattening} is
  \begin{align*}
    \cvec(\tA) =
    \begin{pmatrix}
      \etA_{\colored{1},1,\dots,1}   \\
      \etA_{\colored{2},1,\dots,1}   \\
      \vdots               \\
      \etA_{\colored{N_1},1,\dots,1} \\
      \etA_{\colored[VectorPink]{1},2,\dots,1}   \\
      \vdots               \\
      \etA_{\colored[VectorPink]{N_1},2,\dots,1} \\
      \vdots               \\
      \etA_{N_1,N_2,\dots,N_A}
    \end{pmatrix}
    \in \sR ^{N_1 \cdots N_A}\,.
  \end{align*}
\end{definition}

\begin{definition}[$\rvec$, \Cref{basics/flattening}]\label{def:rvec}
  The last-varies-fastest flattening of tensor $\tA$ from \Cref{setup:flattening} is
  \begin{align*}
    \rvec(\tA) =
    \begin{pmatrix}
      \etA_{1,\dots,1,\colored{1}}   \\
      \etA_{1,\dots,1,\colored{2}}   \\
      \vdots               \\
      \etA_{1,\dots,1,\colored{N_A}} \\
      \etA_{1,\dots,2,\colored[VectorPink]{1}}   \\
      \vdots               \\
      \etA_{1,\dots,2,\colored[VectorPink]{N_A}} \\
      \vdots               \\
      \etA_{N_1,\dots,N_{A-1},N_A}
    \end{pmatrix}
    \in \sR ^{N_A \cdots N_1}\,.
  \end{align*}
\end{definition}

In code, we will sometimes require partial flattening of a subset of contiguous indices, instead of all indices (\eg to turn a tensor into a matrix by first flattening the row indices, followed by flattening the column indices).
The definitions are analogous, but the flattened indices are surrounded by static ones.

\switchcolumn[0]
\subsubsection{Jacobians, JVP, VJPs}
Building up to curvature approximations that tackle the approximation of second-order partial derivatives, we start with first-order derivatives.
These are collected into a matrix called the Jacobian, which depends on the flattening convention.
We can multiply with the Jacobian and its transpose via automatic differentiation, without building up the matrix in memory.
These operations are called Jacobian-vector products (JVPs) and vector-Jacobian products (VJPs), respectively.

Machine learning libraries like JAX and PyTorch offer routines for computing Jacobians, VJPs, and JVPs.
However, their interface is functional.
Here, we provide an alternative implementation that accepts nodes of an evaluated computation graph rather than unevaluated functions as input and will be beneficial for modular implementations of neural networks, as we consider later.
We also provide examples for important Jacobians, namely the output-parameter Jacobian of an affine map, \ie a linear layer.
These Jacobians exhibit a Kronecker structure, which is the foundation for the `K' in KFAC.
We verify this structure numerically and observe how the flattening convention affects it.

\begin{setup}[Vector-to-vector function]\label{setup:vector_to_vector_function}
  Let function $f\colon \sR^A \to \sR^B, \va \mapsto \vb = f(\va)$ denote a vector-to-vector function.
\end{setup}

\begin{definition}[Jacobian of a vector-to-vector function]\label{def:vector_jacobian}
  The Jacobian of a vector-to-vector function $f$ from \Cref{setup:vector_to_vector_function}, $\jac_{\va}\vb \in \sR^{B \times A}$, collects the first-order partial derivatives into a matrix such that
  \begin{align*}
    [\jac_{\va} \vb]_{i,j} = \frac{\partial [f(\va)]_i}{\partial [\va]_j}\,.
  \end{align*}
\end{definition}
\Cref{def:vector_jacobian} is limited to vector-to-vector functions.
The more general Jacobian of a tensor-to-tensor function can be indexed with combined indices from the input and output domains:

\begin{setup}[Tensor-to-tensor function]\label{setup:jacobians}
  Consider a tensor-to-tensor function $f\colon \sR^{A_1 \times \dots \times A_N} \to \sR^{B_1 \times \dots \times B_M}, \tA \mapsto \tB = f(\tA)$ from a rank-$N$ tensor $\tA$ into a rank-$M$ tensor $\tB$.
\end{setup}

\begin{definition}[General Jacobian, \Cref{basics/jacobians}]\label{def:general_jacobian}
  The general Jacobian of $f$ from \Cref{setup:jacobians}, $\tJ_{\tB}\tA$, is a rank-$(M+N)$ tensor that collects the first-order partial derivatives such that
  \begin{align*}
    [\tJ_{\tA}\tB]_{\colored{i_1, \dots, i_M}, \colored[VectorPink]{j_1, \dots, j_N}}
    =
    \frac{\partial [f(\tA)]_{\colored{i_1, \dots, i_M}}}{\partial [\tA]_{\colored[VectorPink]{j_1, \dots, j_N}}}\,.
  \end{align*}
\end{definition}
For $M=N=1$, the general Jacobian reduces to the Jacobian of a vector-to-vector function from \Cref{def:vector_jacobian}.

\switchcolumn[1]*
\codeblock{basics/jacobian_products}
\switchcolumn[0]

\paragraph{Jacobian multiplication.} In practice, this general Jacobian can be prohibitively large and therefore one must almost always work with it in a matrix-free fashion, \ie through VJPs and JVPs.

\begin{definition}[Vector-Jacobian products (VJPs), \Cref{basics/jacobian_products}]\label{def:vjp}
  Given a tensor-to-tensor function $f$ from \Cref{setup:jacobians} and a tensor $\tV \in \sR^{B_1 \times \dots \times B_M}$ in the output domain, the vector-Jacobian product (VJP) $\tU$ of $\tV$ and $\tJ_{\tA}\tB$ lives in $f$'s input domain and follows by contracting the \colored{output indices},
  \begin{align*}
    & [\tU]_{j_1, \dots, j_N}
    \\
    & =
      \colored{\sum_{i_1, \dots, i_M}}
      [\tV]_{\colored{i_1, \dots, i_M}}
      [\tJ_{\tA}\tB]_{\colored{i_1, \dots, i_M}, j_1, \dots, j_N}\,.
  \end{align*}
\end{definition}
For $M=N=1$, $\tV, \tU \to \vv, \vu$ are column vectors, $\tJ_{\tA}\tB \to \jac_{\va}\vb$ is a matrix, and the VJP is $\vu^{\top} = \vv^{\top} (\jac_{\va}\vb)$ or $\vu = (\jac_{\va}\vb)^{\top} \vv$, \ie multiplication with the transpose Jacobian.

VJPs are at the heart of reverse-mode automatic differentiation, aka backpropagation (this is why $\tU$ is often called the \emph{pull-back} or \emph{backpropagation} of $\tV$ through $f$).
Therefore, they are easy to implement with standard functionality (\eg \texttt{autograd.grad} in PyTorch).

The other relevant contraction is between the Jacobian and a vector from the input domain:

\begin{definition}[Jacobian-vector products (JVPs), \Cref{basics/jacobian_products}]\label{def:jvp}
  Given a tensor-to-tensor function $f$ from \Cref{setup:jacobians} and a tensor $\tV \in \sR^{A_1 \times \dots \times A_N}$ in the input domain, the Jacobian-vector product (JVP) $\tU$ between $\tV$ and $\tJ_{\tA}\tB$ lives in $f$'s output domain and follows by contracting the \colored[VectorPink]{input indices},
  \begin{align*}
    & [\tU]_{j_1, \dots, j_M}
    \\
    & =
      \colored[VectorPink]{\sum_{i_1, \dots, i_N}}
      [\tJ_{\tA}\tB]_{j_1, \dots, j_M, \colored[VectorPink]{i_1, \dots, i_N}}
      [\tV]_{\colored[VectorPink]{i_1, \dots, i_N}}\,.
  \end{align*}
\end{definition}
For the vector case, $\tU, \tV, \tJ_{\tA}\tB \to \vu, \vv, \jac_{\va}\vb$, the JVP is $\vu = (\jac_{\va}\vb) \vv$, as suggested by its name.
JVPs are common in forward-mode automatic differentiation ($\tU$ is often called the \emph{push-forward} of $\tV$ through $f$).
Only recently has this mode garnered attention.
The current JVP functionality in ML libraries usually follows a functional API.
To obtain an implementation that accepts variables from a computation graph and is more compatible with the modular approach we chose in this tutorial, we can use a trick that implements a JVP using two VJPs \cite{townsend2017new}.

\switchcolumn[1]*
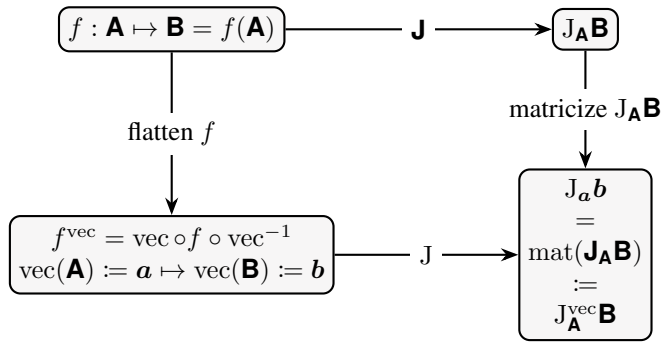
\begin{figure}[!h]
  \centering
  \begin{tikzpicture}[%
    thick,
    box/.style = {rectangle, draw=black, rounded corners, fill=VectorGray!50},%
    ]
    \node[box] (A) at (0,0) {$f: \tA \mapsto \tB = f(\tA)$};
    \node[box] (B) at (5.5,0) {$\jac_{\tA}\tB$};
    \node[box, align=center] (C) at (0,-3) {%
      $f^{\vec} = \vec \circ f \circ \vec^{-1}$\\%
      $\vec(\tA) \coloneq \va \mapsto \vec(\tB) \coloneq \vb$%
    };
    \node[box, align=center] (D) at (5.5,-3) {%
      $\jac_{\va} \vb$\\%
      $=$\\%
      $\mat(\tJ_{\tA}\tB)$\\%
      $\coloneq$\\%
      $\jac^{\vec}_{\tA}\tB$%
    };
    \draw[-Stealth] (A.east) -- node[fill=white] {$\tJ$} (B.west);
    \draw[-Stealth] (A.south) -- node[fill=white] {flatten $f$} (C.north);
    \draw[-Stealth] (C.east) -- node[fill=white] {$\jac$} (D.west);
    \draw[-Stealth] (B.south) -- node[fill=white] {matricize $\jac_{\tA}\tB$} (D.north);
  \end{tikzpicture}
  \caption{\textbf{Flattening and taking the Jacobian commute and lead to the same matricized Jacobian.}
    $\vec$ denotes one of the flattening conventions from \Cref{def:cvec,def:rvec}.
    $\mat$ denotes matricization (two partial flattenings for row and column dimensions, respectively).}\label{fig:commutative-diagram-jacobian}
\end{figure}
\switchcolumn[0]

\paragraph{Matricization.}
Jacobian products are efficient, but somewhat abstract to work with, as we cannot `touch' the full tensor.
Often, we would also like to think about this tensor as a matrix to be able to present derivations in linear algebra notation.

We can reduce the general Jacobian tensor back to the Jacobian matrix in two different ways: We can either (i) directly matricize the tensor, or (ii) `flatten' the function $f \to f^{\vec}$ such that it consumes and produces vectors instead of tensors, then compute its Jacobian.
Both ways and their resulting Jacobian matrices depend on the flattening convention we choose.
The following definitions are consistent in the sense that both of the aforementioned approaches yield the same result, illustrated by the commutative diagram in \cref{fig:commutative-diagram-jacobian}.

\switchcolumn[1]
\codeblock{basics/jacobians}
\switchcolumn[0]

For this tutorial, the two matrices of interest are the $\cvec$- and $\rvec$-Jacobians.
The $\cvec$-Jacobian is used in mathematical derivations in the literature.
The $\rvec$-Jacobian is common in code.

\begin{definition}[$\cvec$-Jacobian, \Cref{basics/jacobians}]\label{def:cvec_jacobian}
  For a tensor-to-tensor function $f$ from \Cref{setup:jacobians}, its $\cvec$-Jacobian $\jac^{\cvec}_{\tA}\tB \in \sR^{(B_1 \cdots B_M) \times (A_1 \cdots A_N)}$ is attained by flattening the input and output tensors with $\cvec$ and applying the Jacobian definition for vectors,
  \begin{align*}
    [\jac^{\cvec}_{\tA}\tB]_{i,j}
    =
    \frac{\partial [\cvec(f(\tA))]_i}{\partial [\cvec(\tA)]_j}\,.
  \end{align*}
\end{definition}

\begin{definition}[$\rvec$-Jacobian, \Cref{basics/jacobians}]\label{def:rvec_jacobian}
  For a tensor-to-tensor function $f$ from \Cref{setup:jacobians}, its $\rvec$-Jacobian $\jac^{\rvec}_{\tA}\tB \in \sR^{(B_M \cdots B_1) \times (A_N \cdots A_1)}$ is attained by flattening the input and output tensors with $\rvec$ and applying the Jacobian definition for vectors,
  \begin{align*}
    [\jac^{\rvec}_{\tA}\tB]_{i,j}
    =
    \frac{\partial [\rvec(f(\tA))]_i}{\partial [\rvec(\tA)]_j}\,.
  \end{align*}
\end{definition}

\paragraph{Example.} The two Jacobians usually differ from each other, albeit in subtle ways.
We highlight their differences on a linear layer, which will be useful later on when we discuss KFAC (\Cref{ex:linear_layer_jacobians}, numerically verified in \Cref{basics/jacobians_linear_layer}).
This example reveals two insights:
\begin{itemize}
\item There is a Kronecker structure in the linear layer's Jacobian \wrt its weight.
  This structure is the foundation for the `K' in KFAC.

\item The order of Kronecker factors is reversed depending on the flattening scheme.
  Therefore, we need to be careful when translating results from one flattening convention to the other.
\end{itemize}

\switchcolumn[1]
\begin{example}[$\cvec$- and $\rvec$-Jacobians of a linear layer \wrt its weights, \Cref{basics/jacobians_linear_layer}]\label{ex:linear_layer_jacobians}
  Consider an affine map with weight matrix $\mW \in \sR^{D_{\text{out}} \times D_{\text{in}}}$, bias vector $\vb \in \sR^{D_{\text{out}}}$, input vector $\vx \in \sR^{D_{\text{in}}}$ and output vector $\vz \in \sR^{D_{\text{out}}}$,
  \begin{align*}
    \vz
    \coloneqq
    \mW \vx + \vb
    =
    \begin{pmatrix}
      \mW & \vb
    \end{pmatrix}
    \begin{pmatrix}
      \vx \\ 1
    \end{pmatrix}
    \coloneqq
    \tilde{\mW}
    \tilde{\vx}\,.
  \end{align*}
  To express this operation as matrix-vector multiplication, we combine weight and bias into a single matrix $\tilde{\mW}$ and augment the input with a one, yielding $\tilde{\vx}$, to account for the bias contribution.

  The linear layer's $\cvec$-Jacobian \wrt the combined weight is
  \begin{align*}
    \jac^{\cvec}_{\tilde{\mW}}\vz
    =
    \tilde{\vx}^{\top}
    \otimes
    \mI_{D_{\text{out}}}\,.
  \end{align*}
  In contrast, the $\rvec$-Jacobian is
  \begin{align*}
    \jac^{\rvec}_{\tilde{\mW}}\vz
    =
    \mI_{D_{\text{out}}}
    \otimes
    \tilde{\vx}^{\top}\,,
  \end{align*}
  see \Cref{basics/jacobians_linear_layer}.
  Note that the order of Kronecker factors is \emph{reversed}, depending on the flattening scheme.
\end{example}
\switchcolumn[0]

\switchcolumn[1]
\codeblock{basics/jacobians_linear_layer}
\switchcolumn[0]

%
%
\begin{comment}
  \begin{example}[$\cvec$- and $\rvec$-weight Jacobians of a linear layer with weight sharing]
    Consider the same affine map from above, but now processing multiple input vectors $\mX = \begin{pmatrix}\vx_1 & \dots & \vx_S\end{pmatrix} \in \sR^{D_{\text{in}}\times S}$, yielding a sequence $\mZ = \begin{pmatrix} \vz_1 & \dots & \vz_S\end{pmatrix} \in \sR^{D_{\text{out}}\times S}$ where each $\vz_s$ is produced like above.
    The parameters are \emph{shared} over all vectors in the input sequence.
    In matrix notation,
    \begin{align*}
      \mZ
      & \coloneqq
        \mW \mX + \vb \vone^{\top}_S
      \\
      & =
        \begin{pmatrix}
          \mW & \vb
        \end{pmatrix}
        \begin{pmatrix}
          \mX \\ \vone^{\top}_S
        \end{pmatrix}
        \coloneqq
        \tilde{\mW}
        \tilde{\mX}\,.
    \end{align*}
    The $\cvec$-Jacobian \wrt the combined weight is
    \begin{align*}
      \jac^{\cvec}_{\tilde{\mW}}\mZ
      =
      \tilde{\mX}^{\top}
      \otimes
      \mI_{D_{\text{out}}}\,.
    \end{align*}
    In contrast, the $\rvec$-Jacobian is
    \begin{align*}
      \jac^{\rvec}_{\tilde{\mW}}\mZ
      =
      \mI_{D_{\text{out}}}
      \otimes
      \tilde{\mX}^{\top}\,.
    \end{align*}
  \end{example}

  \switchcolumn[1]
  \codeblock{basics/jacobians_shared_linear_layer}
\end{comment}

%
%
%
%

\switchcolumn[0]
\subsubsection{Hessians, HVPs}
Now that we have covered first-order derivatives, let's move on to second-order derivatives and develop the necessary concepts to understand KFAC, as well as their implementation.
Second-order derivatives are collected into an object called \emph{the Hessian}.
For our purposes, it will be sufficient to consider the Hessian of functions producing a scalar-valued output.
Let's start with the definition of the Hessian of a vector-to-scalar function.

\begin{setup}[Vector-to-scalar function]\label{setup:vector_to_scalar_function}
  Let $f\colon \sR^A \to \sR, \va \mapsto b = f(\va)$ be a vector-to-scalar function.
\end{setup}

\begin{definition}[Hessian of a vector-to-scalar function]\label{def:vector_hessian}
  The Hessian of a vector-to-scalar function $f$ from \Cref{setup:vector_to_scalar_function} is a matrix $\hess_{\va}b \in \sR^{A \times A}$ collecting the second-order partial derivatives of $f$ into a matrix with
  \begin{align*}
    [\hess_{\va}b]_{i,j}
     & =
    \frac{\partial^2 b}{\partial [\va]_i \partial [\va]_j}\,.
  \end{align*}
\end{definition}
This definition is limited to functions with vector-valued arguments. The extension to tensor-to-scalar functions is straightforward; however, it yields a tensor that is less convenient to work with in terms of notation:

\begin{setup}[Tensor-to-scalar function]\label{setup:hessians}
  Consider a tensor-to-scalar function $f\colon \sR^{A_1 \times \dots \times A_N} \to \sR, \tA \mapsto b = f(\tA)$ from a rank-$N$ tensor $\tA$ into a scalar $b$.
\end{setup}

\begin{definition}[General Hessian of a tensor-to-scalar function, \Cref{basics/hessians}]\label{def:general_hessian}
  The general Hessian of $f$ from \Cref{setup:hessians}, $\tH_{\tA}b \in \sR^{A_1 \times \dots \times A_N \times A_1 \times \dots \times A_N}$, collects the second-order partial derivatives of $f$ into a tensor with
  \begin{align*}
     & [\tH_{\tA}b]_{\colored{i_1, \dots, i_N}, \colored[VectorPink]{j_1, \dots, j_N}}
    \\
     & =
    \frac{\partial^2 b}{\partial [\tA]_{\colored{i_1, \dots, i_N}} \partial [\tA]_{\colored[VectorPink]{j_1, \dots, j_N}}}\,.
  \end{align*}
\end{definition}

\switchcolumn[1]*
\codeblock{basics/hessian_product}
\switchcolumn[0]

\paragraph{Hessian multiplication.}
Just like for Jacobians, the Hessian tensor is usually too large to be stored in memory.
Hence, one usually works with it implicitly through matrix-vector products, which can be done without computing the Hessian:

\begin{definition}[Hessian-vector products (HVPs), \Cref{basics/hessian_product}]\label{def:hvp}
  Given a tensor-to-scalar function $f$ from \Cref{setup:hessians} and a tensor $\tV \in \sR^{A_1 \times \dots \times A_N}$ in the input domain, the Hessian-vector product (HVP) $\tU$ of $\tV$ with $\tH_{\tA}b$ is the result of contraction with one of the Hessian's \colored{input indices},
  \begin{align*}
     & [\tU]_{i_1, \dots, i_N}
    \\
     & =
    \colored{\sum_{j_1, \dots, j_N}}
    [\tH_{\tA}b]_{i_1, \dots, i_N, \colored{j_1, \dots, j_N}} [\tV]_{\colored{j_1, \dots, j_N}}\,.
  \end{align*}
\end{definition}
For the vector case $N=1$, we have $\tV, \tA, \tH_{\tA}b \to \vv, \va, \hess_{\va}b$ and $\tU \to \vu = \hess_{\va} b$ as suggested by the name `Hessian-vector product'.

One way to multiply by the Hessian uses the so-called Pearlmutter trick~\cite{pearlmutter1994fast}.
It relies on the fact that multiplication with higher-order derivatives can be done by nested first-order differentiation.
Hence, multiplication with the Hessian can be done with two VJPs (\Cref{basics/hessian_product}).
In fact, this snippet implements a slightly more general Hessian that can handle differentiating twice \wrt \emph{different}, in contrast to the same, arguments.
It is not essential for understanding KFAC, and we will only use it to visualize curvature matrices in \Cref{subsec:curvature-matrices}.
In the context of KFAC, we do not care about these mixed-layer derivatives.

\switchcolumn[1]
\begin{figure}[!h]
  \centering
  \begin{tikzpicture}[%
      thick,
      box/.style = {rectangle, draw=black, rounded corners, fill=VectorGray!50},%
    ]
    \node[box] (A) at (0,0) {$f: \tA \mapsto b = f(\tA)$};
    \node[box] (B) at (5.5,0) {$\tJ_{\tA}b$};
    \node[box, align=center] (C) at (0,-3) {%
      $f^{\vec} = f \circ \vec^{-1}$\\%
      $\vec(\tA) \coloneq \va \mapsto b$%
    };
    \node[box, align=center] (D) at (5.5,-3) {%
      $\hess_{\va} b$\\%
      $=$\\%
      $\mat(\tH_{\tA}b)$\\%
      $\coloneq$\\%
      $\hess^{\vec}_{\tA}b$%
    };
    \draw[-Stealth] (A.east) -- node[fill=white] {$\tJ$} (B.west);
    \draw[-Stealth] (A.south) -- node[fill=white] {flatten $f$} (C.north);
    \draw[-Stealth] (C.east) -- node[fill=white] {$\jac$} (D.west);
    \draw[-Stealth] (B.south) -- node[fill=white] {matricize $\tH_{\tA}b$} (D.north);
  \end{tikzpicture}
  \caption{\textbf{Flattening and taking the Hessian commute and lead to the same matricized Hessian.}
    $\vec$ denotes one of the flattening conventions from \Cref{def:cvec,def:rvec}.
    $\mat$ denotes matricization and involves two partial flattenings for row and column dimensions.}\label{fig:commutative-diagram-hessian}
\end{figure}
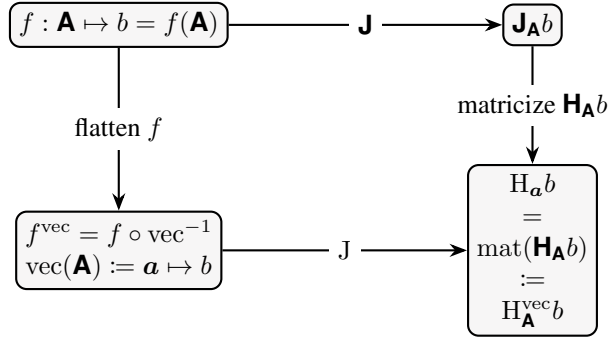
\switchcolumn[0]

\paragraph{Matricization.} For notational convenience, we will also define matricized versions of the general Hessian from \Cref{def:general_hessian}; the $\cvec$-, and $\rvec$-Hessian. Just like for the Jacobians, it does not matter whether we first flatten the function's input space then compute the Hessian, or compute the general Hessian then matricize it (\Cref{fig:commutative-diagram-hessian}).
The following definitions are consistent for both ways.

\switchcolumn[1]
\codeblock{basics/hessians}
\switchcolumn[0]

\begin{definition}[$\cvec$-Hessian, \Cref{basics/hessians}]\label{def:cvec_hessian}
  For a tensor-to-scalar function $f$ from \Cref{setup:hessians}, the $\cvec$-Hessian $\hess_{\tA}^{\cvec}b \in \sR^{(A_N \cdots A_1) \times (A_N \cdots A_1)}$ results from flattening the input tensor with $\cvec$ and applying the Hessian from \Cref{def:vector_hessian},
  \begin{align*}
    [\hess^{\cvec}_{\tA}b]_{i, j}
     & =
    \frac{\partial^2 b}{\partial [\cvec(\tA)]_{i} \partial [\cvec(\tA)]_{j}}\,.
  \end{align*}
\end{definition}

\begin{definition}[$\rvec$-Hessian, \Cref{basics/hessians}]\label{def:rvec_hessian}
  For a tensor-to-scalar function $f$ from \Cref{setup:hessians}, the $\rvec$-Hessian $\hess_{\tA}^{\rvec}b \in \sR^{(A_1 \cdots A_N) \times (A_1 \cdots A_N)}$ results from flattening the input tensor with $\rvec$ and applying the Hessian from \Cref{def:vector_hessian},
  \begin{align*}
    [\hess^{\rvec}_{\tA}b]_{i, j}
     & =
    \frac{\partial^2 b}{\partial [\rvec(\tA)]_{i} \partial [\rvec(\tA)]_{j}}\,.
  \end{align*}
\end{definition}

Whenever we consider vector-to-scalar functions, both Hessians are identical, and we thus suppress the flattening scheme and write $\hess_{\va}b$.

\paragraph{Examples.}
Let's look at important Hessian examples we will return to later.
We also use them to verify our Hessian and HVP implementations.

\switchcolumn[1]
\codeblock{basics/hessian_ce_loss}
\switchcolumn[0]

\begin{example}[Softmax cross-entropy loss Hessian, \Cref{basics/hessian_ce_loss}]\label{ex:hessian-crossentropyloss}
  Consider the softmax cross-entropy loss function between the vector-valued logits $\vf \in \sR^C$ and a class label $y \in \{1, \dots, C\}$ from \Cref{ex:cross_entropy_loss}:
  \begin{align*}
    c(\vf, y)
     & =
    -\log([\vsigma(\vf)]_y)\,.
  \end{align*}
  with $\vsigma(\vf) = \softmax(\vf) \in \sR^C$.
  Its Hessian \wrt $\vf$ is diagonal-minus-rank-one,
  \begin{align*}
    \hess_{\vf} c(\vf, y)
    =
    \diag(\vsigma) - \vsigma \vsigma^\top\,.
  \end{align*}
  See \eg~\citet{dangel2020modular} for a derivation.
\end{example}

\switchcolumn[1]
\codeblock{basics/hessian_mse_loss}

\switchcolumn[0]
\begin{example}[Square loss Hessian, \Cref{basics/hessian_mse_loss}]\label{ex:square_loss_hessian}
  Consider the square loss function between a vector-valued input $\vf \in \sR^C$ and its associated target $\vy \in \sR^C$ from \Cref{ex:square_loss}:
  \begin{align*}
    c(\vf, \vy)
     & =
    \frac{1}{2}\left\lVert
    \vf - \vy
    \right\rVert^2
    \\
     & =
    \frac{1}{2}(\vf - \vy)^{\top} \mI_C (\vf - \vy)\,.
  \end{align*}
  Its Hessian \wrt $\vf$ is the identity,
  \begin{align*}
    \hess_{\vf} c(\vf, \vy)
    =
    \mI_C\,.
  \end{align*}
\end{example}

\switchcolumn[0]
\subsubsection{Partial Linearization, Generalized Gauss-Newtons (GGNs), GGNVPs}\label{sec:partial_linearization}
The last trick for our bag to embark on the KFAC journey is linearization.
It is a useful tool whenever we encounter a composition of functions that we would like to convexify.

\paragraph{Convexification by partial linearization.}
Consider \eg the function $f = g \circ h$ with $f,g,h\colon \sR \to \sR$ for simplicity.
We know that convexity is preserved under function composition, so if both $g$ and $h$ are convex, then $f$ will be convex.
But what if only one of the two composites is convex, let's say $g$, but $h$ is not?
Well, then we can replace $h$ with an approximation $\bar{h}$ that is convex and approximates the original function.
One straightforward approximation is a linear approximation.

\paragraph{Linearization.} Let's say we are interested in only a neighbourhood around $x_0$.
Then, we can obtain a simple, convexified approximation $\bar{f} \approx f$ in that neighbourhood by linearizing $h$ around $x_0$, resulting in a function $\lin_{x_0}(h)$.
This involves a first-order Taylor approximation:

\begin{definition}[Linearization (vector case, \Cref{basics/linearization})]\label{def:vector_linearization}
  Consider a vector-to-vector function $f$ from \Cref{setup:vector_to_vector_function}.
  The linearization of $f$ at an anchor point $\va_0 \in \sR^A$ denoted by $\lin_{\va_0}(f)\colon \sR^A \to \sR^B$ is its first-order Taylor expansion,
  \begin{align*}
    (\lin_{\va_0}(f))(\va) = f(\va_0) + \jac_{\va_0}f(\va_0) (\va - \va_0)
  \end{align*}
  with the Jacobian from \Cref{def:vector_jacobian}.
  Note that $\lin_{\va_0}(f)$ is linear in $\va$ and coincides with the original function at the anchor point, $(\lin_{\va_0}(f))(\va_0) = f(\va_0)$.
  Its Jacobian also coincides with that of the original function at the anchor point.
\end{definition}

The generalization to the tensor case is straightforward.
We just need to replace the Jacobian-vector product with its generalization:

\switchcolumn[1]
\codeblock{basics/linearization}
\switchcolumn[0]

\begin{definition}[Linearization (tensor case, \Cref{basics/linearization})]\label{def:tensor_linearization}
  The linearization of a tensor-to-tensor function from \Cref{setup:jacobians} at an anchor point $\tA_0 \in \sR^{A_1 \times \ldots \times A_N}$, denoted by $\lin_{\tA_0}(f)$ is defined per-entry as
  \begin{align*}
    & \left[
      (\lin_{\tA_0}(f))(\tA)
      \right]_{i_1, \ldots, i_M} = f(\tA_0)
    \\
    & \hspace{2em}+
      \sum_{j_1, \ldots, j_N}
      \left[
      \jac_{\tA_0}f(\tA_0)
      \right]_{i_1, \ldots, i_M, j_1, \ldots, j_N}
    \\
    & \hspace{6.2em}\left[
      \tA - \tA_0
      \right]_{j_1, \ldots, j_N}\,,
  \end{align*}
  with the Jacobian from \Cref{def:general_jacobian}. Note that this is nothing but the function evaluated at the anchor point plus the JVP (\Cref{def:jvp}) with the distance to the anchor.
\end{definition}

\paragraph{GGN \& relevance to deep learning.}
In deep learning, we often face the situation where $f = g \circ h, \tA \mapsto b = f(\tA)$ and $h$ is non-convex while $g$ is convex
(\eg in the empirical risk, where the criterion is typically convex, while the neural net is typically non-convex).
This means the Hessian of $f$ can be indefinite, even though many algorithms require a positive definite approximation to the Hessian.
We can obtain that by considering the partially linearized function $\bar{f} = g \circ \lin_{\tA_0}(h)$, whose Hessian is positive semi-definite.
The Hessian of this partially linearized function is called the generalized Gauss-Newton (GGN) matrix.

Let's stick to our one-dimensional example for a moment, \ie let $f(x) = (g \circ h)(x) \in \sR$.
If we use the chain rule twice, we obtain $f'' = g' h'' g' + h' g''$ for the second derivative. Using our Jacobian and Hessian notation, this translates into
\begin{align*}
  &\hess_x f(x)
  \\
  &=\hess_x (g \circ h)(x)
  \\
  &=\jac_x h(x) \cdot \hess_{h(x)} g(h(x)) \cdot \jac_x h(x)
  \\
  &\phantom{= }+
    \hess_x h(x) \cdot \jac_{h(x)} g(h(x))\,.
\end{align*}
Now, if we take the Hessian of the partially linearized function $\bar{f}(x) = (g \circ \lin_{x_0}(h))(x)$, and use the shorthand $\bar{h} = \lin_{x_0}(h)$ the second term disappears as the linear function's Hessian is zero:
\begin{align*}
  &\hess_x \bar{f}(x)
  \\
  &=\hess_x (g \circ \bar{h})(x)
  \\
  &=\jac_x \bar{h}(x) \cdot \hess_{\bar{h}(x)} g(\bar{h}(x)) \cdot \jac_x \bar{h}(x)
  \\
  &\phantom{= }+
    \underbrace{\hess_x \bar{h}(x)}_{= 0} \cdot \jac_{\bar{h}(x)} g(\bar{h}(x))\,.
\end{align*}
Evaluating both equations at the anchor point (setting $x = x_0$), we see that their first terms coincide (remember from \Cref{def:vector_linearization} that the linearized function's value and Jacobian coincide with those of the original function when evaluated at the anchor point).

\switchcolumn[1]*
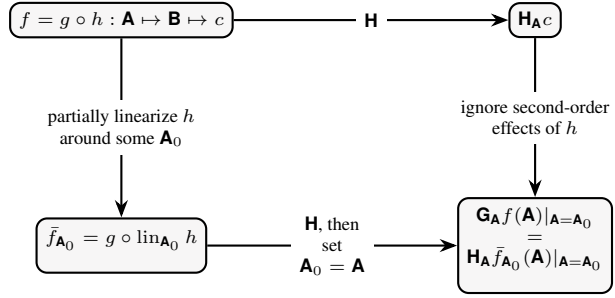
\begin{figure}[!h]
  \centering
  \begin{tikzpicture}[%
    font=\scriptsize,%
    thick,
    box/.style = {rectangle, draw=black, rounded corners, fill=VectorGray!50},%
    ]
    \node[box] (A) at (0,0) {$f = g \circ h: \tA \mapsto \tB \mapsto c$};
    \node[box] (B) at (5.5,0) {$\tH_{\tA}c$};
    \node[box, align=center] (C) at (0,-3) {%
      $\bar{f}_{\tA_0} = g \circ \lin_{\tA_0} h$\\%
    };
    \node[box, align=center] (D) at (5.5,-3) {%
      $\tG_{\tA}f(\tA)|_{\tA = \tA_0}$
      \\
      $=$\\%
      $\tH_{\tA} \bar{f}_{\tA_0}(\tA)|_{\tA = \tA_0}$\\%
    };
    \draw[-Stealth] (A.east) -- node[fill=white] {$\tH$} (B.west);
    \draw[-Stealth] (A.south) -- node[fill=white, align=center] {partially linearize $h$ \\
      around some $\tA_0$} (C.north);
    \draw[-Stealth] (C.east) -- node[fill=white, align=center] {$\tH$, then\\ set\\ $\tA_0 = \tA$} (D.west);
    \draw[-Stealth] (B.south) -- node[fill=white, align=center] {ignore second-order\\ effects of $h$} (D.north);
  \end{tikzpicture}
  \caption{\textbf{Taking the Hessian and partial linearization commute and lead to the same GGN.}
    All operations also commute with flattening the function versus matricizing the Hessian tensor, so we can apply any flattening scheme on top to obtain the $\rvec$- and $\cvec$-GGN matrices.}\label{fig:commutative-diagram-ggn}
\end{figure}
\switchcolumn[0]

Hence, our one-dimensional example's GGN is
\begin{align*}
  &\ggn_{x_0} f(x_0)
  \\
  &=\jac_{x_0} h(x_0) \cdot \hess_{h(x_0)} g(h(x_0)) \cdot \jac_{x_0} h(x_0)\,.
\end{align*}
To summarize, we obtain the GGN either by (i) partially linearizing, then taking the Hessian, or (ii) taking the Hessian and discarding second-order derivatives of the non-convex composite (\ie partially linearizing).
This commutative relationship is visualized in \Cref{fig:commutative-diagram-ggn}.
Our next steps are to generalize the GGN to vector-valued functions that produce scalars, illustrate it on a toy problem, and then develop the fully general formulation for tensor-valued functions along with the functionality to work with it in code.

\switchcolumn[1]*
\begin{figure}[H]
  \centering
  \includegraphics[width=\linewidth]{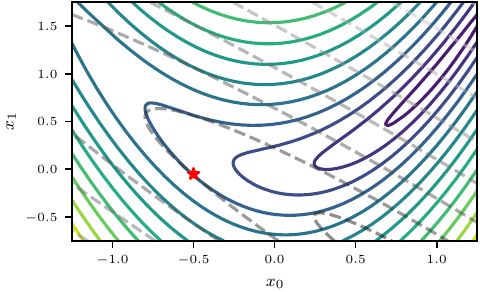}
  \caption{\textbf{The 2d Rosenbrock function} $f_{\alpha=10}$ (solid contour lines) and its convex approximation $\bar{f}_{\alpha=10} = g \circ \bar{h}_{\alpha=10}$ (dashed contour lines) around an anchor $\hat{\vx}$ (star) from partial linearization.
    The convex approximation is a quadratic form with the GGN.}\label{fig:2d-rosenbrock}
\end{figure}
\switchcolumn[0]

\paragraph{Vector case with basic example.}
We already introduced notation for the GGN.
Let's define it for the vector case and apply it to an example.

\switchcolumn[1]
\codeblock{basics/ggn_rosenbrock}
\switchcolumn[0]

\begin{setup}[Composite vector-to-vector-to-scalar function]\label{setup:composite_vector_to_vector_to_scalar_function}
  Let
  \begin{align*}
    f\colon \sR^{A} & \to \sR
    \\
    \va        & \mapsto c = f(\va)
  \end{align*}
  be the composite of a vector-to-vector function $h$ and a vector-to-scalar function $g$, that is
  \begin{align*}
    f = g \circ h\colon \sR^A & \to \sR^B \to \sR
    \\
    \va                  & \mapsto \vb = h(\va) \mapsto c = g(\vb)\,.
  \end{align*}
\end{setup}

\begin{definition}[GGN matrix (vector case)]\label{def:vector_ggn}
  The GGN matrix of a vector-to-vector-to-scalar function $f$ from \Cref{setup:composite_vector_to_vector_to_scalar_function} is
  \begin{align*}
    \ggn_{\va} f(\va)
    =
    (\jac_{\va} \vb)^{\top}
    (\hess_{\vb} c)
    (\jac_{\va} \vb) \in \sR^{A \times A}\,,
  \end{align*}
  \ie the second composite's Hessian, pulled back with the first composite's Jacobian.
\end{definition}

One obtains this form by following through the same steps as above, but for a vector-to-vector-to-scalar instead of a purely scalar function composition. \Cref{ex:ggn-rosenbrock} presents fully-worked-out expressions of the Rosenbrock function's GGN.

\begin{example}[GGN for the Rosenbrock function, \Cref{basics/ggn_rosenbrock}]\label{ex:ggn-rosenbrock}
  Consider the 2d Rosenbrock function $f_{\alpha}\colon \sR^2 \to \sR$ with
  \begin{align*}
    f_{\alpha}(\vx)
    =
    (1 - x_1)^2 + \alpha (x_2 - x_1^2)^2\,.
  \end{align*}
  with some $\alpha > 0$.

  \paragraph{Composition.}
  Following \citet{brunet2010basics}, we express the Rosenbrock function as vector-to-vector-to-scalar function
  \begin{align*}
    f_{\alpha} &= g \circ h_{\alpha}\colon \sR^2 \to \sR^2 \to \sR
                 \shortintertext{with}
                 h_{\alpha}(\vx) &= \begin{pmatrix}
                   1 - x_1 \\
                   \sqrt{\alpha} (x_2 - x_1^2)
                 \end{pmatrix}
                                   \shortintertext{and convex}
                                   g(\vh) &= \vh^\top \vh\,,
  \end{align*}
  namely, the square loss.

  \paragraph{Linearization.}
  The Rosenbrock function is non-convex (see \Cref{fig:2d-rosenbrock} for contour lines) because $h_{\alpha}$ is non-convex.
  We want to convexify it by partial linearization.
  Linearizing $h_{\alpha}$ \wrt $\vx$ around an anchor $\hat{\vx}$ gives
  \begin{align*}
    \bar{h}_{\alpha}(\vx)
    &=
      h_{\alpha}(\hat{\vx}) + (\jac_{\hat{\vx}}h_{\alpha}(\hat{\vx})) (\vx - \hat{\vx})
      \shortintertext{with the Jacobian}
      \jac_{\vx}h_{\alpha}(\vx)
    &=
      \begin{pmatrix}
        -1                   & 0             \\
        -2 \sqrt{\alpha} x_1 & \sqrt{\alpha}
      \end{pmatrix}\,.
  \end{align*}
  Inserting all expressions, we have
  \begin{align*}
    \bar{h}_{\alpha}(\vx)
    =
    \begin{pmatrix}
      1 - \hat{x}_1 \\
      \sqrt{\alpha} (\hat{x}_2 - \hat{x}_1^2)
    \end{pmatrix}
    \\
    +
    \begin{pmatrix}
      \hat{x}_1 - x_1 \\
      2 \sqrt{\alpha} \hat{x}_1 (\hat{x}_1 - x_1) + \sqrt{\alpha} (x_2 - \hat{x}_2)
    \end{pmatrix}
    \\
    =
    \begin{pmatrix}
      1 - x_1
      \\
      2 \sqrt{\alpha} \hat{x}_1 \left( \frac{1}{2} \hat{x}_1 - x_1 \right) +  \sqrt{\alpha} x_2
    \end{pmatrix}
  \end{align*}
  Note that this function is indeed linear in $x_{1,2}$.

  \paragraph{Partial linearization.}
  With $\bar{h}_{\alpha}(\vx)$, we can evaluate the partially linearized Rosenbrock function $\bar{f}_{\alpha} = g \circ \bar{h}_{\alpha}$ as
  \begin{align*}
    \bar{f}_{\alpha}(\vx)
    &=
      (\bar{h}_{\alpha}(\vx))^{\top} \bar{h}_{\alpha}(\vx)
    \\
    &=
      (1 - x_1)^2
    \\
    &\phantom{= }+
      \alpha \left[ 2 \hat{x}_1 \left(\frac{1}{2} \hat{x}_1 - x_1 \right) +  x_2 \right]^2
    \\
    &= 1 - 2x_1 + x_1^2 + \alpha [ 2 \hat{x}_1^4 - 4 \hat{x}_1^3 x_1
    \\
    &\phantom{= }+ 4 \hat{x}_1^2 x_1^2 - 2 \hat{x}_1^2 x_2 - 4 \hat{x}_1 x_1 x_2 + x_2^2 ]\,.
  \end{align*}

  \paragraph{GGN via linearize-then-differentiate.}
  The Rosenbrock's GGN is the Hessian of its partial linearization, evaluated at $\vx = \hat{\vx}$
  \begin{align*}
    \ggn_{\hat{\vx}} f_{\alpha}(\hat{\vx})
    &=
      \hess_{\vx} \bar{f}_{\alpha}(\vx)|_{\vx=\hat{\vx}}
    \\
    &=
      2
      \begin{pmatrix}
        1 + 4 \alpha \hat{x}_1^2 & - 2\alpha \hat{x}_1
        \\
        - 2\alpha \hat{x}_1 & \alpha
      \end{pmatrix}\,.
  \end{align*}

  \paragraph{GGN via differentiate-then-linearize.}
  Let's double-check our result by evaluating the GGN in \cref{def:vector_ggn}, which follows by differentiating twice, then neglecting non-linearities of $h_{\alpha}$,
  \begin{align*}
    \ggn_{\hat{\vx}} f_{\alpha}(\hat{\vx})
    =&
       (\jac_{\vx} h_{\alpha}(\vx) )^{\top}|_{\vx = \hat{\vx}}
    \\
     &\hess_{h_{\alpha}(\vx)} g(h_{\alpha}(\vx)) |_{h_{\alpha}(\vx) = h_{\alpha}(\hat{\vx})}
    \\
     &\jac_{\vx} h_{\alpha}(\vx) |_{\vx = \hat{\vx}}\,.
       \shortintertext{Using the Rosenbrock's Jacobian from above, and the square loss Hessian from \Cref{ex:square_loss_hessian} which is proportional to the identity, yields}
       =&
          2
          (\jac_{\vx} h_{\alpha}(\vx) )^{\top}|_{\vx = \hat{\vx}}
    \\
     &
       \jac_{\vx} h_{\alpha}(\vx) |_{\vx = \hat{\vx}}
    \\
    =&
       2
       \begin{pmatrix}
         -1                   & -2 \sqrt{\alpha} x_1 \\
         0 & \sqrt{\alpha}
       \end{pmatrix}
    \\
     &
       \begin{pmatrix}
         -1                   & 0             \\
         -2 \sqrt{\alpha} x_1 & \sqrt{\alpha}
       \end{pmatrix}
    \\
    =&
       2
       \begin{pmatrix}
         1 + 4 \alpha \hat{x}_1^2 & - 2\alpha \hat{x}_1
         \\
         - 2\alpha \hat{x}_1 & \alpha
       \end{pmatrix}\,.
  \end{align*}
  This is the same expression we obtained with the linearize-then-differentiate approach.
 We also see that the GGN is positive semi-definite, as it is a scaled outer product of Jacobians.

 \paragraph{Comparison of Hessian and GGN.}
 As expected, the Rosenbrock function's Hessian, evaluated at $\hat{\vx}$, is different from the GGN:
  \begin{align*}
    &\hess_{\vx} f_{\alpha}(\vx)|_{\vx = \hat{\vx}}
      \\
    &=
    2
    \begin{pmatrix}
      1 + 6 \alpha \hat{x}_1^2 - 2 \alpha \hat{x}_2 & -2 \alpha \hat{x}_1 \\
      -2 \alpha \hat{x}_1                      & \alpha
    \end{pmatrix}\,.
  \end{align*}
\end{example}

\paragraph{Generalization to tensor case.} The above example illustrated the GGN for a vector-to-vector-to-scalar function.
We will spend the remaining part of this section generalizing the GGN to tensor-to-tensor-to-scalar functions.
Note that we already generalized all ingredients (linearization from \Cref{def:tensor_linearization}, the Jacobian tensor from \Cref{def:general_jacobian}, and the Hessian tensor from \Cref{def:general_hessian}) to that scenario.
All we have to do is put them together.
After that, we will see a tensor case example of the GGN that will be extremely relevant for our KFAC journey.

\switchcolumn[1]
\codeblock{basics/ggns}
\switchcolumn[0]

\begin{setup}[Composite tensor-to-tensor-to-scalar function]\label{setup:composite_tensor_to_tensor_to_scalar_function}
  Let the tensor-to-scalar map
  \begin{align*}
    f\colon \sR^{A_1 \times \ldots \times A_N} & \to \sR
    \\
    \tA                                   & \mapsto c = f(\tA)
  \end{align*}
  be the composite of a tensor-to-tensor function $h$ and a tensor-to-scalar function $g$, that is
  \begin{align*}
    f = g \circ h\colon & \sR^{A_1 \times \ldots \times A_N} \to \sR^{B_1 \times \ldots \times B_M}  \to \sR\!\!
    \\
                   & \tA \mapsto \tB = h(\tA) \mapsto c = g(\tB)\,.
  \end{align*}
\end{setup}

\begin{definition}[Generalized Gauss-Newton (GGN) tensor (\Cref{basics/ggns})]\label{def:general_ggn}%
  The GGN tensor of a tensor-to-tensor-to-scalar map $f$ from \Cref{setup:composite_tensor_to_tensor_to_scalar_function}, $\tG_{\tA} f(\tA) \in \sR^{A_1 \times \ldots \times A_N \times A_1 \times \ldots \times A_N}$ is the Hessian of the partially linearized function $\bar{f} = g \circ \lin_{\tA_0}(h)$, evaluated at the anchor point $\tA = \tA_0$.
  \begin{align*}
    \tG_{\tA} f(\tA)
    =
    \tH_{\tA} \bar{f}(\tA)|_{\tA = \tA_0}\,.
  \end{align*}
  where $\bar{f} = g \circ \lin_{\tA_0}(h)$.

  Similar to the vector case (\Cref{def:vector_ggn}), we can express this tensor as a contraction between the Jacobians and Hessians of the two composites, \ie
  \begin{align*}
    [\tG_{\tA} f(\tA)]_{i_1, \ldots, i_N, j_1, \ldots, j_N}
    \\
    =
    \textcolor{VectorPink}{\sum_{k_1, \dots, k_M}}
    \textcolor{VectorBlue}{\sum_{l_1, \dots, l_M}}
    & [\jac_{\tA} \tB]_{\textcolor{VectorPink}{k_1, \dots, k_M}, i_1, \dots, i_N} \hspace{-1.3ex}
    \\
    & [\hess_{\tB} c]_{\textcolor{VectorPink}{k_{1}, \ldots, k_{M}}, \textcolor{VectorBlue}{l_{1}, \ldots, l_{M}}} \hspace{-1.3ex}
    \\
    & [\jac_{\tA} \tB]_{\textcolor{VectorBlue}{l_1, \ldots, l_M}, j_1, \ldots, j_N}\,. \hspace{-1.3ex}
  \end{align*}
\end{definition}

\paragraph{GGN multiplication.}
This expression is daunting, so we will convert things back to matrix notation soon.
Before doing so, let's introduce multiplication with the GGN to work with it numerically:

\switchcolumn[1]
\codeblock{basics/ggn_product}
\switchcolumn[0]

\begin{definition}[GGN-vector-product (GGNVP, \Cref{basics/ggn_product})]\label{def:ggnvp}%
  Consider the GGN tensor of a tensor-to-tensor-to-scalar map $f$ from \Cref{setup:composite_tensor_to_tensor_to_scalar_function,def:general_ggn}.
  The GGN-vector-product (GGNVP) $\tU \in \sR^{A_1 \times \ldots \times A_N}$ with a tensor $\tV \in \sR^{A_1 \times \ldots \times A_N}$ from the input domain is
  \begin{align*}
    & [\tU]_{i_1, \dots, i_N}
    \\
    & =
      \colored{\sum_{j_1, \dots, j_N}}
      [\tG_{\tA} f(\tA)]_{i_1, \dots, i_N, \colored{j_1, \dots, j_N}}
      [\tV]_{\colored{j_1, \dots, j_N}}\,,
  \end{align*}
  and decomposes into a JVP, HVP, and VJP when applying the composition from \Cref{def:general_ggn}.
\end{definition}
This is easiest to see for the vector-to-vector-to-scalar case where $\tA, \tB, \tU, \tV \to \va, \vb, \vu, \vv$ and the GGNVP becomes $\vu = (\jac_{\va}\vb)^{\top} (\hess_{\vb} c) (\jac_{\va} \vb) \vv$, which can be written as a matrix chain and computed without explicitly building up any of the matrices in memory~\cite{schraudolph2002fast}.

\paragraph{Matricization.} As for the Hessian and Jacobian, we can flatten both composite functions before applying the partial linearization and taking the Hessian to obtain the GGN.
This is equivalent to matricizing the GGN tensor from \Cref{def:general_ggn}.
And it is also equivalent to matricizing the Hessians and Jacobians in the general definition.

\begin{definition}[$\cvec$- and $\rvec$-GGN matrices, \Cref{basics/ggns}]\label{def:vec_ggns}
  For a tensor-to-tensor-to-scalar function $f$ from \Cref{setup:composite_tensor_to_tensor_to_scalar_function}, we define the $\cvec$- and $\rvec$-GGN matrices by flattening the composite functions before applying the partial linearization and taking the Hessian. This yields the flattened GGN matrices $\ggn^{\vec}_{\tA}f(\tA) \in \sR^{(A_1 \cdots A_N) \times (A_1 \cdots A_N)}$ where $\vec \in \{\cvec, \rvec\}$ which can be written as matrix chain
  \begin{align*}
    \ggn^{\vec}_{\tA} f(\tA)
    =
    (\jac^{\vec}_{\tA} \tB)^{\top}
    (\hess^{\vec}_{\tB} c)
    (\jac^{\vec}_{\tA} \tB)\,.
  \end{align*}
  using the previously defined flattened Hessians (\Cref{def:cvec_hessian,def:cvec_hessian}) and Jacobians (\Cref{def:cvec_jacobian,def:rvec_jacobian}).
\end{definition}

\switchcolumn[1]
\begin{example}[GGN of linear regression (\Cref{basics/ggns_linear_regression})]\label{ex:ggn-linear-regression}
  Consider the least squares objective
  \begin{align*}
    \gL(\mW) = \sum_n \ell_n(\mW)
  \end{align*}
  where $\ell_n = c_n \circ f_n$ is the composition of a linear classifier and square loss
  on a data point labeled $n$, that is
  $f_n(\mW) = \mW \vx_n$ and $c_n(\vz) = \frac{1}{2}\left\lVert \vz - \vy_n \right\rVert^2_2$.

  Using the shorthands $\vf_n \coloneq f_n(\mW)$ and $c_n \coloneq c_n(\vf_n)$, the matrix GGNs are
  \begin{align*}
    \ggn_{\mW}^{\vec}\gL(\mW)
    \coloneq
    \sum_n
    \ggn_{\mW}^{\vec} \ell_n(\mW)
    \\
    =
    \sum_n
    (\jac^{\vec}_{\mW} \vf_n)^{\top}
    (\hess^{\vec}_{\vf_n} c_n)
    (\jac^{\vec}_{\mW} \vf_n)\,.
  \end{align*}
  We can use the results from previous examples, specifically the Jacobian of an affine map from \Cref{ex:linear_layer_jacobians}, and the square loss Hessian from \Cref{ex:square_loss_hessian}, to obtain
  \begin{align*}
    \ggn_{\mW}^{\cvec}\gL(\mW)
    & =
      \left(
      \sum_n \vx_n \vx_n^{\top}
      \right)
      \otimes \mI\,,
    \\
    \ggn_{\mW}^{\rvec}\gL(\mW)
    & =
      \mI
      \otimes
      \left(
      \sum_n \vx_n \vx_n^{\top}
      \right)\,.
  \end{align*}
\end{example}
\switchcolumn[0]

\switchcolumn[1]
\codeblock{basics/ggns_linear_regression}
\switchcolumn[0]

\paragraph{Example.} Again, it is important to emphasize that the matrix GGN depends on the flattening scheme.
To emphasize this point, we conclude this section by studying the GGN of linear regression (\Cref{ex:ggn-linear-regression}).
Two interesting observations about this example are:
\begin{itemize}
\item Both GGNs are a Kronecker product.
  This already hints at using a Kronecker product to approximate the exact GGN, which is what KFAC aims to do.

\item The order of factors depends on the flattening scheme we are using.
  Therefore, we always need to remind ourselves of the chosen scheme when working with the GGN.
\end{itemize}

\switchcolumn[0]
\subsection{Curvature Matrices in Deep Learning}\label{subsec:curvature-matrices}

We introduced Jacobians, Hessians, partial linearizations, and the resulting generalized Gauss-Newton (GGN) in the language of automatic differentiation for arbitrary functions.
Let's switch gears and apply these concepts to deep learning.

\switchcolumn[1]*
\begin{figure}[!h]
  \centering
  \begin{minipage}[t]{0.495\linewidth}
    \centering
    $\cvec$\vspace{1ex}
    \includegraphics[width=\linewidth]{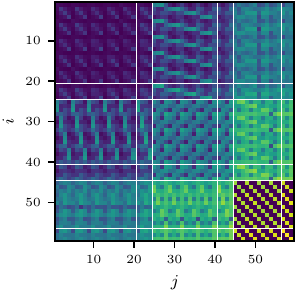}
  \end{minipage}
  \hfill
  \begin{minipage}[t]{0.495\linewidth}
    \centering
    $\rvec$\vspace{1ex}
    \includegraphics[width=\linewidth]{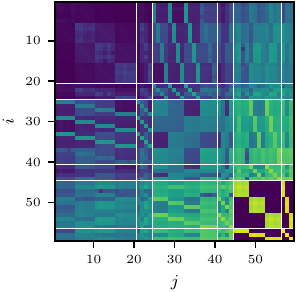}
  \end{minipage}
  \\
  \begin{minipage}[t]{0.495\linewidth}
    \centering
    \includegraphics[width=\linewidth]{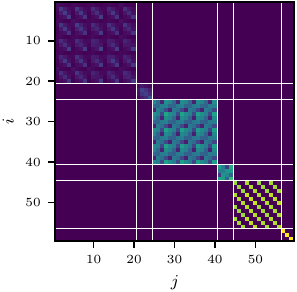}
  \end{minipage}
  \hfill
  \begin{minipage}[t]{0.495\linewidth}
    \centering
    \includegraphics[width=\linewidth]{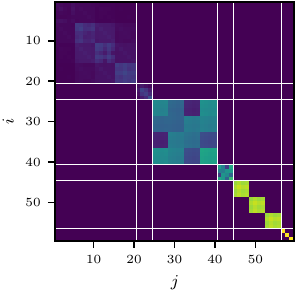}
  \end{minipage}
  \caption{\textbf{The Hessian of a neural net exhibits block structure.}
    We visualize the Hessian and its block-diagonal approximation using different flattening schemes.
    We use synthetic data ($N=100$) on an MLP with three fully-connected layers (5-4-4-3) and ReLU activations and square loss.
    Blocks correspond to second-order derivatives \wrt weights and biases of the different layers, yielding a $6 \times 6$ block structure.
    Hessian blocks are visually highlighted with white lines.
    The left column uses $\cvec$-flattening, the right column $\rvec$-flattening.
    Plots produced with \repofile{plots/synthetic_hessian}.
  }\label{fig:hessian-block-structure}
\end{figure}
\switchcolumn[0]

\paragraph{Parameter list/tuple format.} One new aspect we have to deal with is that ML libraries like PyTorch represent parameters as lists/tuples of variables, each of which is associated with a different layer.
This layer structure in the parameters gives rise to a block structure in the Jacobians, Hessians (see \Cref{fig:hessian-block-structure} for an example), and GGNs.
In the context of KFAC, we will focus on specific blocks.

Typically, a neural net $f(\vx, \vtheta)\colon \gX \times \Theta \to \gF$ has parameters in list/tuple-format,
\begin{align*}
  \vtheta = (\vtheta^{(1)}, \vtheta^{(2)}, \ldots, \vtheta^{(L)}),
\end{align*}
where each $\vtheta^{(l)}$ is an arbitrary parameter tensor.
To be able to use matrix expressions, we will often consider the concatenation of flattened parameters,
\begin{align*}
  \vec(\vtheta)
  =
  \begin{pmatrix}
    \vec(\vtheta^{(1)}) \\
    \vec(\vtheta^{(2)}) \\
    \vdots              \\
    \vec(\vtheta^{(L)})
  \end{pmatrix}
  \in \sR^D
  \,,
\end{align*}
with $\vec \in \{ \rvec, \cvec \}$ one of the previously described flattening operations (\Cref{def:cvec,def:rvec}).
This convention generalizes $\vec$ to handle parameters in list/tuple format by applying the original definition to each element and concatenating the results.
In code, we will still work with the list/tuple format.

\paragraph{Recap: Empirical risk and shorthands.}
We consider a data set $\sD = \{(\vx_n, \vy_n) \in \gX \times \gY \mid n = 1, \dots, N \}$ containing $N$ independent and identically distributed (\iid) samples.
The inputs are processed by a neural net $f\colon \gX \times \Theta \to \gF$ and their predictions are scored with a criterion function $c\colon \gF \times \gY \to \sR$, like square loss (\Cref{ex:square_loss}) or softmax cross-entropy loss (\Cref{ex:cross_entropy_loss}).
For each datum $n$, we define its prediction function $\vf_n\colon \Theta \to \gF, \vtheta \mapsto \vf_n(\vtheta) = f(\vx_n, \vtheta)$ and its criterion function $c_n\colon \gF \to \sR, \vf \mapsto c(\vf, \vy_n)$.
Combining both, we obtain the per-datum loss function (\wrt the net's parameters),
\begin{align*}
  (\ell_n\colon \Theta \to \sR) = (c_n \circ \vf_n\colon \Theta \to \gF \to \sR)
\end{align*}
We will often use the shorthands $c_n, \vf_n, \ell_n$ for the function values (rather than functions) of the per-datum criteria, predictions, and losses.

We accumulate the per-datum losses $\{\ell_1, \dots, \ell_N\}$ into a single scalar, which yields the empirical risk
\begin{align*}
  \gL_{\sD}\colon \Theta \to \sR,
  \qquad
  \vtheta \mapsto \gL_{\sD}(\vtheta) = R \sum_{n=1}^N \ell_n(\vtheta)\,,
\end{align*}
where $R$ is the reduction factor (see \Cref{eq:empirical_risk} for details).
$\sD$ can be any collection of data points, \eg the full data set or a mini-batch.

\subsubsection{The Hessian}\label{sec:basics_dl_hessian}
\switchcolumn[1]*
\begin{figure}[!h]
  \centering
  \begin{minipage}[t]{0.495\linewidth}
    \centering
    $\cvec$\vspace{1ex}
    \includegraphics[width=\linewidth]{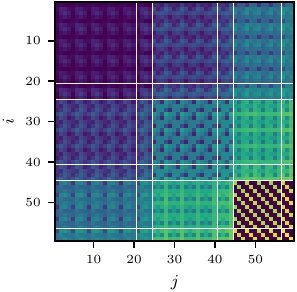}
  \end{minipage}
  \hfill
  \begin{minipage}[t]{0.495\linewidth}
    \centering
    $\rvec$\vspace{1ex}
    \includegraphics[width=\linewidth]{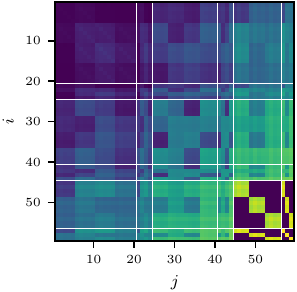}
  \end{minipage}
  \\
  \begin{minipage}[t]{0.495\linewidth}
    \centering
    \includegraphics[width=\linewidth]{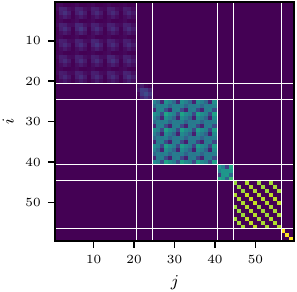}
  \end{minipage}
  \hfill
  \begin{minipage}[t]{0.495\linewidth}
    \centering
    \includegraphics[width=\linewidth]{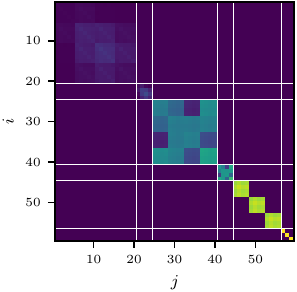}
  \end{minipage}
  \caption{\textbf{The GGN of a neural net exhibits block structure.}
    We visualize the GGN and its block-diagonal approximation using different flattening schemes.
    We use synthetic data ($N=100$) on an MLP with three fully-connected layers (5-4-4-3) and ReLU activations and square loss.
    The GGN blocks are visually highlighted with white lines.
    The left column uses $\cvec$-flattening, the right column $\rvec$-flattening.
    Plots produced with \repofile{plots/synthetic_fisher}
    (this is not a typo; the Fisher and GGN are closely related as we will see in \Cref{subsec:connection-ggn-fisher}).
  }\label{fig:ggn-block-structure}
\end{figure}

\switchcolumn[0]
Consider the empirical risk from above.
The Hessian matrix under a flattening scheme $\vec$ is
\begin{align*}
  \hess_{\vtheta}^{\vec} \gL_{\sD}(\vtheta)
  =
  R
  \sum_{n=1}^N
  \hess_{\vtheta}^{\vec} \ell_n(\vtheta) \in \sR^{D \times D}\,.
\end{align*}
It contains the second-order partial derivatives of $\gL_{\sD}(\vtheta)$ (\Cref{def:vector_hessian}), \ie
\begin{align*}
  [\hess_{\vtheta}^{\vec} \gL_{\sD}(\vtheta)]_{\colored{i},\colored[VectorPink]{j}}
  =
  \frac{\partial^2 \gL_{\sD}(\vtheta)}{\colored{\partial [\vec(\vtheta)]_i} \colored[VectorPink]{\partial [\vec(\vtheta)]_j}}\,.
\end{align*}

\paragraph{Block structure.} Due to the tuple/list structure of $\vtheta$, the Hessian has a block structure (\Cref{fig:hessian-block-structure}).
A block corresponding to parameters $\vtheta^{(k)}, \vtheta^{(l)}$ is
\begin{align*}
  [\hess_{\colored{\vtheta^{(k)}}, \colored[VectorPink]{\vtheta^{(l)}}}^{\vec} \gL_{\sD}]_{i,j}
  =
  \frac{\partial^2 \gL_{\sD}}{\partial [\colored{\vec\left(\vtheta^{(k)}\right)}]_i \partial [\colored[VectorPink]{\vec\left(\vtheta^{(l)}\right)}]_j}\,
\end{align*}
and we usually suppress repeated differentiation arguments, $\hess_{\vtheta^{(k)}}^{\vec} \coloneqq \hess_{\vtheta^{(k)}, \vtheta^{(k)}}^{\vec}$.
With the shorthand $\hess_{k,l}^{\vec} \coloneq \hess_{\vtheta^{(k)}, \vtheta^{(l)}}^{\vec}$,
the Hessian can be written in block form as
\begin{align*}
  \hess_{\vtheta}^{\vec} \gL
  =
  \begin{pmatrix}
    \hess_1^{\vec} \gL
    &
      \hess_{1, 2}^{\vec} \gL
    &
      \cdots
    &
      \hess_{1, L}^{\vec} \gL
    \\
    \hess_{2, 1}^{\vec} \gL
    &
      \hess_2^{\vec} \gL
    &
      \cdots
    &
      \hess_{2, L}^{\vec} \gL
    \\
    \vdots & \cdots & \ddots & \vdots
    \\
    \hess_{L, 1}^{\vec} \gL
    &
      \hess_{L, 2}^{\vec} \gL
    &
      \cdots
    &
      \hess_L^{\vec} \gL
  \end{pmatrix}\,.
\end{align*}
For KFAC, we will only need the block diagonal approximation of this matrix,
\begin{align*}
  \tilde{\hess}_{\vtheta}^{\vec} \gL
  =
  \begin{pmatrix}
    \hess_1^{\vec} \gL & \vzero & \cdots & \vzero
    \\
    \vzero & \hess_2^{\vec} \gL & \ddots & \vdots
    \\
    \vdots & \ddots & \ddots & \vzero
    \\
    \vzero & \cdots & \vzero & \hess_L^{\vec} \gL
  \end{pmatrix}\,.
\end{align*}
\ie, individual blocks $\{ \hess_{\vtheta^{(k)}}^{\vec} \gL_{\sD}(\vtheta)\}_{k=1}^L$.

\subsubsection{The Generalized Gauss-Newton}\label{subsec:curvature-ggn}
\switchcolumn[1]*
\switchcolumn[0]

\paragraph{Partially linearized empirical risk.}
As previously described in \Cref{sec:partial_linearization}, the GGN matrix arises from the partial linearization of a composite function.
We will now apply this to the per-datum loss functions $\ell_n = c_n \circ \vf_n$ in the empirical risk.
Specifically, we will linearize the neural net predictions $\vf_n \to \bar{\vf}_n = \lin_{\vtheta_0}(\vf_n)$, then form the partially-linearized per-sample losses $\bar{\ell}_n = c_n \circ \bar{\vf}_n$.
This gives rise to the partially linearized empirical risk $\bar{\gL}_{\sD}(\vtheta) = R \sum_{n=1}^{N} \bar{\ell}_n(\vtheta)$.
Just like in \Cref{sec:partial_linearization}, the empirical risk's $\vec$-GGN matrix is the Hessian of the partially-linearized empirical risk, evaluated at the anchor,
\begin{align*}
  \ggn^{\vec}_{\vtheta} \gL_{\sD}(\vtheta)
  &=
    \left.\hess_{\vtheta}^{\vec} \bar{\gL}_{\sD}(\vtheta)\right|_{\vtheta_0 = \vtheta} \in \sR^{D \times D}
  \\
  &=
    R \sum_{n=1}^N
    \left.\hess_{\vtheta}^{\vec} \bar{\ell}_n(\vtheta)\right|_{\vtheta_0 = \vtheta},
    \shortintertext{where $\vec$ is one of the flattening operations.
    We can also express the GGN through Jacobians,}
  &=
    R \sum_{n=1}^N
    [\jac_{\vtheta}^{\vec} \vf_n]^{\top}
    \hess^{\vec}_{\vf_n} c_n
    [\jac_{\vtheta}^{\vec} \vf_n]\,.
\end{align*}

\paragraph{Block structure.}
Just like the Hessian, the GGN has a block structure (\Cref{fig:ggn-block-structure}).
Abbreviating
\begin{align*}
  \ggn_{k,l}^{\vec}\gL
  &\coloneq \ggn^{\vec}_{\vtheta^{(k)}, \vtheta^{(l)}} \gL_{\sD}(\vtheta)
  \\
  &= \hess^{\vec}_{\vtheta^{(k)}, \vtheta^{(l)}} \bar{\gL}_{\sD}(\vtheta)
  \\
  &= R \sum_{n=1}^N [\jac_{\vtheta^{(k)}}^{\vec} \vf_n]^{\top} \hess^{\vec}_{\vf_n} c_n [\jac_{\vtheta^{(l)}}^{\vec} \vf_n]
\end{align*}
and $\ggn_{k}^{\vec}\gL \coloneq \ggn^{\vec}_{\vtheta^{(k)}} \gL_{\sD}$, the GGN's block form is
\begin{align*}
  \ggn_{\vtheta}^{\vec} \gL
  =
  \begin{pmatrix}
    \ggn_1^{\vec} \gL
    &
      \ggn_{1, 2}^{\vec} \gL
    &
      \cdots
    &
      \ggn_{1, L}^{\vec} \gL
    \\
    \ggn_{2, 1}^{\vec} \gL
    &
      \ggn_2^{\vec} \gL
    &
      \cdots
    &
      \ggn_{2, L}^{\vec} \gL
    \\
    \vdots & \cdots & \ddots & \vdots
    \\
    \ggn_{L, 1}^{\vec} \gL
    &
      \ggn_{L, 2}^{\vec} \gL
    &
      \cdots
    &
      \ggn_L^{\vec} \gL
  \end{pmatrix}\,.
\end{align*}
For KFAC, we only need its block diagonal matrix
\begin{align*}
  \tilde{\ggn}_{\vtheta}^{\vec} \gL
  =
  \begin{pmatrix}
    \ggn_1^{\vec} \gL & \vzero & \cdots & \vzero
    \\
    \vzero & \ggn_2^{\vec} \gL & \ddots & \vdots
    \\
    \vdots & \ddots & \ddots & \vzero
    \\
    \vzero & \cdots & \vzero & \ggn_L^{\vec} \gL
  \end{pmatrix}\,,
\end{align*}
\ie, individual blocks $\{ \ggn_{\vtheta^{(k)}}^{\vec} \gL_{\sD}(\vtheta)\}_{k=1}^L$.

\paragraph{The GGN as a self-outer product.}
Let us look at one last aspect of the GGN, which makes it convenient to relate it to other curvature matrices we are about to discuss.
Consider the loss contributed by a single datum and suppress the index $_n$ for now, as well as the reduction factor $R$.
Its contribution to the GGN is
\begin{align*}
  \underbrace{[\jac_{\vtheta}^{\vec} \vf]^{\top}}_{\dim(\Theta) \times \dim(\gF)}
  \underbrace{[\hess^{\vec}_{f} c(\vf)]}_{\dim(\gF) \times \dim(\gF)}
  \underbrace{[\jac_{\vtheta}^{\vec} \vf]}_{\dim(\gF) \times \dim(\Theta)}\,.
\end{align*}
We will now make this more symmetric.
By assumption, the criterion function $c$ is convex in $\vf$.
This means that the flattened Hessian $\hess^{\vec}_{\vf} c(\vf)$ is positive semi-definite.
Since any positive semi-definite matrix $\mA \in \sR^{C \times C}$ can be expressed as an outer product $\mA = \mB \mB^{\top}$ where $\mB \in \sR^{C \times \rank(\mA)}$, we can find a symmetric factorization $\mS^{\vec} \in \sR^{\dim(\gF) \times \dim(\gF)}$ of the criterion's Hessian such that $\hess^{\vec}_{\vf} c(\vf) = \mS^{\vec} {\mS^{\vec}}^{\top}$.
With that, we can then write the upper expression as
\begin{align*}
  &[\jac_{\vtheta}^{\vec} \vf]^{\top} \mS^{\vec} {\mS^{\vec}}^{\top} [\jac_{\vtheta}^{\vec} \vf]
  \\
  &=
    \underbrace{([\jac_{\vtheta}^{\vec} \vf]^{\top} \mS^{\vec})}_{\coloneq \mV^{\vec} \in \sR^{\dim(\Theta) \times \dim(\gF)}}
    ([\jac_{\vtheta}^{\vec} \vf]^{\top} \mS^{\vec})^{\top}
  \\
  &=
    \mV^{\vec} {\mV^{\vec}}^{\top}\,.
\end{align*}
In words, we can express the GGN contribution of a single datum as a self-outer product.
While we will not use $\mV^{\vec}$ (see \eg \citet{dangel2022vivit,ren2019efficient} for detailed discussions), we need the loss Hessian factorization ($\mS^{\vec}$) later.

\Cref{ex:mseloss_hessian_factorization,ex:crossentropyloss_hessian_factorization} present the loss Hessian factorizations of the square and softmax cross-entropy losses.
One important insight for these two loss functions is that $\mS^{\vec}$ does not depend on the labels.
We will use this in \Cref{subsec:connection-ggn-fisher} to connect the GGN with the Fisher for regression and classification.
But first, let's introduce the Fisher.

\switchcolumn[1]
\begin{example}[Symmetric factorization of the square loss Hessian, \Cref{basics/hessian_factorizations}]\label{ex:mseloss_hessian_factorization}
  Consider the square loss $c$ from \Cref{ex:square_loss} and its Hessian from \Cref{ex:square_loss_hessian}.
  The Hessian's symmetric factorization is simply $\mS^{\vec} = \mI$ because $\mS^{\vec} {\mS^{\vec}}^{\top} = \mI = \hess_{\vf}^{\vec}c$.
\end{example}

\begin{example}[Symmetric factorization of the softmax cross-entropy loss Hessian, \Cref{basics/hessian_factorizations}]\label{ex:crossentropyloss_hessian_factorization}
  The Hessian's symmetric factorization is \citep[\eg][]{papyan2019measurements}
  \begin{align*}
    \mS^{\vec} = \diag(\sqrt{\vsigma}) - \vsigma \sqrt{\vsigma}^{\top}
  \end{align*}
  where $\vsigma = \softmax(\vf)$ and the square root is applied element-wise.
  To see this, we can form $\mS^{\vec} {\mS^{\vec}}^{\top}$ which yields
  \begin{align*}
    \diag(\vsigma) - 2 \vsigma \vsigma^{\top} + \vsigma \sqrt{\vsigma}^{\top} \sqrt{\vsigma} \vsigma^{\top}
    = \diag(\vsigma) - \vsigma \vsigma^{\top}
  \end{align*}
  using the fact that $\vsigma^{\top} \vone = 1 = \sqrt{\vsigma}^{\top} \sqrt{\vsigma}$.
  This expression equals the Hessian from \Cref{ex:hessian-crossentropyloss}.
\end{example}

\codeblock{basics/hessian_factorizations}
\switchcolumn[0]

\subsubsection{The Fisher}\label{sec:fisher}
\switchcolumn[1]
\codeblock{basics/mc_fishers}

\begin{figure}[!h]
  \centering
  \includegraphics[width=\linewidth]{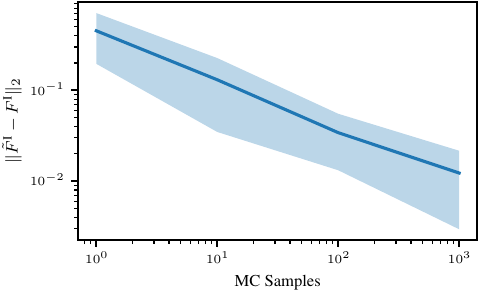}
  \caption{ \textbf{Convergence of the Monte Carlo-approximated type-I Fisher to the type-I Fisher.}
    We visualize the spectral norm of their residual as a function of MC samples.
    The Fishers were evaluated on synthetic data ($N=100$) using an MLP with three fully-connected layers and ReLU activations (5-4-4-3) and square loss.
    In this setting, the exact type-I Fisher equals the GGN, and we used GGNVPs (\Cref{def:ggnvp}) to compute it.
    Plot produced with \repofile{plots/synthetic_fisher}.
  }\label{fig:mc-fisher-converges-to-fisher}
\end{figure}

\codeblock{basics/mc_fisher_product}
\switchcolumn[0]

\paragraph{Recap: Probabilistic interpretation.}
In \Cref{subsec:probabilistic-interpretation} we demonstrated that a neural net often models a likelihood for the labels given the inputs using parameters.
This probabilistic perspective provides a new way to compare two models $f_1 \coloneq f(\bullet, \vtheta_1), f_2 \coloneq f(\bullet, \vtheta_{2})$ where $\vtheta_2 = \vtheta_1 + \Delta \vtheta$: Instead of assessing their dissimilarity by their parameters' dissimilarity, \eg using the Euclidean distance measure $d^2(f_1, f_2) = \left\lVert \vtheta_2 - \vtheta_1 \right\rVert_2^2 = \left\lVert \Delta \vtheta \right\rVert_2^2$, we can instead use the KL divergence of the represented probability densities, $d^2(f_1, f_2) = \mathrm{KL}(p(\rvx, \rvy \mid \vtheta_1) \mid\mid p(\rvx, \rvy \mid \vtheta_2))$.

\paragraph{The Fisher as metric of the KL divergence.}
Using $p(\rvx, \rvy \mid \vtheta) = p(\rvy \mid \rvx, \vtheta)p_{\text{data}}(\rvx)$, we have
\begin{align*}
  & \mathrm{KL}(p(\rvx, \rvy \mid \vtheta_1) \mid\mid p(\rvx, \rvy \mid \vtheta_2))                               \\
  = & \E_{p_{\text{data}}(\rvx)} [\mathrm{KL}(p(\rvy \mid \rvx, \vtheta_1) \mid\mid p(\rvy \mid \rvx, \vtheta_2))].
\end{align*}
While the KL divergence is not a valid distance measure for general pairs $(\vtheta_1, \vtheta_2)$ (\eg it is not symmetric \wrt its arguments), it can be locally approximated by a quadratic Taylor series for small $\Delta \vtheta$.
The curvature matrix in that Taylor series is the Fisher,
\begin{align*}
  \mathrm{KL}(p(\rvx, \rvy \mid \vtheta) \mid\mid p(\rvx, \rvy \mid \vtheta + \Delta \vtheta))
  \\
  = \frac{1}{2} {\Delta \vtheta}^{\top} \mF(\vtheta) \Delta \vtheta + \gO( \Vert\Delta \vtheta\Vert^3)\,,
\end{align*}
with the Fisher information matrix
\begin{align*}
  \mF(\vtheta) &\coloneqq \E_{p(\rvx, \rvy \mid \vtheta)} [-\hess_{\vtheta} \log p(\rvx, \rvy \mid \vtheta)]\,.
                 \intertext{Let's rewrite this expression in terms of the likelihood $p(\rvy \mid \vx, \vtheta)$, and its reparameterized likelihood $r(\rvy \mid f(\rvx, \vtheta))$ with the neural net prediction,}
  &= \E_{p_{\text{data}}(\rvx)}\E_{p(\rvy \mid \rvx, \vtheta)} [-\hess_{\vtheta} \log p(\rvy \mid \rvx, \vtheta)]
  \\
               &=
                 \E_{p_{\text{data}}(\rvx)}\E_{r(\rvy \mid f(\rvx, \vtheta))} [
  \\
               &\qquad\qquad\qquad-\hess_{\vtheta} \log r(\rvy \mid f(\rvx, \vtheta))]\,.
\end{align*}
The expectation over the data-generating distribution's marginal $p_{\text{data}}(\rvx)$ is intractable.
Therefore, like before, we replace it with the empirical distribution over our data set in the following.
Following \citet{soen2024tradeoffs}, we present two forms of the Fisher: While the \emph{type-II Fisher} allows to elegantly connect the Fisher and the GGN and highlights its connection to the Hessian (a curvature matrix), the \emph{type-I Fisher} highlights that the Fisher is a gradient covariance matrix, and can therefore be computed with first-order derivatives.

\paragraph{Type-II Fisher.}
Let's replace the true marginal distribution over the inputs with its empirical distribution $p_{\sD}(\rvx) = \frac{1}{N}\sum_n \delta(\rvx - \vx_n)$.
This yields the type-II Fisher information matrix (because it is defined via second-order derivatives),
\begin{align*}
  \mF^{\text{II}}(\vtheta)
  & =
    \E_{p_{\sD}(\rvx)}\E_{r(\rvy \mid f(\rvx, \vtheta))} [
  \\
  &\qquad\qquad\qquad
    -\hess_{\vtheta} \log r(\rvy \mid f(\rvx, \vtheta))]
  \\
  & =
    \frac{1}{N} \sum_n
    \E_{r(\rvy \mid \vf_n)} [-\hess_{\vtheta} \log r(\rvy \mid \vf_n)]
    \shortintertext{(remember $\vf_n \coloneq f(\rvx = \vx_n, \vtheta)$).
    We can apply the chain rule to the log-likelihood's Hessian (remember the 1d example from \Cref{sec:partial_linearization}) and use that $\E_{r(\rvy \mid f(\rvx, \vtheta))}[\nabla_{f(\rvx, \vtheta)} \log r(\rvy \mid f(\rvx, \vtheta))] = \vzero$ to simplify the type-II Fisher into}
  &= \frac{1}{N} \sum_n \E_{r(\rvy \mid \vf_n)} [
    \begin{aligned}[t]
      & {\jac_{\vtheta} \vf_n}^{\top}                          \\
      & \left( -\hess_{\vf_n} \log r(\rvy \mid \vf_n)  \right) \\
      & \jac_{\vtheta} \vf_n ]\,
    \end{aligned}
    \shortintertext{and, as the Jacobians do not depend on $\rvy$,}
  & = \frac{1}{N} \sum_n
    \begin{aligned}[t]
      & {\jac_{\vtheta} \vf_n}^{\top} \\
      & \left(
        \E_{r(\rvy \mid \vf_n)} [
        -\hess_{\vf_n} \log r(\rvy \mid \vf_n)
        ]
        \right)                          \\
      & \jac_{\vtheta} \vf_n \,.
    \end{aligned}
\end{align*}
This form resembles the GGN a lot.
In fact, for square loss and softmax cross-entropy loss ($c = - \log$), we derived that the loss Hessian we average in the type-II Fisher does not depend on the labels modelled by the distribution (see \Cref{ex:hessian-crossentropyloss,ex:square_loss_hessian}).
This means we can drop the expectation for such losses.
We just showed that the (type-II) \emph{Fisher equals the GGN for regression with square loss and classification with softmax cross-entropy loss} (more details in \Cref{subsec:connection-ggn-fisher}).

\paragraph{Type-I Fisher.} There is another way to rewrite the Fisher in terms of first-order derivatives of the log-likelihood.
We will call this the type-I Fisher information matrix.
To derive it, we use the property that the negative log-likelihood's Hessian equals its gradient covariance in expectation,
\begin{align*}
  & \E_{p(\rvx, \rvy \mid \vtheta)}
    \left[
    -\hess_{\vtheta} \log p(\rvx, \rvy \mid \vtheta)
    \right]
  \\
  = & \E_{p(\rvx, \rvy \mid \vtheta)} [
      \begin{aligned}[t]
        & (-\nabla_{\vtheta} \log p(\rvx, \rvy \mid \vtheta)) \\
        & (-\nabla_{\vtheta} \log p(\rvx, \rvy\mid  \vtheta))^{\top}].
      \end{aligned}
\end{align*}
Hence, we can write the Fisher as
\begin{align*}
  &\mF(\vtheta)
  \\
  & = \E_{p(\rvx, \rvy \mid \vtheta)} [
    \begin{aligned}[t]
      & (-\nabla_{\vtheta} \log p(\rvx, \rvy \mid \vtheta))        \\
      & (-\nabla_{\vtheta} \log p(\rvx, \rvy \mid \vtheta))^{\top}]
    \end{aligned}
    \shortintertext{(factorize $p(\rvx, \rvy \mid \vtheta) = p_{\text{data}}(\rvx) p(\rvy \mid \rvx, \vtheta)$)}
  & = \E_{p_{\text{data}}(\rvx)}\E_{p(\rvy \mid \rvx, \vtheta))} [
    \begin{aligned}[t]
      & (-\nabla_{\vtheta} \log p(\rvy \mid \rvx, \vtheta))        \\
      & (-\nabla_{\vtheta} \log p(\rvy \mid \rvx, \vtheta))^{\top}]
    \end{aligned}
    \shortintertext{(reparameterize $p(\rvy \mid \rvx, \vtheta) = r(\rvy \mid f(\rvx, \vtheta))$)}
  & = \E_{p_{\text{data}}(\rvx)}\E_{r(\rvy \mid f(\rvx, \vtheta))} [
  \\
  &\qquad \qquad\qquad
    (-\nabla_{\vtheta} \log r(\rvy \mid f(\rvx, \vtheta)))
  \\
  &\qquad \qquad\qquad
    (-\nabla_{\vtheta} \log r(\rvy \mid f(\rvx, \vtheta)))^{\top}].
\end{align*}
Let's apply the chain rule for the gradient, substitute the empirical distribution for $p_{\text{data}}(\rvx)$, and simplify.
This gives the type-I Fisher
\begin{align*}
  & \mF^{\text{I}}(\vtheta)
  \\
  & =
    \frac{1}{N} \sum_n
    \E_{r(\rvy \mid \vf_n)}
    [\begin{aligned}[t]
      & (-\nabla_{\vtheta} \log r(\rvy \mid \vf_n)) \\
      & (-\nabla_{\vtheta} \log r(\rvy \mid \vf_n))^{\top}]
    \end{aligned}
  \\
  & =
    \frac{1}{N} \sum_n
    \E_{r(\rvy \mid \vf_n)}
    [\begin{aligned}[t]
      & (\jac_{\vtheta}\vf_n)^{\top} \\
      & (-\nabla_{\vf_n} \log r(\rvy \mid \vf_n)) \\
      & (-\nabla_{\vf_n} \log r(\rvy \mid \vf_n))^{\top} \\
      & \jac_{\vtheta}\vf_n]
    \end{aligned}
  \\
  \shortintertext{and, as the Jacobians do not depend on $\rvy$,}
  & =
    \frac{1}{N} \sum_n
    \begin{aligned}[t]
      & (\jac_{\vtheta}\vf_n)^{\top} \\
      & \E_{r(\rvy \mid \vf_n)}
        [(-\nabla_{\vf_n} \log r(\rvy \mid \vf_n)) \\
      & \phantom{\E_{r(\rvy \mid \vf_n)}[}(-\nabla_{\vf_n} \log r(\rvy \mid \vf_n))^{\top}] \\
      & \jac_{\vtheta}\vf_n \,.
    \end{aligned}
\end{align*}
This form highlights that the Fisher can be computed with first-order derivatives, specifically vector-Jacobian products.

\paragraph{Monte Carlo-approximated type-I Fisher.} For computations with the type-I Fisher, we need to replace the expectation over $r(\rvy \mid \vf_n)$ with an estimator, \eg Monte Carlo approximation with $M$ sampled labels $\smash{\tilde{\vy}_{n,1}, \dots, \tilde{\vy}_{n, M} \stackrel{\text{i.i.d}}{\sim} r(\rvy \mid \vf_n)}$ for the prediction of datum $n$.
Thankfully, we have already developed the functionality to sample such labels in \Cref{subsec:probabilistic-interpretation} (specifically, \Cref{basics/label_sampling}).

\begin{definition}[Monte Carlo-approximated type-I Fisher (\Cref{basics/mc_fishers})]\label{def:mc_fisher}%
  The Monte Carlo-approximated type-I Fisher of the likelihood $\log r(\rvy \mid f(\vx_n, \vtheta))$,
  $\tilde{\mF}^{\text{I}}(\vtheta) \in \sR^{D \times D}$ is
  \begin{align*}
    & \tilde{\mF}^{\text{I}}(\vtheta) \\
    & = \frac{1}{NM} \sum_{n,m}
      \begin{aligned}[t]
        & (-\nabla_{\vtheta} \log r(\rvy = \tilde{\vy}_{n,m} \mid \vf_n))        \\
        & (-\nabla_{\vtheta} \log r(\rvy = \tilde{\vy}_{n,m} \mid \vf_n))^{\top} \\
      \end{aligned} \\
    & = \frac{1}{NM} \sum_{n,m}
      \begin{aligned}[t]
        & \left(\jac_{\vtheta}\vf_n\right)^{\top}                              \\
        & (-\nabla_{\vf_n} \log r(\rvy = \tilde{\vy}_{n,m} \mid \vf_n))        \\
        & (-\nabla_{\vf_n} \log r(\rvy = \tilde{\vy}_{n,m} \mid \vf_n))^{\top} \\
        & \jac_{\vtheta}\vf_n.
      \end{aligned}
  \end{align*}
\end{definition}
As we increase the number of Monte Carlo (MC) samples, the MC-approximated type-I Fisher converges to the Fisher, as illustrated in \Cref{fig:mc-fisher-converges-to-fisher}.

\paragraph{Would-be gradients \& self-outer product form.}
To compute with the type-I Fisher, we need to evaluate the gradient of the loss using sampled labels, rather than the labels from the data set (see \Cref{basics/mc_fisher_product}, which implements multiplication with the MC Fisher).
As coined by \citet{papyan2020traces}, we call these gradients `would-be' gradients, because they depend on hypothetical labels.

Let's denote the $m$-th negative would-be gradient for sample $n$ by $\tilde{\vg}_{n,m} \coloneq -\nabla_{\vf_n} \log r(\rvy = \tilde{\vy}_{n,m} \mid \vf_n)$
We can then form the matrix
\begin{align*}
  \tilde{\mS}_n
  \coloneq
  \frac{1}{\sqrt{M}}
  \begin{pmatrix}
    \tilde{\vg}_{n,1} & \cdots & \tilde{\vg}_{n,M}
  \end{pmatrix}
  \in \sR^{\dim(\gF) \times M}
\end{align*}
for each $n$, and express the MC Fisher as
\begin{align*}
  \tilde{\mF}^{\text{I}}(\vtheta)
  & =
    \frac{1}{N} \sum_{n,m}
    (\jac_{\vtheta}\vf_n)^{\top}
    \tilde{\mS}_n
    \tilde{\mS}_n^{\top}
    (\jac_{\vtheta}\vf_n)
  \\
  & =
    \frac{1}{N} \sum_n
    \tilde{\mV}_n
    \tilde{\mV}_n^{\top}
\end{align*}
where $\tilde{\mV}_n = (\jac_{\vtheta}\vf_n)^{\top} \tilde{\mS}_n \in \sR^{D \times M}$.

Did you notice what we just did here?
By stacking the negative would-be gradients into columns of a matrix, we were able to express the Fisher as the self-outer product of a matrix $\tilde{\mV}_n$.
This notation looks very similar to the GGN's self-outer product notation.
In fact, we can think of $\tilde{\mV}_n \in \sR^{D \times M}$ as a \emph{randomization} of $\mV_n \in \sR^{D \times C}$ which can be much smaller (depending on $M$ and $C$) and therefore cheaper to compute in exchange for inaccuracy.
The different self-outer product forms of curvature matrices will give rise to multiple flavours of KFAC in \Cref{sec:kfac-overview}.

\paragraph{Reduction factor of the Fisher vs.\,empirical risk.}
In the above derivation, we arrived at an expression of the Fisher which uses a reduction factor $R = \nicefrac{1}{N}$.
We will handle this factor flexibly in practice by using the reduction factor of our empirical risk.

\paragraph{(Optional) efficient would-be gradient computation.}
One way to compute would-be gradients is to repeatedly sample a label, compute its loss, and then use automatic differentiation to obtain the gradient.
This incurs \texttt{for} loops, which may be sub-optimal for performance.
In this tutorial, we follow the less efficient \texttt{for} loop approach as it is less complex.
To improve efficiency, we could draw all labels in parallel and compute their gradients manually instead of using autodiff, which is cheaper, as demonstrated by \Cref{ex:square-loss-gradient,ex:cross-entropy-loss-gradient}.
They provide the gradients of square loss and softmax cross-entropy loss, which are straightforward to evaluate manually.
However, we will not apply this optimization in our tutorial.

\switchcolumn[1]
\begin{example}[Gradients of square loss]\label{ex:square-loss-gradient}
  Consider the square loss from \Cref{ex:square_loss} and its probabilistic interpretation from \Cref{ex:square_loss_probabilistic}
  For some label $\vy$, the gradient is given by
  \begin{align*}
    \nabla_{\vf} c(\vf, \vy)
    & =
      - \nabla_{\vf} \log r(\rvy =\vy \mid \vf)
    \\
    & =
      \vf - \vy\,.
  \end{align*}
\end{example}

\begin{example}[Gradients of softmax cross-entropy loss]\label{ex:cross-entropy-loss-gradient}
  Consider the softmax cross-entropy loss from \Cref{ex:cross_entropy_loss} and its probabilistic interpretation from \Cref{ex:cross_entropy_loss_probabilistic}. For some label $y$, the gradient is given by
  \begin{align*}
    \nabla_{\vf} c(\vf, y)
    & =
      - \nabla_{\vf} \log r(\ry = y \mid \vf)
    \\
    & =
      \vsigma(\vf) - \mathrm{onehot}(y)\,.
  \end{align*}
\end{example}
\switchcolumn[0]

\subsubsection{Connection between GGN \& Fisher}\label{subsec:connection-ggn-fisher}
The GGN, type-I, and type-II Fisher are all weighted sums of matrices sandwiched between the per-sample Jacobians (remember that we can mentally set $-\log r(\rvy \mid \vf_n) = c(\vf_n, \rvy)$ for square and softmax cross-entropy loss):
\begin{align*}
  \mG(\vtheta)
  & =
    R \sum_n
    \begin{aligned}[t]
      & (\jac_{\vtheta}\vf_n)^\top                           \\
      & \textcolor{VectorBlue}{\hess_{\vf_n}c(\vf_n, \vy_n)} \\
      & \jac_{\vtheta}\vf_n\,,
    \end{aligned}
  \\
  \mF^{\text{II}}(\vtheta)
  & =
    R \sum_n
    \begin{aligned}[t]
      & (\jac_{\vtheta}\vf_n)^\top                                                                 \\
      & \textcolor{VectorPink}{\E_{r(\rvy \mid \vf_n)}[-\hess_{\vf_n} \log( r(\rvy \mid \vf_n)) ]} \\
      & \jac_{\vtheta}\vf_n\,,
    \end{aligned}
  \\
  \mF^{\text{I}}(\vtheta)
  & =
    R \sum_n
    \begin{aligned}[t]
      & (\jac_{\vtheta}\vf_n)^\top
      \\
      & \colored[VectorTeal]{\E_{r(\rvy \mid \vf_n)}[}
      \\
      &\begin{aligned}[t]
        &\quad \colored[VectorTeal]{-\nabla_{\vf_n} \log( r(\rvy \mid \vf_n))}
        \\
        &\quad \colored[VectorTeal]{(-\nabla_{\vf_n} \log( r(\rvy \mid \vf_n)))^{\top}]}
      \end{aligned}
      \\
      & \jac_{\vtheta}\vf_n\,.
    \end{aligned}
\end{align*}
In previous sections, we showed that for square loss and softmax cross-entropy, the criterion function's Hessian $\hess_\vf c(\vf, \vy) = -\hess_{\vf} \log( r(\rvy = \vy \mid \vf)$ does not depend on the value of the target random variable $\rvy$!
Therefore, the expectation in the type-II Fisher effectively disappears, and we are free to set $\rvy = \vy_n$ because this does not change the Hessian.
This means the type-II Fisher and GGN are equivalent for these losses.
Note that we cannot drop the expectation in the type-I Fisher, though\footnote{This is precisely why we needed a separate definition for the Monte Carlo-approximated type-I Fisher to make it computationally tractable.}.
But from the equivalence of type-I and type-II Fisher, we know that it also equals the GGN in the above scenarios.

\subsubsection{The Empirical Fisher (EF)}\label{sec:emp_fisher}
Finally, we introduce the empirical Fisher.
This matrix serves as an approximation to the type-I Fisher that does not require Monte Carlo samples.
Several contemporary optimizers, \eg the Adam variants, gained inspiration from second-order optimization, in particular preconditioning with the empirical Fisher.
For example, Adam \cite{kingma2015adam} stores the moving average of the mini-batch loss' squared gradients, which is motivated by (but is not equivalent to) preconditioning with the \emph{diagonal} of the empirical Fisher\footnote{See~\citet{lin2024can} for the exact preconditioner Adam approximates, which is a different approximation of the Fisher information matrix.}.
We obtain the empirical Fisher by replacing the model's likelihood in the type-I Fisher with the empirical likelihood and evaluating the expectation over it:

\begin{definition}[Empirical Fisher (\Cref{basics/emp_fishers})]\label{def:emp_fisher}%
  The empirical Fisher matrix of the likelihood $\log r(\rvy \mid \vf_n)$,
  $\mE(\vtheta) \in \sR^{D \times D}$, is
  \begin{align*}
    & \mE(\vtheta) \\
	& = \frac{1}{N} \sum_{n}
	\begin{aligned}[t]
	   & (-\nabla_{\vtheta} \log r(\rvy = \vy_n \mid \vf_n))        \\
	   & (-\nabla_{\vtheta} \log r(\rvy = \vy_n \mid \vf_n))^{\top} \\
	\end{aligned} \\
    & = \frac{1}{N} \sum_{n}
    \begin{aligned}[t]
       & \left(\jac_{\vtheta}\vf_n\right)^{\top}                  \\
       & (-\nabla_{\vf_n} \log r(\rvy = \vy_n \mid \vf_n))        \\
       & (-\nabla_{\vf_n} \log r(\rvy = \vy_n \mid \vf_n))^{\top} \\
       & \jac_{\vtheta}\vf_n.
    \end{aligned}
  \end{align*}
\end{definition}

\switchcolumn[1]
\codeblock{basics/emp_fishers}
\codeblock{basics/emp_fisher_product}
\switchcolumn[0]

\Cref{basics/emp_fisher_product} implements multiplication with the EF.

\paragraph{Empirical Fisher $\neq$ Fisher.}
The subtle difference between empirical and type-I Fisher is the \emph{expectation} over the gradient outer product
\begin{align*}
  (-\nabla_{\vtheta} \log r(\rvy \mid \vf_n))
  (-\nabla_{\vtheta} \log r(\rvy \mid \vf_n))^{\top}
\end{align*}
\wrt the model's predictive distribution $r(\rvy \mid \vf_n)$, whereas $\mE(\vtheta)$ plugs in the \emph{ground-truth} labels $\vy_n$ into the gradient outer product.
While the computation of the empirical Fisher is more efficient than Monte Carlo approximating the type-I Fisher with number of samples $M > 1$, this subtle difference can impact the utility of this matrix in optimization.
In particular,~\citet{kunstner2019limitations} show that preconditioning with the EF can have detrimental effects in simple problems.
The empirical Fisher's success in some settings (\eg, through Adam) may be attributed to its ability to attenuate \emph{gradient noise}, not its properties as a curvature estimate.
To some extent, this is still an open question.

\paragraph{Visual comparison.} For our purposes, it suffices to acknowledge that the empirical Fisher is a popular curvature approximation that is often used due to its computational advantages over the Fisher and GGN but differs from them, and can also be approximated using KFAC.
The differences between GGN, MC type-I Fisher, and empirical Fisher, as well as their variation under different flattening conventions, are visualized in \Cref{fig:visual-comparison-mc-empirical-fisher}.

\begin{figure}[!h]
  \centering
  \begin{minipage}[t]{0.49\linewidth}
    \centering
    $\cvec 1$ sample\vspace{1ex}
    \includegraphics[width=1.0\linewidth]{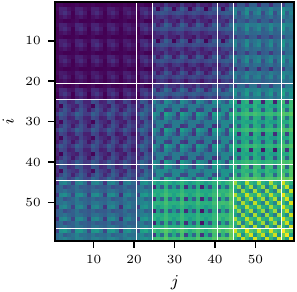}
  \end{minipage}
  \hfill
  \begin{minipage}[t]{0.49\linewidth}
    \centering
    $\rvec 1$ sample\vspace{1ex}
    \includegraphics[width=1.0\linewidth]{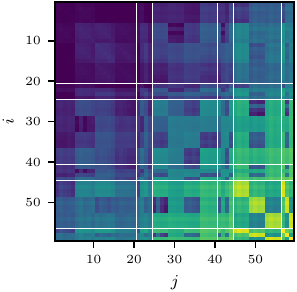}
  \end{minipage}
  \\
  \begin{minipage}[t]{0.49\linewidth}
    \centering
    $\cvec 100$ samples\vspace{1ex}
    \includegraphics[width=1.0\linewidth]{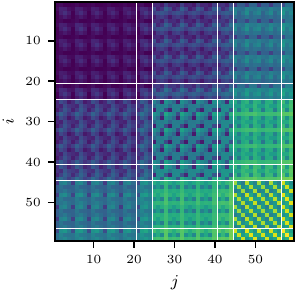}
  \end{minipage}
  \hfill
  \begin{minipage}[t]{0.49\linewidth}
    \centering
    $\rvec 100$ samples\vspace{1ex}
    \includegraphics[width=1.0\linewidth]{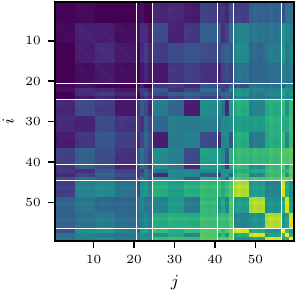}
  \end{minipage}
  \begin{minipage}[t]{0.49\linewidth}
    \centering
    $\cvec$ exact\vspace{1ex}
    \includegraphics[width=1.0\linewidth]{kfs/plots/synthetic_cvec_ggn.pdf}
  \end{minipage}
  \hfill
  \begin{minipage}[t]{0.49\linewidth}
    \centering
    $\rvec$ exact\vspace{1ex}
    \includegraphics[width=1.0\linewidth]{kfs/plots/synthetic_rvec_ggn.pdf}
  \end{minipage}
  \begin{minipage}[t]{0.49\linewidth}
    \centering
    $\cvec$ empirical\vspace{1ex}
    \includegraphics[width=1.0\linewidth]{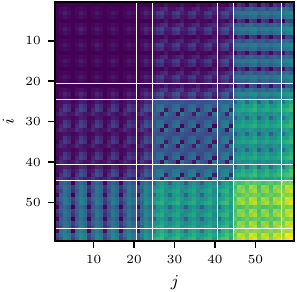}
  \end{minipage}
  \hfill
  \begin{minipage}[t]{0.49\linewidth}
    \centering
    $\rvec$ empirical\vspace{1ex}
    \includegraphics[width=1.0\linewidth]{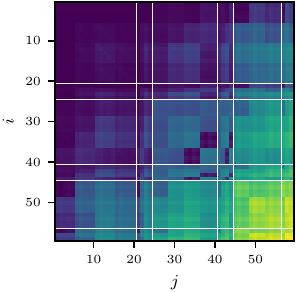}
  \end{minipage}
  \caption{\textbf{Visual comparison of Fisher matrices.}
    The Fisher blocks are visually highlighted with white lines.
    The left column uses $\cvec$-flattening, the right column $\rvec$-flattening.
    The penultimate row uses the exact Fisher.
    The last row shows the empirical Fisher, highlighting its difference from the previous curvature estimates.
    The Fishers were evaluated on synthetic data ($N=100$) using an MLP with three fully-connected layers and ReLU activations (5-4-4-3) and square loss.
    Created with \repofile{plots/synthetic_fisher}.
  }\label{fig:visual-comparison-mc-empirical-fisher}
\end{figure}

  \switchcolumn[0]*
\end{paracol}
\clearpage

\begin{paracol}{1}
  \section{Cheatsheet: Basics}\label{sec:cheatsheet-basics}
  \begin{itemize}
\item Risk minimization and tractable empirical risk minimization (\Cref{subsec:empirical-risk-minimization})
  \begin{align*}
    \argmin_{\vtheta} \gL_{p_{\text{data}}(\rvx, \rvy)}(\vtheta)
    \qquad
    &\text{where}\qquad
      \gL_{p_{\text{data}}(\rvx, \rvy)}(\vtheta) \coloneq \E_{(\vx, \vy) \sim p_{\text{data}}(\rvx, \rvy)}
      \left[
      c(f(\vx, \vtheta), \vy)
      \right]
    &\text{(intractable)}
      \shortintertext{(use empirical density $p_{\sD}(\rvx, \rvy) = \frac{1}{N} \sum_n \delta(\rvx - \vx_n) \delta(\rvy - \vy_n)$ from data $\sD = \{ (\vx_n, \vy_n) \sim p_{\text{data}} \mid n=1, \dots, N \}$)}
      \argmin_{\vtheta} \gL_{\sD}(\vtheta)
      \qquad
    &\text{where}\qquad
      \gL_{\sD}(\vtheta) \coloneq R \sum_{n=1}^N c(f(\vx_n; \vtheta), \vy_n)\,.
    &\text{(tractable)}
  \end{align*}
  (changing the reduction factor $R$ does not change the optimization problem's solution)

\item Common criterion functions and their reduction constants (\Cref{ex:square_loss,ex:cross_entropy_loss})
  \begin{align*}
    &\begin{matrix}
      \text{Square loss}
      \\
      \text{(\texttt{MSELoss})}
    \end{matrix}
      \qquad
    &c(\vf, \vy) = \frac{1}{2} \left\lVert \vf - \vy \right\rVert_2^2\,,
      \qquad
    &R =
      \begin{cases}
        2 & \text{\texttt{reduction="sum"}} \\
        \frac{2}{N \dim(\vy)} & \text{\texttt{reduction="mean"}}
      \end{cases}
    \\
    &\begin{matrix}
      \text{Softmax cross-entropy loss}\\
      \text{(\texttt{CrossEntropyLoss})}
    \end{matrix}
      \qquad
    &c(\vf, y) = - \log( [\softmax(\vf)]_y)\,,
      \qquad
    &R =
      \begin{cases}
        1 & \text{\texttt{reduction="sum"}} \\
        \frac{1}{N \dim(\vf)} & \text{\texttt{reduction="mean"}}
      \end{cases}
  \end{align*}

\item Probabilistic interpretation of a neural net: Parameterize $p(\rvx, \rvy \mid \vtheta) = p_{\text{data}}(\rvx) p(\rvy \mid \rvx, \vtheta) = p_{\text{data}}(\rvx) r(\rvy \mid f(\rvx, \vtheta))$
  \begin{align*}
    \argmin_{\vtheta} \mathrm{KL}( p_{\text{data}}(\rvx, \rvy) \mid\mid p(\rvx, \rvy \mid \vtheta) )
    \quad\Leftrightarrow\quad
    &\argmin_{\vtheta} \E_{p_{\text{data}}(\rvx)} \E_{r(\rvy \mid \rvx, \vtheta)} \left[
      - \log r(\rvy \mid f(\rvx, \vtheta))
      \right]
    &\text{(intractable)}
      \shortintertext{(use empirical density to make tractable)}
      \qquad
    &\argmin_{\vtheta} - R \sum_{n=1}^N \log r(\rvy=\vy_n \mid f(\rvx=\vx_n, \vtheta))
    &\text{(tractable)}
  \end{align*}

\item Common criteria are negative log-likelihoods: $c(\vf, \vy) = - \log r(\rvy=\vy \mid f(\rvx, \vtheta) = \vf)$ (\Cref{ex:square_loss_probabilistic,ex:cross_entropy_loss_probabilistic})
  \begin{align*}
    &\text{Square loss (\texttt{MSELoss})}
      \qquad
    &r(\rvy \mid \vf) = \gN( \rvy \mid \vmu = \vf, \mSigma = \mI)
    \\
    &\text{Softmax cross-entropy loss (\texttt{CrossEntropyLoss})}
      \qquad
    &r(\ry \mid \vf) = \gC( \ry \mid \vsigma = \softmax(\vf))
  \end{align*}

\item Shorthands: Per-datum prediction $\vf_n(\vtheta) = f(\vx_n, \vtheta)$, criterion $c_n(\vf_n) = c(\vf_n, \vy_n)$, and loss $\ell_n(\vtheta) = c_n(\vf_n(\vtheta))  $

\item Net Jacobian $\jac_{\vtheta}\vf \in \sR^{\dim(\gF) \times D}$, $[\jac_{\vtheta} \vf]_{i,j} = \frac{\partial [\vf]_i}{\partial [\vtheta]_j}$, criterion Hessian $\hess_{\vf} c \in \sR^{\dim(\gF) \times \dim(\gF)}$, $[\hess_{\vf}c]_{i,j} = \frac{\partial^2 c}{\partial [\vf]_i \partial [\vf]_j}$

\item Hessian, generalized Gauss-Newton, type-II/I/empirical Fishers ($\vf_n \coloneq f(\vx_n, \vtheta)$, $\rvf = f(\rvx, \vtheta)$, $\tilde{\vy}_{n,m} \sim r(\rvy \mid \rvf = \vf_n)$)
  \begin{align*}
    \hess_{\vtheta} \gL(\vtheta)
    &= R \sum_{n=1}^N \hess_{\vtheta} c(\vf_n, \vy_n)
      = -R \sum_{n=1}^N \hess_{\vtheta} \log r(\rvy = \vy_n \mid \rvf = \vf_n)
    \\
    \mG(\vtheta)
    &= R \sum_{n=1}^N
      \jac_{\vtheta} \vf_n^{\top}
      \left(
      \hess_{\vf_n} c(\vf_n, \vy_n)
      \right)
      \jac_{\vtheta} \vf_n
      =
      R \sum_{n=1}^N
      \jac_{\vtheta} \vf_n^{\top}
      \left(
      -\hess_{\vf_n} \log r(\rvy = \vy_n \mid \rvf = \vf_n)
      \right)
      \jac_{\vtheta} \vf_n
    \\
    \mF^{\text{II}}(\vtheta)
    &=
      \lim_{M \to \infty}
      \frac{R}{M}
      \sum_{n=1}^N
      \jac_{\vtheta} \vf_n^{\top}
      \left[
      \hess_{\vf_n}(\underbrace{- \log r(\rvy = \tilde{\vy}_{n,m} \mid \rvf = \vf_n)}_{= c(\vf_n, \tilde{\vy}_{n,m})} )
      \right]
      \jac_{\vtheta} \vf_n
    \\
    \mF^{\text{I}}(\vtheta)
    &=
      \lim_{M \to \infty}
      \frac{R}{M}
      \sum_{n=1}^N
      \jac_{\vtheta} \vf_n^{\top}
      \left[
      -\nabla_{\vf_n} \log r(\rvy = \tilde{\vy}_{n,m} \mid \rvf = \vf_n)
      (-\nabla_{\vf_n} \log r(\rvy = \tilde{\vy}_{n,m} \mid \rvf = \vf_n))^{\top}
      \right]
      \jac_{\vtheta} \vf_n
    \\
    \mE(\vtheta)
    &=
      R
      \sum_{n=1}^N
      (\nabla_{\vtheta} c(\vf_n, \vy_n))
      (\nabla_{\vtheta} c(\vf_n, \vy_n))^{\top}
  \end{align*}
  \begin{itemize}
  \item In expectation notation (coloured parts coincide for the above criterion functions, hence GGN = Fisher)
    \begin{align*}
      \hess_{\vtheta} \gL_{\sD}(\vtheta)
      &\propto
        \E_{p_{\sD}(\rvx)}
        \E_{p_{\sD}(\rvy \mid \rvx)}
        \left[
        -\hess_{\vtheta} \log r(\rvy \mid \rvf)
        \right]
      \\
      \mF(\vtheta)
      &\propto
        \E_{p_{\sD}(\rvx)}
        \E_{r(\rvy \mid \rvf)}
        \left[
        -\hess_{\vtheta} \log r(\rvy \mid \rvf)
        \right]
      \\
      \mG(\vtheta)
      &\propto
        \E_{p_{\sD}(\rvx)}
        \left[
        \jac_{\vtheta} \rvf^{\top}
        \textcolor{VectorBlue}{
        \E_{p_{\sD}(\rvy \mid \rvx)}
        \left[
        -\hess_{\rvf} \log r(\rvy \mid \rvf)
        \right]
        }
        \jac_{\vtheta} \rvf
        \right]
      \\
      \mF^{\text{II}}(\vtheta)
      &\propto
        \E_{p_{\sD}(\rvx)}
        \left[
        \jac_{\vtheta} \rvf^{\top}
        \textcolor{VectorBlue}{
        \E_{r(\rvy \mid \rvf)}
        \left[
        -\hess_{\rvf} \log r(\rvy \mid \rvf)
        \right]
        }
        \jac_{\vtheta} \rvf
        \right]
      \\
      \mF^{\text{I}}(\vtheta)
      &\propto
        \E_{p_{\sD}(\rvx)}
        \left[
        \jac_{\vtheta} \rvf^{\top}
        \textcolor{VectorBlue}{
        \E_{r(\rvy \mid \rvf)}
        \left[
        -\nabla_{\rvf} \log r(\rvy \mid \rvf)
        (
        -\nabla_{\rvf} \log r(\rvy \mid \rvf)
        )^{\top}
        \right]
        }
        \jac_{\vtheta} \rvf
        \right]
      \\
      \mE(\vtheta)
      &\propto
        \E_{p_{\sD}(\rvx)}
        \left[
        \jac_{\vtheta} \rvf^{\top}
        \E_{p_{\sD}(\rvy \mid \rvx)}
        \left[
        -\nabla_{\rvf} \log r(\rvy \mid \rvf)
        (
        -\nabla_{\rvf} \log r(\rvy \mid \rvf)
        )^{\top}
        \right]
        \jac_{\vtheta} \rvf
        \right]
    \end{align*}
  \end{itemize}
\item Gradients (\Cref{ex:square-loss-gradient,ex:cross-entropy-loss-gradient}), Hessians (\Cref{ex:hessian-crossentropyloss,ex:square_loss_hessian}), and symmetric Hessian decompositions (\Cref{ex:mseloss_hessian_factorization,ex:crossentropyloss_hessian_factorization}) of criterion functions
  \begin{align*}
    &
      \qquad&\nabla_{\vf} c(\vf, \vy)
              \qquad&\hess_{\vf} c(\vf, \vy)
                      \qquad&\mS,\, \mS \mS^{\top} = \hess_{\vf} c(\vf, \vy)
    \\
    &
      \begin{matrix}
        \text{Square loss}\\
        \text{(\texttt{MSELoss})}
      \end{matrix}
    & \vf - \vy
      \qquad& \mI
              \qquad& \mI
    \\
    &
      \begin{matrix}
        \text{Softmax cross-entropy loss}\\
        (\text{\texttt{CrossEntropyLoss}})\\
        (\vsigma = \softmax(\vf))
      \end{matrix}
      \qquad& \vsigma - \onehot(y)
              \qquad& \diag(\vsigma) - \vsigma \vsigma^{\top}
                      \qquad& \diag(\sqrt{\vsigma}) - \vsigma \sqrt{\vsigma}^{\top}
  \end{align*}
\end{itemize}

\end{paracol}
\clearpage

\begin{paracol}{2}
  \section{KFAC: An Overview}\label{sec:kfac-overview}
  \switchcolumn[1]*
\codeblock{kfac/scaffold}
\switchcolumn[0]

KFAC has many nuances that complicate its understanding.
However, we can break down its structure into a general scaffold that simplifies its complexity into manageable sub-tasks (\Cref{kfac/scaffold}).
Here, we discuss the components.

\paragraph{Outline.} At its core, KFAC approximates curvature matrices using Kronecker products, significantly reducing computational and memory costs.
In this section, we start from the general form of relevant curvature matrices, discuss how their structure enables a Kronecker-factored approximation, and introduce the core components of KFAC.
This scaffold---accompanied by the code snippets on the right---serves as the foundation for a systematic implementation, offering intuition on how Kronecker factors are computed and why such a decomposition is beneficial.
We keep the discussion general and highlight how to adapt the code to approximate any curvature matrix introduced in the previous section.
The next section (\Cref{sec:kfac-linear}) will focus on the specific case of KFAC for the generalized Gauss-Newton (GGN) of linear layers.

\paragraph{Curvature matrix structure.} Many relevant matrices---like the GGN, the type-I, type-II, and empirical Fisher---share a common structure
\begin{align}\label{eq:common-structure}
  \begin{split}
    \!\!\mC&(\vtheta^{(i)}) \\
           &= R \sum_n
             (\jac_{\vtheta^{(i)}} \vf_n)^{\top}
             \left[ \bullet(\vf_n, \vy_n) \right]
             (\jac_{\vtheta^{(i)}} \vf_n)\,
  \end{split}
\end{align}
where $\bullet \in \sR^{\dim(\gF) \times \dim(\gF)}$ is a positive semi-definite matrix in the prediction space that depends on the prediction $\vf_n$ and target $\vy_n$.
This term is sandwiched between the Jacobians of the network's prediction with respect to the parameters $\vtheta^{(i)}$ in layer $i$.
Directly computing and storing these matrices is usually intractable, which motivates approximation schemes such as KFAC.
\paragraph{Key idea.} The idea behind KFAC is to exploit a Kronecker-product structure in the Jacobians $\jac_{\vtheta^{(i)}}\vf_n$ to approximate
\begin{align*}
  \mC(\vtheta^{(i)})
  \approx
  \kfac(\mC(\vtheta^{(i)}))
  \coloneqq \mA^{(i)} \otimes \mB^{(i)}.
\end{align*}
For a composition of functions $$f \coloneqq f^{(L)} \circ f^{(L-1)} \circ \dots \circ f^{(1)}$$ with parameters $$\vtheta = \left[ \vtheta^{(1)}, \dots, \vtheta^{(L)} \right],$$ KFAC yields a block-diagonal approximation of the full curvature matrix $\mC(\vtheta)$, where each block corresponds to a layer and is of the form $\kfac(\mC(\vtheta^{(i)})) = \mA^{{(i)}} \otimes \mB^{(i)}$.

In this approximation, $\mA^{(i)}$ is computed from the inputs to layer $i$, and we refer to it as \emph{input-based Kronecker factor}.
Similarly, $\mB^{(i)}$ is computed from gradients \wrt layer $i$'s output, and we call it the \emph{grad-output-based Kronecker factor}.

\subsection{Why Use a Kronecker Structure?}
Having a Kronecker approximation for $\mC(\vtheta^{(i)})$ has multiple advantages and motivations that coincide well with the block diagonal structure.

\subsubsection{Kronecker Products Are Computationally Efficient}
\label{sec:mem_comp_eff_kron}
Instead of working with the dense representation
\begin{align*} \mA \otimes& \mB  \in \R^{n_1m_1 \times n_2m_2} \\
                          &=\begin{bmatrix}
                            a_{1,1} & \dots & a_{1,n_2} \\
                            \vdots & \ddots & \vdots \\
                            a_{n_1,1} & \dots & a_{n_1,n_2}
                          \end{bmatrix}
  \\
                          &\otimes
                            \begin{bmatrix}
                              b_{1,1} & \dots & b_{1,m_2} \\
                              \vdots & \ddots & \vdots \\
                              b_{m_1,1} &  \dots & b_{m_1,m_2}
                            \end{bmatrix} \\
                          &=\begin{bmatrix}
                            a_{1,1} \mB & \dots & a_{1,n_2} \mB \\
                            \vdots & \ddots & \vdots \\
                            a_{n_1,1} \mB & \dots & a_{n_1,n_2} \mB
                          \end{bmatrix}\,,
\end{align*}
we can express many operations with the Kronecker factors $\mA \in \sR^{n_1 \times n_2}$ and $\mB \in \sR^{m_1 \times m_2}$.
This drastically reduces memory, as we only need to store and handle these smaller matrices (\ie, $\mathcal{O}(n_1n_2 + m_1m_2)$) rather than the full Kronecker product (\ie $\mathcal{O}(n_1n_2m_1m_2)$).
Examples are:

\paragraph{Matrix-vector products.}
Let $\vv \in \R^{n_2m_2}$, then
$$ (\mA \otimes \mB) \vv = \cvec(\mB \mV \mA^{\top}) $$
with $\mV = \cvec^{-1}(\vv) \in \R^{n_2\times m_2}$.
Similarly,
$$ (\mA \otimes \mB) \vv = \rvec(\mA \mV \mB^{\top}) $$
with $\mV = \rvec^{-1}(\vv) \in \R^{m_2\times n_2}$.

\paragraph{Matrix transpose.} $(\mA \otimes \mB)^{\top} = \mA^{\top} \otimes \mB^{\top}$.

\paragraph{Matrix inverse.} $(\mA \otimes \mB)^{-1} = \mA^{-1} \otimes \mB^{-1}$ (assuming the Kronecker factors are invertible).

\paragraph{Matrix multiplication.}
Let $\mC \in \R^{n_2 \times d_1}$ and $\mD \in \R^{m_2 \times d_2}$, then we have
$$ (\mA \otimes \mB)(\mC \otimes \mD) = \mA \mC \otimes \mB \mD\,, $$ \ie we can multiply the Kronecker factors.

\subsubsection{Kronecker Products Naturally Emerge in Curvature Matrices}

Kronecker structure arises naturally in the expression for $\mC(\vtheta^{(i)})$ due to the structure of the layer Jacobians, providing a strong motivation for its use.
To see this, we rewrite $\mC(\vtheta^{(i)})$ from \Cref{eq:common-structure}.

\paragraph{Factorizing the loss Hessian.}
Since $\bullet(\vf_n, \vy_n)$ is a positive semi-definite matrix, it can be decomposed as a sum of outer products of at most $\dim(\gF)$ vectors: $$\bullet(\vf_n, \vy_n) = \sum_{c=1}^{\dim(\gF)} \blacktriangle_c(\vf_n, \vy_n) (\blacktriangle_c(\vf_n, \vy_n))^{\top}$$
where we define $\blacktriangle_{n,c} := \blacktriangle_c(\vf_n, \vy_n) \in \sR^{\dim(\gF)}$ as a vector we will specify later.
Substituting this into \Cref{eq:common-structure}, we obtain
\begin{align*}
  \mC(&\vtheta^{(i)}) \\
  =&
     R \sum_n \sum_{c}
     (\jac_{\vtheta^{(i)}} \vf_n)^{\top}
     \blacktriangle_{n,c} \blacktriangle_{n,c}^{\top}
     (\jac_{\vtheta^{(i)}} \vf_n)\,.
\end{align*}

\paragraph{Separate the layer's Jacobian.}
By applying the chain rule to split the Jacobian at the layer output $\vx^{(i)} = f^{(i)}(\vx^{(i-1)}, \vtheta^{(i)})$, we get
\begin{align*}
  (\jac_{\vtheta^{(i)}} \vf_n)^{\top}
  =
  (\jac_{\vtheta^{(i)}} \vx^{(i)}_n)^{\top}
  (\jac_{\vx_n^{(i)}} \vf_n)^{\top}.
\end{align*}
Substituting this back, the curvature matrix is
\begin{align*}
  \mC(\vtheta&^{(i)}) \\ =& R \sum_n \sum_{c}
                          &&(\jac_{\vtheta^{(i)}} \vx^{(i)}_n)^{\top} \\
             & &&(\jac_{\vx_n^{(i)}} \vf_n)^{\top}
                  \blacktriangle_{n,c}
                  \blacktriangle_{n,c}^{\top}
                  (\jac_{\vx_n^{(i)}} \vf_n) \\
             & &&(\jac_{\vtheta^{(i)}} \vx^{(i)}_n).
\end{align*}

\switchcolumn[1]*
\begin{example}[Curvature matrix of a linear layer as sum of Kronecker products]\label{ex:curvature-matrix-sum-of-kronecker-products}
  Let's illustrate the statement that the curvature matrix becomes a sum of Kronecker products, where one factor stems from the layers Jacobian and one from a backpropagated vector, for a linear layer with weight matrix $\mW$, and using $\cvec$ flattening.
  Let's define the shorthand $\vg_{n,c}^{{(i)}} \coloneqq (\jac_{\vx_n^{(i)}} \vf_n)^{\top} \blacktriangle_{n,c}$.
  Then, inserting the layer's Jacobian from \Cref{ex:linear_layer_jacobians}, we get
  \begin{align*}
      \mC^{(i)}(\cvec \mW)
      &=
      R \sum_{n,c}
      \left(
      {\vx_n^{(i-1)}}^{\top}\!\!\! \otimes \mI
      \right)^\top
      \!\!\!\vg_{n,c}^{(i)}
      {\vg_{n,c}^{(i)}}^\top
      \left(
      {\vx_n^{(i-1)}}^{\top} \!\!\!\otimes \mI
      \right)\!\!
      \\
      &=
      R \sum_{n,c}
      \left(
      \vx_n^{(i-1)} \otimes \vg_{n,c}^{(i)}
      \right)
      \left(
      {\vx_n^{(i-1)}}^\top \otimes {\vg_{n,c}^{(i)}}^\top
      \right)
      \\
      &=
      R \sum_{n,c}
      \vx_n^{(i-1)} {\vx_n^{(i-1)}}^\top
      \otimes
      \vg_{n,c}^{(i)} {\vg_{n,c}^{(i)}}^\top\,
  \end{align*}
  simply by using the Kronecker product's properties from \Cref{sec:mem_comp_eff_kron}.
  This illustrates that the exact curvature matrix indeed becomes a sum of Kronecker products.
\end{example}
\switchcolumn[0]

\paragraph{Insert the layer's Jacobian.}
The Kronecker structure emerges naturally in the output-parameter Jacobian $\smash{\jac_{\vtheta^{(i)}} \vx_n^{(i-1)}}$ of a layer.
We saw this in previous chapters for a linear layer (\Cref{ex:linear_layer_jacobians}), whose Jacobian is a Kronecker product of the layer input $\smash{\vx^{(i-1)}_n}$ and an identity matrix.
Inserting the Jacobian, this reduces the \emph{exact} form of $\mC^{(i)}$ to a sum of Kronecker products (fully worked out in \Cref{ex:curvature-matrix-sum-of-kronecker-products}),
\begin{align*}
  \mC^{(i)} = R \sum_n \sum_c \va_n \va_n^{\top} \otimes \vb_{n,c} \vb_{n,c}^{\top}
\end{align*}
where $\va_n$ stems from the layer's Jacobian, \ie the layer's input $\vx_n^{(i-1)}$, and $\vb_{n,c}$ stems from the vector $\blacktriangle_{n,c}$ that is backpropagated to the layer's output by applying the Jacobian $(\jac_{\vx_n^{(i)}} \vf_n)^{\top}$.

Our remaining steps are (i) to specify what $\va_n$ and $\vb_{n,c}$ are, which will depend on the curvature matrix we want to approximate (\Cref{subsec:backpropagated-vectors}), and (ii) how to approximate the sum of Kronecker products as a single Kronecker product (KFAC's expectation approximation, \Cref{def:kfac_exp_approx}).

\switchcolumn[1]*
\codeblock{kfac/backpropagated_vectors}
\switchcolumn[0]

\subsubsection{Kronecker Factor Dependencies}

For the scaffold in \Cref{kfac/scaffold}, let's quickly think about the dependencies of the Kronecker matrices in the above expression.
KFAC results in two distinct Kronecker factors: one based on layer inputs (input-based Kronecker factor $\mA^{(i)}$) and the other based on gradients with respect to the layer output (grad-output-based Kronecker factor $\mB^{(i)}$).
As we will see, we can interpret $(\jac_{\vtheta^{(i)}} \vx_n^{(i)})^{\top} \blacktriangle_{n,c}$ as a \emph{pseudo-gradient}.
In fact, setting
$$\blacktriangle_{n,c} \leftarrow \nabla_{\vf_n} c(\vf_n, \vy_n)\,$$
we obtain
$$(\jac_{\vx_n^{(i)}} \vf_n)^{\top} \blacktriangle_{n,c} = \nabla_{\vx_n^{(i)}} c(\vf_n, \vy_n),$$
which is the per-datum loss gradient with respect to layer $i$'s output.
In summary, we identify the following dependencies of the Kronecker factors $\mA^{(i)}$ and $\mB^{(i)}$, justifying their names:
\begin{align*}
  \mA^{(i)} &\text{\quad depends on \quad} \{\vx_{n}^{(i-1)}\}_n \,,
  \\
  \mB^{(i)} &\text{\quad depends on \quad} \{ (\jac_{\vx_n^{(i)}}\vf_{n})^{\top} \blacktriangle_{n,c}\}_{n,c}\,.
\end{align*}

\subsection{Algorithmic Outline}

We now know enough to discuss the scaffold from \Cref{kfac/scaffold}.

\paragraph{Step 1:} Perform a forward pass to compute $\{\vf_n\}_n $, storing the layer inputs $\{\vx_n^{(i-1)} \}_n$ and outputs $\{\vx_n^{(i)}\}_n$ of all layers $i$ whose parameter curvature matrix we want to approximate with KFAC.

\paragraph{Step 2:} Compute the input-based Kronecker factors $\mA^{(i)}$ using the layer inputs $\{\vx_n^{(i-1)}\}_n$.
The details of that will depend on the layer type, and we still have to specify them in later chapters.

\paragraph{Step 3:} Generate the vectors $\{\blacktriangle_{n,c}\}_{n,c}$ to be backpropagated, and backpropagate them to the layer outputs to get the pseudo-gradients $\{(\jac_{\vx_n^{(i)}} \vf_n)^{\top} \blacktriangle_{n,c} \}_{n,c}$.
This step depends on the loss function we are using, and the curvature matrix we want to approximate (\Cref{subsec:backpropagated-vectors}).
Finally, compute the output-based Kronecker factors $\mB^{(i)}$.
The details will again depend on the layer type, and we will specify them in later chapters.

\paragraph{Step 4:} Account for scaling caused by the loss function's reduction $R$.

\paragraph{Step 5:} Return the KFAC approximation in the form $\mA^{(i)}, \mB^{(i)}$ for all supported layers $i$.

In summary, the scaffold disentangles the task of computing KFAC into three components: (i) computing the input-based Kronecker factors, (ii) generating and backpropagating the vectors $\blacktriangle_{n,c}$, and (iii) computing the grad-output-based Kronecker factors.
The next section describes how to accomplish step (ii).
After that, we can add new KFAC implementations simply by specifying (i) and (iii) for each new layer.

\subsection{Backpropagated Vectors}\label{subsec:backpropagated-vectors}
The computational scaffold can be flexibly adapted to various curvature matrices by modifying the choice of backpropagated vectors $\blacktriangle_{n,c}$ and the reduction factor $R$.
This can be done by pattern-matching each curvature matrix to the general form in \Cref{eq:common-structure}.
Below, we recap the different curvature definitions; see \Cref{subsec:curvature-matrices} for a self-contained introduction.
\Cref{kfac/backpropagated_vectors} implements the construction of backpropagated vectors.

\paragraph{Generalized Gauss-Newton/type-II Fisher.} Remember the GGN from \Cref{sec:partial_linearization},
\begin{align*}
  \mG(\vtheta)= R \sum_n
  \left[\jac_{\vtheta} \vf_n\right]^{\top}
  \left[\hess_{\vf_n} c(\vf_n, \vy_n)
  \right]
  \left[\jac_{\vtheta} \vf_n\right]\,.
\end{align*}
With the symmetric decomposition (\Cref{subsec:curvature-ggn})
$$\mS_n \mS_n^{\top} = \hess_{\vf_n} c(\vf_n, \vy_n),$$ we identify for the backpropagated vector
\begin{align*}
  \blacktriangle_{n,c} = [\mS_n]_{:,c}.
\end{align*}
In words, to approximate the GGN with KFAC, we must backpropagate each of the $\dim(\gF)$ columns of the Hessian decomposition $\mS_n$.
For common loss functions, the Type-II Fisher from \Cref{sec:fisher} coincides with the GGN, hence we can use the same backpropagated vectors as for the GGN to approximate it.

\paragraph{MC-sampled type-I Fisher.}
The type-I Fisher, approximated via Monte Carlo (MC) sampling, is
\begin{align*}
  &\begin{aligned}
    \mF^{\text{I}}(\vtheta) = \lim_{M \to \infty} \frac{R}{M} \sum_{n=1}^N\sum_{m=1}^M
    &\left( \jac_{\vtheta} \vf_n\right)^{\top} \\
    &
      \begin{aligned}
        \big[
        &\nabla_{\vf_n} c(\vf_n, \tilde{\vy}_{n,m}) \\
        &\nabla_{\vf_n} c(\vf_n, \tilde{\vy}_{n,m})^{\top}
          \big]
      \end{aligned} \\
    &\jac_{\vtheta} \vf_n\,,
  \end{aligned}
\end{align*}
(\Cref{sec:fisher}).
We can immediately see that
\begin{align*}
  \blacktriangle_{n,m}
  &= \nabla_{\vf_n}  c(\vf_n, \tilde{\vy}_{n,m})
\end{align*}
where $\tilde{\vy}_{n,m} \stackrel{\text{\iid}}{\sim} r(\rvy \mid \vf = \vf_n)$ is a sample from the model's predictive distribution.
In words, to approximate the Type-I Fisher with KFAC, we backpropagate $M$ `would-be' gradients for each datum $\vf_n$.
The number of Monte Carlo samples controls the number of backpropagations:
\begin{itemize}
\item For \emph{computational efficiency}, $M<\dim(\gF)$ makes sense because this reduces the number of backpropagations from $\dim(\gF)$ (as in the type-II case) to $M$.
\item For \emph{practical settings}, $M$ is usually set to $1$.
\item For \emph{verification}, a larger $M$ can be used to ensure convergence to the expected value.
\end{itemize}

\paragraph{Empirical Fisher.}
The empirical Fisher (\Cref{sec:emp_fisher}) replaces the expectation over the model's likelihood in the type-I Fisher by the expectation over the empirical distribution:
\begin{align*}
  &\begin{aligned}\mE(\vtheta) = R \sum_{n=1}^N
    &\jac_{\vtheta} \vf_n^{\top} \\
                               & \nabla_{\vf_n} c(\vf_n, \tilde{\vy}_{n}) (\nabla_{\vf_n} c(\vf_n, \tilde{\vy}_n ))^{\top} \\
                               &\jac_{\vtheta} \vf_n\,.
  \end{aligned}
\end{align*}
We identify that we only need to backpropagate a single vector per datum,
\begin{align*}
  \blacktriangle_{n,1}
  &= \nabla_{\vf_n}  c(\vf_n, \vy_n)\,.
\end{align*}
It is the same vector that is backpropagated to compute the gradient.
This means that to approximate the empirical Fisher with KFAC, we can recycle the backward pass from the gradient computation.

  \clearpage

  \section{KFAC for Linear Layers (No Weight Sharing)}\label{sec:kfac-linear}
  \switchcolumn[1]*
\begin{figure}[!h]
  \centering
  \begin{minipage}[t]{0.485\linewidth}
    \centering
    \textbf{Full}
  \end{minipage}
  \hfill
  \begin{minipage}[t]{0.485\linewidth}
    \centering
    \textbf{KFAC}
  \end{minipage}
  \\
  \begin{minipage}[t]{0.485\linewidth}
    \centering
    GGN ($\cvec$)\vspace{1ex}
    \includegraphics[width=1.0\linewidth]{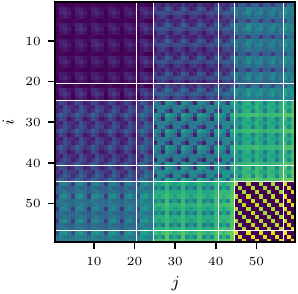}
  \end{minipage}
  \hfill
  \begin{minipage}[t]{0.485\linewidth}
    \centering
    GGN ($\cvec$)\vspace{1ex}
    \includegraphics[width=1.0\linewidth]{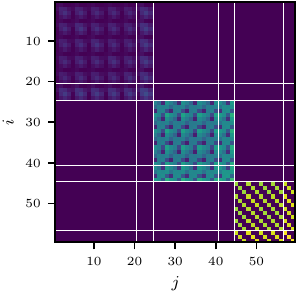}
  \end{minipage}
  \\
  \begin{minipage}[t]{0.485\linewidth}
    \centering
    MC-Fisher ($\cvec$)\vspace{1ex}
    \includegraphics[width=1.0\linewidth]{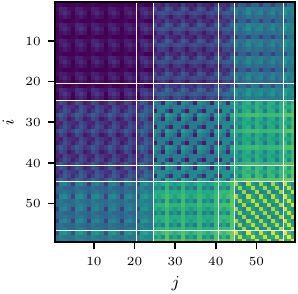}
  \end{minipage}
  \hfill
  \begin{minipage}[t]{0.485\linewidth}
    \centering
    MC-Fisher ($\cvec$)\vspace{1ex}
    \includegraphics[width=1.0\linewidth]{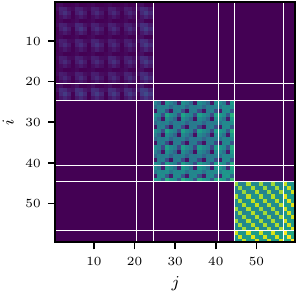}
  \end{minipage}
  \\
  \begin{minipage}[t]{0.485\linewidth}
    \centering
    Emp-Fisher ($\cvec$)\vspace{1ex}
    \includegraphics[width=1.0\linewidth]{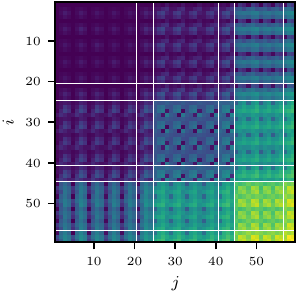}
  \end{minipage}
  \hfill
  \begin{minipage}[t]{0.485\linewidth}
    \centering
    Emp-Fisher ($\cvec$)\vspace{1ex}
    \includegraphics[width=1.0\linewidth]{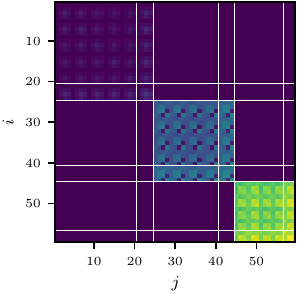}
  \end{minipage}
  \caption{\textbf{Full curvatures and their corresponding KFAC approximation using $\cvec$-flattening}.
    All curvatures were evaluated on synthetic data ($N = 100$) using an MLP with three fully-connected layers and ReLU activations (5-4-4-3) and square loss.
    For the MC-sampled Fisher, we consider $M = 100$ samples.
    This follows the setup in \Cref{fig:hessian-block-structure}.
    KFAC concatenates the weight and bias of each layer.
    Produced with \repofile{plots/synthetic_kfac}.}
  \label{fig:cvec-kfac-full-comparison}
\end{figure}
\switchcolumn[0]

Building on the scaffold and code from the previous chapter, we now introduce, implement, and test the KFAC approximation for the weights (or combined weights and bias) of fully-connected layers (\texttt{torch.nn.Linear}).
See \Cref{fig:rvec-kfac-full-comparison,fig:cvec-kfac-full-comparison} for visualizations.

Our discussion will primarily center on regression settings with deep linear networks---MLPs composed of dense layers without nonlinearities.
This setting provides an ideal framework for understanding the core approximations behind KFAC and verifying them numerically through rigorous testing.
We focus on the formulation from \citet{martens2015optimizing}, which was originally introduced for standard MLP architectures, where linear layers do not exhibit weight sharing.

Other layers, like convolutions and linear layers inside the attention mechanism, exhibit weight sharing.
We deliberately exclude these aspects here and focus on linear layers without weight sharing.
Future versions of this tutorial may include them.

\switchcolumn[1]
\begin{figure}[!h]
  \centering
  \begin{minipage}[t]{0.485\linewidth}
    \centering
    \textbf{Full}
  \end{minipage}
  \hfill
  \begin{minipage}[t]{0.485\linewidth}
    \centering
    \textbf{KFAC}
  \end{minipage}
  \\
  \begin{minipage}[t]{0.485\linewidth}
    \centering
    GGN ($\rvec$)\vspace{1ex}
    \includegraphics[width=\linewidth]{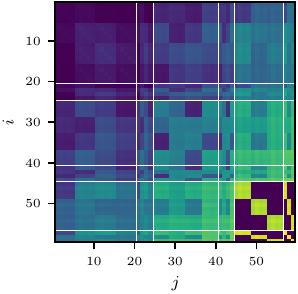}
  \end{minipage}
  \hfill
  \begin{minipage}[t]{0.485\linewidth}
    \centering
    GGN ($\rvec$)\vspace{1ex}
    \includegraphics[width=\linewidth]{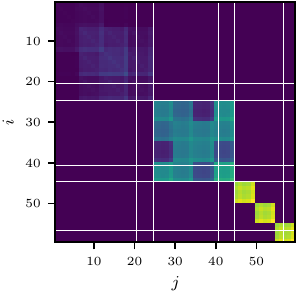}
  \end{minipage}
  \\
  \begin{minipage}[t]{0.485\linewidth}
    \centering
    MC-Fisher ($\rvec$)\vspace{1ex}
    \includegraphics[width=\linewidth]{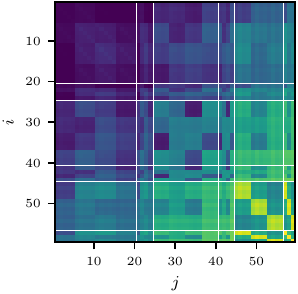}
  \end{minipage}
  \hfill
  \begin{minipage}[t]{0.485\linewidth}
    \centering
    MC-Fisher ($\rvec$)\vspace{1ex}
    \includegraphics[width=\linewidth]{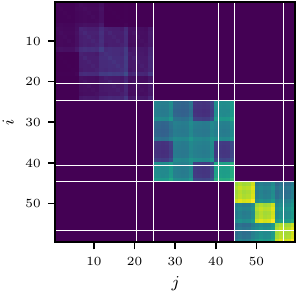}
  \end{minipage}
  \\
  \begin{minipage}[t]{0.485\linewidth}
    \centering
    Emp-Fisher ($\rvec$)\vspace{1ex}
    \includegraphics[width=\linewidth]{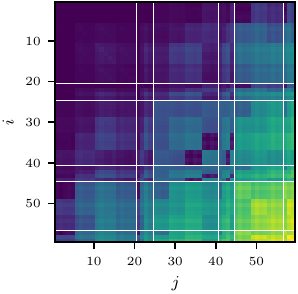}
  \end{minipage}
  \hfill
  \begin{minipage}[t]{0.485\linewidth}
    \centering
    Emp-Fisher ($\rvec$)\vspace{1ex}
    \includegraphics[width=\linewidth]{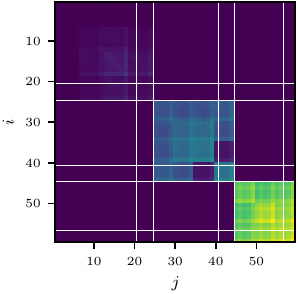}
  \end{minipage}
  \caption{\textbf{Full curvatures and their corresponding KFAC approximation using $\rvec$-flattening.}
    All curvatures were evaluated on synthetic data ($N = 100$) using an MLP with three fully-connected layers and ReLU activations (5-4-4-3) and square loss.
    For the MC-sampled Fisher, we consider $M = 100$ Monte Carlo samples.
    This follows the setup in \Cref{fig:hessian-block-structure}.
    KFAC concatenates the weight and bias of each layer.
    Produced with \repofile{plots/synthetic_kfac}.}
  \label{fig:rvec-kfac-full-comparison}
\end{figure}

\switchcolumn[0]

Let's formalize the layer whose curvature we approximate with KFAC in this chapter:

\begin{setup}[Linear layer inside a neural net]\label{setup:linear_layer}
  Consider a linear layer with weights $\mW \in \sR^{D_{\text{out}}\times D_{\text{in}}}$ and bias $\vb \in \sR^{D_{\text{out}}}$ in a neural net $f(\cdot, \vtheta)$.
  The network's prediction feeds into a criterion function, and we compute an empirical risk over a dataset of $N$ points, incorporating a reduction factor $R$.
  Our goal is to approximate the curvature matrix block $\mC(\vec \mW)$.

  For each datum $n$, the layer processes an input vector $\vx_n \in \sR^{D_{\text{in}}}$, producing an output vector $\vz_{n} \in \sR^{D_{\text{out}}}$ as follows
  $$ \vz_{n} = \mW \vx_{n} + \vb\,.$$
  Denote the network's prediction for datum $n$ by $\vf_n \in \sR^{\dim(\gF)}$.
  For each backprograted vector $\blacktriangle_{n,c} \in \sR^{\dim(\gF)}$, we denote the layer's output gradient as
  $$\vg_{n,c} = (\jac_{\vz_{n}} \vf_n)^{\top} \blacktriangle_{n,c} \in \sR^{D_{\text{out}}}\,.$$

  We will often neglect the bias and focus on approximating the curvature for the weight matrix $\mW$ alone.
  The bias can always be incorporated into the weight matrix by concatenation into a single weight matrix $\tilde{\mW}$ and by augmenting the input vectors with an additional constant:
  \begin{align*}
    \tilde{\mW} &= \begin{pmatrix} \mW & \vb \end{pmatrix} \in \sR^{D_{\text{out}} \times (D_{\text{in}} + 1)}\,, \\
    \tilde{\vx} &= \begin{pmatrix} \vx \\ 1 \end{pmatrix} \in \sR^{D_{\text{in}} + 1}\,.
  \end{align*}
  Thus, the derivations below hold equivalently for $\tilde{\mW}, \tilde{\vx}$ in place of $\mW, \vx$.
\end{setup}

\subsection{Derivation of KFAC}

We now derive the KFAC approximation for a linear layer with weights $\mW$, input vector $\vx$, and output vector $\vz = \mW\vx$ (\Cref{setup:linear_layer}).
Following convention in the literature, we assume column-major ($\cvec$) flattening.

\paragraph{Exact curvature matrix.}
Recall from \Cref{eq:common-structure} that the curvature matrix \wrt $\mW$ takes the form
\begin{align*}
  &\mC(\cvec \mW) \\
  &\begin{aligned}=
    R \sum_n \sum_c &(\jac^{\cvec}_{\mW}\vz_n)^{\top} \\
                    &(\jac^{\cvec}_{\vz_n}\vf_n)^{\top}
                      \blacktriangle_{n,c} \blacktriangle_{n,c}^{\top}
                      (\jac^{\cvec}_{\vz_n}\vf_n) \\
                    &(\jac^{\cvec}_{\mW}\vz_n)\,.
  \end{aligned}
\end{align*}
We recall from \Cref{ex:linear_layer_jacobians} that the Jacobian of a linear layer's output \wrt its weights is
$$ \jac^{\cvec}_{\mW}\vz_n = \vx_n^{\top} \otimes \mI_{D_{\text{out}}}\, $$
where $\vx_n \in \R^{D_\text{in}}$ is the layer's input.
With that, and using the Kronecker product properties from \Cref{sec:mem_comp_eff_kron}, we see that the curvature matrix is a sum of Kronecker products:
\begin{align*}
  &\mC(\cvec \mW)
  \\
  &\begin{aligned}
    = R \sum_n \sum_c &(\vx_n \otimes \mI_{D_{\text{out}}}) \vg_{n,c} \\
                      & \vg_{n,c}^{\top} (\vx_n^{\top} \otimes \mI_{D_{\text{out}}})
  \end{aligned}\\
  &\begin{aligned}
    = R \sum_n \sum_c &(\vx_n \otimes \mI_{D_{\text{out}}}) \vg_{n,c} \\
                      &\left[(\vx_n \otimes \mI_{D_{\text{out}}}) \vg_{n,c} \right]^{\top}
  \end{aligned}\\
  &\begin{aligned}
    = R \sum_n \sum_c &\left( \vx_n \otimes \vg_{n,c} \right)\left( \vx_n \otimes \vg_{n,c} \right)^{\top}
  \end{aligned}\\
  &= R \sum_n \sum_c \left( \vx_n \otimes \vg_{n,c} \right)
    \left( \vx_n^{\top} \otimes \vg_{n,c}^{\top} \right) \\
  &= R \sum_n \sum_c \left(\vx_n \vx_n^{\top}\right) \otimes \left(\vg_{n,c} \vg_{n,c}^{\top}\right)\,.
\end{align*}

\switchcolumn[1]*
\begin{example}[Motivation for KFAC's expectation approximation]
  \label{ex:just_kfac_exp_approx}
  KFAC's expectation approximation can be derived from an optimality condition under specific assumptions to preserve a Kronecker structure.

  Consider the curvature matrix
  \begin{align*}
    \mC
    &=
      \textstyle
      R \sum_{n=1}^N
      (\vx_n \otimes \mI_{D_{\text{out}}})
      \vg_n \vg_n^{\top}
      (\vx_n^{\top} \otimes \mI_{D_{\text{out}}})\,.
  \end{align*}
  Our goal is to approximate $\mC$ with a single Kronecker product.

  To make this expression more compact, stack the vectors $\vx_n$ and $\vg_n$ into matrices:
  \begin{align*}
    \mX
    &=
      \begin{pmatrix}
        \vx_1 & \vx_2 & \ldots & \vx_N
      \end{pmatrix}
      \in \sR^{D_{\text{in}} \times N}\,,
    \\
    \mG
    &=
      \begin{pmatrix}
        \vg_1 & \vg_2 & \ldots & \vg_N
      \end{pmatrix}
      \in \sR^{D_{\text{out}} \times N}\,.
  \end{align*}
  and rewrite $\mC$ in terms of these matrices:
  \begin{align*}
    \mC = R (\mX \otimes \mI_{D_{\text{out}}}) \cvec(\mG) \cvec(\mG)^{\top}(\mX^{\top} \otimes \mI_{D_{\text{out}}}).
  \end{align*}
  Looking at this expression, we realize that an easy way to achieve a single Kronecker product is to approximate $\cvec(\mG) \cvec(\mG)^{\top} \approx \mI_{D_{\text{in}}} \otimes \mB$ with $\mB \in \sR^{D_{\text{out}}\times D_{\text{out}}}$.
  Then we would have
  \begin{align*}
    \mC
    &\approx
      R (\mX \otimes \mI_{D_{\text{out}}})
      (\mI_{D_{\text{in}}} \otimes \mB)
      (\mX^{\top} \otimes \mI_{D_{\text{out}}})
    \\
    &=
      R \mX \mX^{\top} \otimes \mB
      =
      \textstyle
      R \left(\sum_{n=1}^N \vx_n \vx_n^{\top}\right) \otimes \mB\,.
  \end{align*}
  To obtain $\mB$, we minimize the squared Frobenius norm residual:
  \begin{align*}
    \begin{split}
      &\argmin_{\mB}
        \left\lVert
        \cvec(\mG) \cvec(\mG)^{\top} - \mI_{D_{\text{in}}} \otimes \mB
        \right\rVert_2^2.
    \end{split}
    \intertext{Expanding the Kronecker structure, this is equivalent to}
    \begin{split}
      &\argmin_{\mB}
        \left\lVert
        \begin{pmatrix}
          \vg_1 \vg_1^{\top} -\mB & \dots & \vg_1 \vg_N^{\top} \\
          \vdots & \ddots & \vdots \\
          \vg_N \vg_1^{\top} & \dots & \vg_N \vg_N^{\top} - \mB \\
        \end{pmatrix}
        \right\rVert_2^2
      \\
      &=
        \argmin_{\mB}
        \left\lVert
        \begin{pmatrix}
          \vg_1 \vg_1^{\top} - \mB & \dots & \vzero \\
          \vdots & \ddots & \vdots \\
          \vzero & \hdots & \vg_N \vg_N^{\top} - \mB \\
        \end{pmatrix}
        \right\rVert_2^2
      \\
      &=
        \argmin_{\mB}
        \sum_{n=1}^N
        \left\lVert
        \vg_n \vg_n^{\top} - \mB
        \right\rVert_2^2
      \\
      &=
        \argmin_{\mB}
        \sum_{n=1}^N
        \sum_{i,j = 1}^{D_{\text{out}}}
        \left(
        [\vg_n]_i [\vg_n]_j - [\mB]_{i,j}
        \right)^2 \,.
    \end{split}
  \end{align*}
  Taking the derivative \wrt $\mB$ and setting it to zero yields for all $k, l \in \{1, \dots, D_{\text{out}}\}$,
  \begin{align*}
    &\frac{\partial \left(
      \sum_{n=1}^N
      \sum_{i,j = 1}^{D_{\text{out}}}
      \left(
      [\vg_n]_i [\vg_n]_j - [\mB]_{i,j}
      \right)^2
      \right)
      }{
      \partial [\mB]_{k,l}
      }
    \\
    &=
      2 \sum_{n=1}^N
      \left(
      [\mB]_{k,l} -  [\vg_n]_k [\vg_n]_l
      \right)
    \\
    &=
      2 \left( N[\mB]_{k,l} - \sum_{n=1}^N [\vg_n]_k [\vg_n]_l \right)
      \stackrel{!}{=} 0
    \\
    &\implies
      \mB = \frac{1}{N} \sum_{n=1}^N \vg_n \vg_n^{\top}
      = \frac{1}{N} \mG \mG^{\top}.
  \end{align*}
  Thus, under this optimality condition, the best choice for $\mB$ is the empirical covariance of the gradient vectors:
  \begin{align*}
    \mC
    \approx
    \left(R \sum_n \vx_n \vx_n^\top \right)
    \otimes
    \left(\frac{1}{N} \sum_{n'} \vg_{n'} \vg_{n'}^\top \right).
  \end{align*}
  This is precisely the expectation approximation used in KFAC.
\end{example}
\switchcolumn[0]

\paragraph{From sum of Kronecker products to single Kronecker product.}
The last step is to approximate the sum of Kronecker products by a single Kronecker product.
To do that, we use the so-called KFAC's expectation approximation (see \Cref{ex:just_kfac_exp_approx} for further motivation):

\begin{definition}[KFAC's expectation approximation]
  \label{def:kfac_exp_approx}
  Consider two sets of $N$ vectors $\{\vx_{n}\}_{n=1}^N$ and $\{\vg_{n}\}_{n=1}^N$ where each $\vx_n$ is input-based and each $\vg_n$ is output-gradient-based.
  KFAC approximates the sum of Kronecker products of their self-outer products by a single Kronecker product, using the following expectation approximation:
  \begin{subequations}\label{eq:expectation_approximation}
    \begin{align}
      \begin{split}
        &\sum_{n=1}^{\textcolor{VectorOrange}{N}} \vx_n\vx_n^{\top} \otimes \vg_n \vg_n^{\top}
        \\
        &\qquad\approx
          \left( \sum_{n=1}^N \vx_n \vx_n^{\top} \right)
          \otimes
          \frac{
          \left( \textcolor{VectorOrange}{\sum_n} \vg_n \vg_n^{\top} \right)
          }{\textcolor{VectorOrange}{N}}.
      \end{split}
      \intertext{In words, we sum the outer products on the left and right of the Kronecker product independently and divide by the number of outer products in the term containing the gradient vectors.
      In expectation notation (expectation meaning averaging over all values of $n$, \ie $\E_n[\bullet_n] = \nicefrac{1}{N} \sum_{n=1}^N \bullet_n$), we can write this approximation more conveniently as}
      \begin{split}
        &\E_n \left[ \vx_n \vx_n^{\top} \otimes \vg_n \vg_n^{\top} \right]
        \\
        &\qquad\approx N
          \E_n \left[ \vx_n \vx_n^{\top} \right]
          \otimes
          \E_n \left[ \vg_n \vg_n^{\top} \right]\,,
      \end{split}
    \end{align}
  \end{subequations}
  \ie `pull' the expectation into the Kronecker factors.
\end{definition}

Applying \Cref{def:kfac_exp_approx}, we obtain the curvature matrix's KFAC approximation:
\begin{align*}
  \mC&(\cvec \mW) \\
  \approx&
           \underbrace{\left( R \sum_n \vx_n \vx_n^{\top} \right)}_{\text{input-based}}
           \otimes
           \underbrace{\left( \frac{1}{N} \sum_n \sum_c \vg_{n,c} \vg_{n,c}^{\top} \right)}_{\text{grad-output-based}}
  \\
  =& \kfac(\mC(\cvec \mW))\,.
\end{align*}

\paragraph{Changing flattening swaps the Kronecker factors.}
Had we done the same derivation using the $\rvec$-flattening scheme where the output-parameter Jacobian takes the form (\Cref{ex:linear_layer_jacobians})
$$ \jac^{\rvec}_{\mW}\vz_n = \mI_{D_{\text{out}}} \otimes \vx_n^{\top}\,,$$
we would have found the order of Kronecker factors to reverse,
\begin{align*}
  \mC&(\rvec \mW) \\
  \approx&
           \left( \frac{1}{N} \sum_n \sum_c \vg_{n,c} \vg_{n,c}^{\top} \right)
           \otimes
           \left( R \sum_n \vx_n \vx_n^{\top} \right)
  \\
  =& \kfac(\mC(\rvec \mW))\,.
\end{align*}
This is important to keep in mind for KFAC implementations, as libraries like PyTorch use $\rvec$-flattening.
In summary, we have the following:

\switchcolumn[1]
\codeblock{kfac/expand_Linear}
\switchcolumn[0]

\begin{definition}[KFAC for a linear layer without weight sharing, \Cref{kfac/expand_Linear}]\label{def:kfac_expand_linear}
  Consider a linear layer inside a neural network from \Cref{setup:linear_layer}.
  The \emph{KFAC approximation} of a curvature matrix \wrt the layer's weights is given by
  \begin{subequations}\label{eq:kfac_expand_linear}
    \begin{align}
      \begin{split}
        &\kfac(\vec \mW) \approx \mC(\vec \mW)
        \\
        &\quad=
          \begin{cases}
            \mA \otimes \mB & \vec = \cvec
            \\
            \mB \otimes \mA & \vec = \rvec
          \end{cases}\,,
      \end{split}
    \end{align}
    with the input-based factor
    \begin{align}
      \mA &= R \sum_{n=1}^N \vx_{n} \vx_{n}^{\top}\in \sR^{D_{\text{in}} \times D_{\text{in}}}
                        \intertext{and the output-gradient-based factor}
                        \mB &= \frac{1}{N}\sum_{n=1}^N \sum_c \vg_{n,c} \vg_{n,c}^{\top}\in \sR^{D_{\text{out}} \times D_{\text{out}}}
    \end{align}
  \end{subequations}
  and $\vg_{n,c}$ the backpropagated vector of a curvature matrix at layer output $\vz$ given by $$\vg_{n,c} = (\jac_{\vz_{n,s}}^{\vec}\vf_n)^{\top} \blacktriangle_{n,c}\,.$$
\end{definition}

\subsection{Tests}

To verify our KFAC implementation, we need tests.
This raises an important question: When does the approximation in \Cref{eq:expectation_approximation} become exact?
In general, KFAC's expectation approximation becomes exact whenever one of $\{\vx_n\}_n$ or $\{\vg_{n,c}\}_{n,c}$ is data-independent, \ie, does not depend on $n$.
There are two notable cases where this happens, one of which is relatively obvious.

\switchcolumn[1]
\codeblock{kfac_tests/mlp_batch_size_1}
\switchcolumn[0]

\subsubsection{Data Set Size 1}
A particularly straightforward case occurs when summation over $n$ disappears, \ie when the dataset is only a single data point or $N$ identical data points.
Hence, we have the following

\begin{test}[KFAC for linear layers in an MLP with one data point, \Cref{kfac_tests/mlp_batch_size_1}]\label{test:kfac_expand_linear_no_weight_sharing_batch_size_1}
  Consider a multi-layer perceptron (MLP)
  \begin{align*}
    f = \phi^{(L)} \circ f^{(L)} \circ \ldots \circ \phi^{(1)} \circ f^{(1)}\,.
  \end{align*}
  Each layer consists of a dense (linear) transformation $f^{(l)}$ (satisfying \Cref{setup:linear_layer}) followed by a pointwise activation $\phi^{(l)}$.
  Let $\mW^{(l)}$, $\vb^{(l)}$, and $\tilde{\mW}^{(l)} = \begin{pmatrix} \mW^{(l)} & \vb^{(l)} \end{pmatrix}$ denote the weight, bias, and concatenated parameters of the dense layer $l$.

  Now, assume that the dataset consists of only one data point, $\sD = \{ (\vx, \vy) \}$, and we have a criterion function $c = - \log r$, which can be interpreted as negative log-likelihood.

  Then, KFAC becomes exact for all layers $l = 1, \dots, L$ and any flattening scheme $\vec$ in the following sense:
  \begin{itemize}[leftmargin=0.5cm]
  \item KFAC-type-II equals the GGN
    \begin{align*}
      \kfac(\mG(\vec \tilde{\mW}^{(l)})) = \mG(\vec \tilde{\mW}^{(l)})
    \end{align*}
  \item KFAC-MC converges to the GGN
    \begin{align*}
      \lim_{M \to \infty} &\kfac(\mF^{\text{I}}(\vec\tilde{\mW}^{(l)}))
      \\
                          &= \mG(\vec \tilde{\mW}^{(l)})
    \end{align*}
  \item KFAC-empirical coincides with the empirical Fisher
    \begin{align*}
      \kfac(\mE(\vec\tilde{\mW}^{(l)})) = \mE(\vec \tilde{\mW}^{(l)})
    \end{align*}
  \end{itemize}
\end{test}

This test serves as a basic functionality check for KFAC, particularly regarding the criterion function and its backpropagated vectors $\blacktriangle_{n,c}$.

\switchcolumn[1]
\codeblock{kfac_tests/deep_linear_regression}
\switchcolumn[0]

\subsubsection{Regression with Deep Linear Nets}

A more subtle and interesting case where the KFAC approximation becomes exact is for deep linear networks with square loss.
In this setting, the backpropagated vectors $\vg_{n,c}$ become independent of $n$ because the network's prediction is linear \wrt each layer's parameter $\mW^{(l)}$ (but not linear \wrt the \emph{full} parameter vector $\vtheta$).

Recall that $\vg_{n,c}$ depends on the backpropagated vector $\blacktriangle_{n,c}$ and the Jacobian of the prediction \wrt the layer's output $\jac_{\vz_n} \vf_n$.
For a square loss function, $\blacktriangle_{n,c}$ is often independent of $n$:
\begin{itemize}
\item In type-II Fisher, it corresponds to a column of the identity matrix.
\item In Monte Carlo (MC) Fisher, it is simply a normally distributed random number.
\end{itemize}
Further, if all layers in the net are linear, then all prediction-intermediate Jacobians remain constant and independent of $n$.
Thus, in this setting, the KFAC approximation becomes exact in the limit, as originally found by \citet{bernacchia2018exact}.

\begin{test}[KFAC for regression with a deep linear net (no nonlinear layers), \Cref{kfac_tests/deep_linear_regression}]
  Consider a deep linear network consisting of $L$ fully-connected layers
  \begin{align*}
    f = f^{(L)} \circ f^{(L-1)} \circ \ldots \circ f^{(1)}
  \end{align*}
  with weights $\mW^{(l)}$, bias $\vb^{(l)}$, and concatenated parameters $\tilde{\mW}^{(l)} = \begin{pmatrix} \mW^{(l)} & \vb^{(l)} \end{pmatrix}$.
  Assume the network processes a vector-valued input per datum and performs regression with square loss on an arbitrary dataset $\sD = \{ (\vx_n, \vy_n) \mid n = 1, \dots, N \}$.
  Then, for each layer $l=1,\dots,L$ and any chosen flattening scheme $\vec$, the following identities hold:
  \begin{itemize}[leftmargin=0.5cm]
  \item KFAC-type-II equals the GGN
    \begin{align*}
      \kfac(\mG(\vec \tilde{\mW}^{(l)}))
      =
      \mG(\vec\tilde{\mW}^{(l)})
    \end{align*}
  \item KFAC-MC converges to the GGN
    \begin{align*}
      \lim_{M \to \infty}& \kfac(\mF^{\text{I}}(\vec \tilde{\mW}^{(l)}))
      \\&=
      \mG(\vec \tilde{\mW}^{(l)})
    \end{align*}

  \end{itemize}
  However, KFAC-empirical \emph{differs} from the empirical Fisher,
  \begin{align*}
    \kfac(\mE(\vec \tilde{\mW}^{(l)}))
    \neq
    \mE(\vec \tilde{\mW}^{(l)})\,,
  \end{align*}
  as the backpropagated vector $\blacktriangle_{n,c}$ depends on $\vf_n$ and therefore remains dependent on $n$.
\end{test}

This test is useful for preventing scaling bugs in the Kronecker factor computations.
Additionally, it remains valid when incorporating an independent sequence dimension $S$, \ie in the presence of weight sharing, provided that the network treats sequence elements independently.
Although weight sharing is not of our interest in this chapter, we have already included such a slightly more general test case in \Cref{kfac_tests/deep_linear_regression}.

\end{paracol}
\clearpage

\begin{paracol}{1}
  \section{Cheatsheet: KFAC}\label{sec:kfac-cheatsheet}
  \begin{itemize}
  \item General KFAC scaffold: \qquad
    $
    \mC(\vtheta^{(i)})
    \approx
    \kfac(\mC(\vtheta^{(i)}))
    \coloneqq \mA^{(i)} \otimes \mB^{(i)}
    $
  \quad with
  \begin{itemize}
    \item $\mC(\vtheta^{(i)})
    = R \sum_n
    (\jac_{\vtheta^{(i)}} \vf_n)^{\top}
    \left[ \bullet(\vf_n, \vy_n) \right]
    (\jac_{\vtheta^{(i)}} \vf_n)$
    \item $\bullet(\vf_n, \vy_n) = \sum_{c=1}^{\dim(\gF)} \blacktriangle_c(\vf_n, \vy_n) (\blacktriangle_c(\vf_n, \vy_n))^{\top}$ \quad (depends on curvature type $\mC$)
    \item $\mA^{(i)} = \mA^{(i)}( \{\vx_{n}^{(i-1)}\}_n )$ \quad (input-based factor)
    \item $\mB^{(i)} = \mB^{(i)}( \{ (\jac_{\vx_n^{(i)}}\vf_{n})^{\top} \blacktriangle_{n,c}\}_{n,c})$ \quad (grad-output-based factor)
  \end{itemize}
  \item Backpropagated vectors $\{ \blacktriangle_{n,c} \}_{n,c}$
    \begin{itemize}
    \item GGN/type-II Fisher: $\blacktriangle_{n,c} = [\mS_n]_{:,c}$ for $c = 1, \dots, C$ where $\mS_n \mS_n^{\top} = \hess_{\vf_n} c(\vf_n, \vy_n)$
      \begin{itemize}
      \item Square loss: $\blacktriangle_{n,c} = [\mI]_{:,c}$  (one-hot vector, does not depend on $n$)
      \item Softmax cross-entropy loss: $\blacktriangle_{n,c} =  \sqrt{[\softmax(\vf_n)]_c} (\onehot(c) - \softmax(\vf_n))$
      \end{itemize}

    \item MC-Fisher: $\blacktriangle_{n,m} = - \nabla_{\vf_n} \log r(\rvy = \tilde{\vy}_{n,m} \mid \rvf = \vf_n)$ where $\tilde{\vy}_{n,m} \stackrel{\text{\iid}}{\sim} r(\rvy \mid \rvf = \vf_n)$
      \begin{itemize}
      \item Square loss: $\blacktriangle_{n,m} = \tilde{\vy}_{n,m} - \vf_n$ where $\vy_{n,m} \stackrel{\text{\iid}}{\sim} \gN(\rvy \mid \vmu = \vf_n, \mSigma = \mI)$\\
        $\Leftrightarrow \blacktriangle_{n,m} = \tilde{\vy}$ where $\tilde{\vy} \stackrel{\text{\iid}}{\sim} \gN(\rvy \mid \mu = \vzero, \mSigma = \mI)$ (does not depend on $n$)
        \item Softmax cross-entropy loss: $\blacktriangle_{n,m} = \softmax(\vf_n) - \onehot(y_{n,m})$ where $y_{n,m} \stackrel{\text{\iid}}{\sim} \gC(\ry \mid \vsigma = \softmax(\vf_n))$
      \end{itemize}

    \item Empirical Fisher: $\blacktriangle_{n,1} = - \nabla_{\vf_n} \log r(\rvy = \vy_n \mid \rvf = \vf_n)$
      \begin{itemize}
      \item Square loss: $\blacktriangle_{n,1} = \vy_n - \vf_n$
      \item Softmax cross-entropy loss: $\blacktriangle_{n,1} = \softmax(\vf_n) - \onehot(y_n)$
      \end{itemize}
    \end{itemize}
  \item KFAC for a linear layer $\vz = \mW \vx + \vb$ without weight sharing (also holds for $(\mW, \vx) \leftrightarrow (\tilde{\mW}, \tilde{\vx})$)
    \begin{align*}
      &\kfac(\mC(\vec \mW)) \approx \mC(\vec \mW)
      =
        \begin{cases}
          \mA \otimes \mB & \vec = \cvec\,,
          \\
          \mB \otimes \mA & \vec = \rvec\,,
        \end{cases}\,
    \end{align*}
    with input-based factor $\mA$ and grad-output-based factor $\mB$, and $\vg_{n,c} = (\jac_{\vz_{n,s}}^{\vec}\vf_n)^{\top} \blacktriangle_{n,c}\,$,
    \begin{align*}
      \mA = R \sum_{n=1}^N \vx_{n} \vx_{n}^{\top} \in \sR^{D_{\text{in}} \times D_{\text{in}}}, \qquad
      \mB &= \frac{1}{N}\sum_{n=1}^N \sum_c \vg_{n,c} \vg_{n,c}^{\top}  \in \sR^{D_{\text{out}} \times D_{\text{out}}}
    \end{align*}
  \item Test cases
    \begin{itemize}
      \item KFAC for linear layers in an MLP (no weight sharing) and a dataset with only one datum ($|\sD| = 1$). For all $l$:
        \begin{itemize}
          \item KFAC-type-II coincides with the GGN/type-II Fisher, \ie,
            \begin{align*}
              \kfac(\mG(\vec\tilde{\mW}^{(l)})) = \mG(\vec \tilde{\mW}^{(l)})
            \end{align*}
          \item KFAC-MC converges to the GGN, \ie,
            \begin{align*}
              \lim_{M \to \infty} \kfac(\mF^{\text{I}}(\vec\tilde{\mW}^{(l)})) = \mG(\vec \tilde{\mW}^{(l)})
            \end{align*}
          \item KFAC-empirical coincides with the empirical Fisher, \ie,
            \begin{align*}
              \kfac(\mE(\vec\tilde{\mW}^{(l)})) = \mE(\vec \tilde{\mW}^{(l)})
            \end{align*}
        \end{itemize}
      \item KFAC for regression with a deep linear network (no weight sharing and no nonlinear layers). For all $l$:
        \begin{itemize}
          \item KFAC-type-II coincides with the GGN/type-II Fisher, \ie,
            \begin{align*}
              \kfac(\mG(\vec \tilde{\mW}^{(l)})) = \mG(\vec \tilde{\mW}^{(l)})
            \end{align*}
          \item KFAC-MC converges to the GGN as $M\rightarrow\infty$, \ie,
            \begin{align*}
              \lim_{M \to \infty} \kfac(\mF^{\text{I}}(\vec \tilde{\mW}^{(l)})) = \mG(\vec \tilde{\mW}^{(l)})\,.
            \end{align*}
          \item KFAC-empirical does \emph{not} equal the empirical Fisher, \ie,
            \begin{align*}
              \kfac(\vec \tilde{\mW}^{(l)})
              \neq
              \mE(\vec \tilde{\mW}^{(l)})\,.
            \end{align*}
        \end{itemize}
      \end{itemize}
    \end{itemize}

  \end{paracol}
\clearpage

\begin{paracol}{1}
  \section{Conclusion \& Outlook}\label{sec:outlook}
  \paragraph{Summary.} In this manuscript, we provided a detailed overview of the seminal KFAC curvature approximation from the ground up, in both math and code.
In doing so, we first developed a collection of automatic differentiation utilities to access derivatives.
They play a crucial role in computing curvature information and helped us make these concepts concrete for common operations used in deep learning, while serving as ground truth to verify our implementation.
Building on these fundamentals, we motivated using Kronecker products to approximate curvature matrices in neural networks, and introduced the KFAC approximation for linear layers in the absence of weight sharing, as originally described by \citet{martens2015optimizing}, showed how to apply it to different curvature matrices, and highlighted tests to verify its correctness.

\emph{We hope this tutorial is a helpful resource for both newcomers to the field who want to learn more about curvature matrices, their approximations, and common pitfalls, as well as experienced researchers who are seeking a pedagogical introduction and implementation they can use as a starting point to prototype their research idea.}

\paragraph{Call for contributions \& future work.} Although we believe that we covered the most essential aspects of KFAC, this tutorial does not address many important, albeit advanced, topics related to KFAC.
We invite anyone to contribute to this fully open-source effort (\href{\repourl}{\texttt{github.com/f-dangel/kfac-tutorial}}) to further improve the tutorial over time.
Specifically, we think the following ideas are promising to include in future versions:
\begin{itemize}
\item \textbf{Eigenvalue-corrected KFAC (EKFAC).}
  \citet{george2018fast} proposed the concept of eigenvalue correction to improve KFAC's approximation quality, and the resulting EKFAC has become the de facto standard for KFAC's application to training data attribution with influence functions~\cite{grosse2023studying,mlodozeniec2025influence}.
  The first step in computing EKFAC is to compute KFAC.
  In the second step, we only keep the space spanned by the Kronecker factors, and introduce a diagonal scaling term (the ``eigenvalues''), whose value is determined by minimizing the Frobenius norm residual between the original curvature matrix and the corrected KFAC approximation.
  This can be done cheaply.

\item \textbf{KFAC for linear layers with weight sharing.}
  We focused on explaining KFAC as introduced in the seminal paper by \citet{martens2015optimizing}, which considers multi-layer perceptrons where each layer processes a single vector per datum (\ie no weight sharing).
  However, modern neural networks like convolutional networks, transformers, or graph neural networks, process sequence-valued data and therefore naturally exhibit weight sharing across spatial or temporal dimensions.
  Depending on how the shared axis is treated when computing the loss, there exist two different KFAC approximations for linear layers with weight sharing, coined by \citet{eschenhagen2023kroneckerfactored}: KFAC-expand and KFAC-reduce.
  KFAC-reduce becomes relevant in scenarios where weight sharing occurs, and the shared dimension is explicitly reduced (\eg, summed or averaged) during the forward pass.
  Unlike KFAC-expand, which treats shared dimensions as independent by flattening them into a batch dimension, KFAC-reduce adjusts for this reduction by aggregating inputs and gradients across the shared dimension before forming Kronecker factors.
  The formalization of \citet{eschenhagen2023kroneckerfactored} allows KFAC to be extended to transformers, convolutional networks (containing the originally proposed method from \citet{grosse2016kroneckerfactored}), and graph neural networks.
  While we already introduced the \emph{expand} specifier in our implementation, we have not yet implemented the \emph{reduce} specifier and its related tests.

\item \textbf{Functional-style KFAC implementation.}
  One essential step for computing KFAC is accessing the inputs and outputs of layers.
  This is easy if the neural net consists entirely of \texttt{torch.nn.Module}s---and we assume this in our implementation---by using PyTorch's hook mechanism to intercept the forward pass and obtain the layer inputs (required for KFAC's input-based Kronecker factor) and outputs (required for KFAC's grad-output-based Kronecker factors).
  This is okay because most architectures used in practice are indeed implemented in a modular fashion.

  However, a functional-style implementation of KFAC, which had the ability to automatically trace and capture the inputs and outputs of linear operations inside a function representing the neural network (similar to KFAC's implementation in JAX \cite{botev2022kfac-jax}), would make KFAC accessible to libraries that adopt a functional style \citep[\eg][]{duffield2025scalable}.
  Such a more general approach is also helpful to incorporate more recent extensions of KFAC, for instance, to training Physics-informed neural networks \cite{dangel2024kroneckerfactored}, whose forward pass is done in Taylor mode arithmetic that differs from the standard forward pass and can therefore not be intercepted with module hooks.
  We believe that a functional-style implementation of KFAC in PyTorch can be achieved using the tracing mechanism of \texttt{torch.fx} \cite{reed2022torch}, which provides a modifiable presentation of the computation graph.
\end{itemize}

\subsection*{Acknowledgements}
We would like to thank Disen Liao for providing feedback on this manuscript.
Resources used in preparing this research were provided, in part, by the Province of Ontario, the Government of Canada through CIFAR, and companies sponsoring the Vector Institute.
Bálint Mucsányi acknowledges his membership in the European Laboratory
for Learning and Intelligent Systems (ELLIS) PhD program and thanks the International Max Planck Research School for Intelligent Systems (IMPRS-IS) for its support.
Runa Eschenhagen is supported by ARM, the Cambridge Trust, and the Qualcomm Innovation Fellowship.

\end{paracol}

{\footnotesize
  \bibliographystyle{icml2024.bst}
  \bibliography{references.bib}

\begin{thebibliography}{34}
\providecommand{\natexlab}[1]{#1}
\providecommand{\url}[1]{\texttt{#1}}
\expandafter\ifx\csname urlstyle\endcsname\relax
  \providecommand{\doi}[1]{doi: #1}\else
  \providecommand{\doi}{doi: \begingroup \urlstyle{rm}\Url}\fi

\bibitem[Bae et~al.(2024)Bae, Lin, Lorraine, and Grosse]{bae2024training}
Bae, J., Lin, W., Lorraine, J., and Grosse, R.~B.
\newblock Training data attribution via approximate unrolling.
\newblock In \emph{Advances in Neural Information Processing Systems
  (NeurIPS)}, 2024.

\bibitem[Benzing(2022)]{benzing2022gradient}
Benzing, F.
\newblock Gradient descent on neurons and its link to approximate second-order
  optimization.
\newblock In \emph{International Conference on Machine Learning (ICML)}, 2022.

\bibitem[Bernacchia et~al.(2018)Bernacchia, Lengyel, and
  Hennequin]{bernacchia2018exact}
Bernacchia, A., Lengyel, M., and Hennequin, G.
\newblock Exact natural gradient in deep linear networks and its application to
  the nonlinear case.
\newblock In \emph{Advances in Neural Information Processing Systems
  (NeurIPS)}, 2018.

\bibitem[Botev \& Martens(2022)Botev and Martens]{botev2022kfac-jax}
Botev, A. and Martens, J.
\newblock {KFAC-JAX}, 2022.
\newblock URL \url{https://github.com/google-deepmind/kfac-jax}.

\bibitem[Brunet(2010)]{brunet2010basics}
Brunet, F.
\newblock Basics on continuous optimization.
\newblock \url{https://www.brnt.eu/phd/node10.html}, 2010.
\newblock Accessed: 2025-03-22.

\bibitem[Dangel et~al.(2020{\natexlab{a}})Dangel, Harmeling, and
  Hennig]{dangel2020modular}
Dangel, F., Harmeling, S., and Hennig, P.
\newblock Modular block-diagonal curvature approximations for feedforward
  architectures.
\newblock In \emph{International Conference on Artificial Intelligence and
  Statistics (AISTATS)}, 2020{\natexlab{a}}.

\bibitem[Dangel et~al.(2020{\natexlab{b}})Dangel, Kunstner, and
  Hennig]{dangel2020backpack}
Dangel, F., Kunstner, F., and Hennig, P.
\newblock {B}ack{PACK}: Packing more into backprop.
\newblock In \emph{International Conference on Learning Representations
  (ICLR)}, 2020{\natexlab{b}}.

\bibitem[Dangel et~al.(2022)Dangel, Tatzel, and Hennig]{dangel2022vivit}
Dangel, F., Tatzel, L., and Hennig, P.
\newblock Vi{V}i{T}: Curvature access through the generalized
  gauss-newton{\textquoteright}s low-rank structure.
\newblock \emph{Transactions on Machine Learning Research (TMLR)}, 2022.

\bibitem[Dangel et~al.(2024)Dangel, Müller, and
  Zeinhofer]{dangel2024kroneckerfactored}
Dangel, F., Müller, J., and Zeinhofer, M.
\newblock Kronecker-factored approximate curvature for physics-informed neural
  networks.
\newblock In \emph{Advances in Neural Information Processing Systems
  (NeurIPS)}, 2024.

\bibitem[Dangel et~al.(2025)Dangel, Eschenhagen, Ormaniec, Fernandez, Tatzel,
  and Kristiadi]{dangel2025position}
Dangel, F., Eschenhagen, R., Ormaniec, W., Fernandez, A., Tatzel, L., and
  Kristiadi, A.
\newblock Position: Curvature matrices should be democratized via linear
  operators.
\newblock \emph{arXiv}, 2025.

\bibitem[Daxberger et~al.(2021)Daxberger, Kristiadi, Immer, Eschenhagen, Bauer,
  and Hennig]{daxberger2021laplace}
Daxberger, E., Kristiadi, A., Immer, A., Eschenhagen, R., Bauer, M., and
  Hennig, P.
\newblock Laplace redux - effortless bayesian deep learning.
\newblock In \emph{Advances in Neural Information Processing Systems
  (NeurIPS)}, 2021.

\bibitem[Duffield et~al.(2025)Duffield, Donatella, Chiu, Klett, and
  Simpson]{duffield2025scalable}
Duffield, S., Donatella, K., Chiu, J., Klett, P., and Simpson, D.
\newblock Scalable bayesian learning with posteriors.
\newblock In \emph{International Conference on Learning Representations
  (ICLR)}, 2025.

\bibitem[Eschenhagen et~al.(2023)Eschenhagen, Immer, Turner, Schneider, and
  Hennig]{eschenhagen2023kroneckerfactored}
Eschenhagen, R., Immer, A., Turner, R.~E., Schneider, F., and Hennig, P.
\newblock Kronecker-factored approximate curvature for modern neural network
  architectures.
\newblock In \emph{Advances in Neural Information Processing Systems
  (NeurIPS)}, 2023.

\bibitem[George et~al.(2018)George, Laurent, Bouthillier, Ballas, and
  Vincent]{george2018fast}
George, T., Laurent, C., Bouthillier, X., Ballas, N., and Vincent, P.
\newblock Fast approximate natural gradient descent in a kronecker-factored
  eigenbasis.
\newblock \emph{Advances in Neural Information Processing Systems (NeurIPS)},
  2018.

\bibitem[Grosse \& Martens(2016)Grosse and
  Martens]{grosse2016kroneckerfactored}
Grosse, R. and Martens, J.
\newblock A kronecker-factored approximate {F}isher matrix for convolution
  layers.
\newblock In \emph{International Conference on Machine Learning (ICML)}, 2016.

\bibitem[Grosse et~al.(2023)Grosse, Bae, Anil, Elhage, Tamkin, Tajdini,
  Steiner, Li, Durmus, Perez, Hubinger, Lukošiūtė, Nguyen, Joseph,
  McCandlish, Kaplan, and Bowman]{grosse2023studying}
Grosse, R., Bae, J., Anil, C., Elhage, N., Tamkin, A., Tajdini, A., Steiner,
  B., Li, D., Durmus, E., Perez, E., Hubinger, E., Lukošiūtė, K., Nguyen,
  K., Joseph, N., McCandlish, S., Kaplan, J., and Bowman, S.~R.
\newblock Studying large language model generalization with influence
  functions, 2023.

\bibitem[Kingma \& Ba(2015)Kingma and Ba]{kingma2015adam}
Kingma, D.~P. and Ba, J.
\newblock {A}dam: A method for stochastic optimization.
\newblock In \emph{International Conference on Learning Representations
  (ICLR)}, 2015.

\bibitem[Kunstner et~al.(2019)Kunstner, Hennig, and
  Balles]{kunstner2019limitations}
Kunstner, F., Hennig, P., and Balles, L.
\newblock Limitations of the empirical fisher approximation for natural
  gradient descent.
\newblock 2019.

\bibitem[Lin et~al.(2024)Lin, Dangel, Eschenhagen, Bae, Turner, and
  Makhzani]{lin2024can}
Lin, W., Dangel, F., Eschenhagen, R., Bae, J., Turner, R.~E., and Makhzani, A.
\newblock Can we remove the square-root in adaptive gradient methods? {A}
  second-order perspective.
\newblock In \emph{International Conference on Machine Learning (ICML)}, 2024.

\bibitem[Martens \& Grosse(2015)Martens and Grosse]{martens2015optimizing}
Martens, J. and Grosse, R.
\newblock Optimizing neural networks with {K}ronecker-factored approximate
  curvature.
\newblock In \emph{International Conference on Machine Learning (ICML)}, 2015.

\bibitem[Mlodozeniec et~al.(2025)Mlodozeniec, Eschenhagen, Bae, Immer, Krueger,
  and Turner]{mlodozeniec2025influence}
Mlodozeniec, B., Eschenhagen, R., Bae, J., Immer, A., Krueger, D., and Turner,
  R.
\newblock Influence functions for scalable data attribution in diffusion
  models.
\newblock In \emph{International Conference on Learning Representations
  (ICLR)}, 2025.

\bibitem[Osawa et~al.(2023)Osawa, Ishikawa, Yokota, Li, and
  Hoefler]{osawa2023asdl}
Osawa, K., Ishikawa, S., Yokota, R., Li, S., and Hoefler, T.
\newblock Asdl: A unified interface for gradient preconditioning in pytorch,
  2023.

\bibitem[Papyan(2019)]{papyan2019measurements}
Papyan, V.
\newblock Measurements of three-level hierarchical structure in the outliers in
  the spectrum of deepnet {H}essians.
\newblock In \emph{International Conference on Machine Learning (ICML)}, 2019.

\bibitem[Papyan(2020)]{papyan2020traces}
Papyan, V.
\newblock Traces of class/cross-class structure pervade deep learning spectra.
\newblock \emph{Journal of Machine Learning Research (JMLR)}, 2020.

\bibitem[Paszke et~al.(2019)Paszke, Gross, Massa, Lerer, Bradbury, Chanan,
  Killeen, Lin, Gimelshein, Antiga, Desmaison, Kopf, Yang, DeVito, Raison,
  Tejani, Chilamkurthy, Steiner, Fang, Bai, and Chintala]{paszke2019pytorch}
Paszke, A., Gross, S., Massa, F., Lerer, A., Bradbury, J., Chanan, G., Killeen,
  T., Lin, Z., Gimelshein, N., Antiga, L., Desmaison, A., Kopf, A., Yang, E.,
  DeVito, Z., Raison, M., Tejani, A., Chilamkurthy, S., Steiner, B., Fang, L.,
  Bai, J., and Chintala, S.
\newblock {PyTorch}: An imperative style, high-performance deep learning
  library.
\newblock In \emph{Advances in Neural Information Processing Systems
  (NeurIPS)}. 2019.

\bibitem[Pearlmutter(1994)]{pearlmutter1994fast}
Pearlmutter, B.~A.
\newblock Fast exact multiplication by the {H}essian.
\newblock \emph{Neural Computation}, 1994.

\bibitem[Petersen et~al.(2023)Petersen, Sutter, Borgelt, Huh, Kuehne, Sun, and
  Deussen]{petersen2023isaac}
Petersen, F., Sutter, T., Borgelt, C., Huh, D., Kuehne, H., Sun, Y., and
  Deussen, O.
\newblock {ISAAC} newton: Input-based approximate curvature for newton's
  method.
\newblock In \emph{International Conference on Learning Representations
  (ICLR)}, 2023.

\bibitem[Reed et~al.(2022)Reed, DeVito, He, Ussery, and Ansel]{reed2022torch}
Reed, J., DeVito, Z., He, H., Ussery, A., and Ansel, J.
\newblock torch.fx: Practical program capture and transformation for deep
  learning in python.
\newblock \emph{Proceedings of Machine Learning and Systems (MLSys)}, 2022.

\bibitem[Ren \& Goldfarb(2019)Ren and Goldfarb]{ren2019efficient}
Ren, Y. and Goldfarb, D.
\newblock Efficient subsampled gauss-newton and natural gradient methods for
  training neural networks.
\newblock \emph{arXiv}, 2019.

\bibitem[Schraudolph(2002)]{schraudolph2002fast}
Schraudolph, N.~N.
\newblock Fast curvature matrix-vector products for second-order gradient
  descent.
\newblock \emph{Neural Computation}, 2002.

\bibitem[Soen \& Sun(2024)Soen and Sun]{soen2024tradeoffs}
Soen, A. and Sun, K.
\newblock Tradeoffs of diagonal fisher information matrix estimators.
\newblock 2024.

\bibitem[Tam et~al.(2024)Tam, Bansal, and Raffel]{tam2024merging}
Tam, D., Bansal, M., and Raffel, C.
\newblock Merging by matching models in task parameter subspaces.
\newblock \emph{Transactions on Machine Learning Research (TMLR)}, 2024.

\bibitem[Townsend(2017)]{townsend2017new}
Townsend, J.
\newblock A new trick for calculating jacobian vector products.
\newblock \url{https://j-towns.github.io/2017/06/12/A-new-trick.html}, 2017.
\newblock Accessed: 2025-03-22.

\bibitem[Wang et~al.(2019)Wang, Grosse, Fidler, and Zhang]{wang2019eigendamage}
Wang, C., Grosse, R., Fidler, S., and Zhang, G.
\newblock Eigendamage: Structured pruning in the kronecker-factored eigenbasis.
\newblock In \emph{International Conference on Machine Learning (ICML)}, 2019.

\end{thebibliography}
}

\appendix

\end{document}